\makeatletter
\@ifl@t@r{\fmtversion}{2021/01/01}{}{\@namedef{ver@hyperxmp.sty}{2026/01/01}}
\makeatother
\documentclass[sigconf,nonacm,9pt,balance=false]{acmart}
\setcopyright{none}
\acmYear{2026}
\acmDOI{}\acmISBN{}
\usepackage{amsmath,booktabs,tabularx,graphicx}
\usepackage{microtype}
\usepackage{placeins}

\newcommand{\Rset}{\mathcal R}
\newcommand{\Cset}{\mathcal C}
\hypersetup{pdfauthor={Tianren Li, Dongbo Li, Hongkai Chen, Jie Liu, Guoliang Xing},pdfsubject={Research preprint},pdfkeywords={aerial sensing, active perception, budgeted verification}}
\makeatletter
\def\@mkauthors@iii{%
  \hsize=\textwidth
  \global\setbox\mktitle@bx=\vbox{\noindent
    \unvbox\mktitle@bx\par\medskip
    \centering
    {\large Tianren Li\textsuperscript{1}, Dongbo Li\textsuperscript{1,\dagger},
    Hongkai Chen\textsuperscript{2}, Jie Liu\textsuperscript{1},
    Guoliang Xing\textsuperscript{2}\par}
    \smallskip
    {\normalsize
    \textsuperscript{1}Harbin Institute of Technology, China;\quad
    \textsuperscript{2}The Chinese University of Hong Kong, China\par
    \href{mailto:26s103297@stu.hit.edu.cn}{26s103297@stu.hit.edu.cn},\quad
    \{ldb,jieliu\}@hit.edu.cn,\quad
    \{hkchen,glxing\}@ie.cuhk.edu.hk\par}
    \bigskip}}
\def\@authornotes{\footnotetext[2]{Dongbo Li is the corresponding author (\href{mailto:ldb@hit.edu.cn}{ldb@hit.edu.cn}).}}
\makeatother

\begin{document}
\title[AeroBelief: Budgeted First-Action Verification]{AeroBelief: Budgeted First-Action Verification for Evidence-Guided Aerial Sensing}

\author{Tianren Li}
\affiliation{\institution{Harbin Institute of Technology}\country{China}}
\email{26s103297@stu.hit.edu.cn}
\author{Dongbo Li}

\affiliation{\institution{Harbin Institute of Technology}\country{China}}
\email{ldb@hit.edu.cn}
\author{Hongkai Chen}
\affiliation{\institution{The Chinese University of Hong Kong}\country{China}}
\email{hkchen@ie.cuhk.edu.hk}
\author{Jie Liu}
\affiliation{\institution{Harbin Institute of Technology}\country{China}}
\email{jieliu@hit.edu.cn}
\author{Guoliang Xing}
\affiliation{\institution{The Chinese University of Hong Kong}\country{China}}
\email{glxing@ie.cuhk.edu.hk}
\renewcommand{\shortauthors}{Li et al.}

\begin{abstract}
Aerial sensing must combine uncertain overhead context with current observations while budgeting prediction, movement and image processing. AeroBelief uses the next executable action as the unit of verification against reference scores. It groups candidate views by first action, scans them with an inexpensive proposer, and selectively queries a structured world model (WM). Calibrated group bounds support strict commitment; a six-query controller connects selective prediction with belief updates, acceptance checks and scheduled sensing. Across four layout families, 40 worlds and 960 closed-loop executions, the AeroBelief controller completes 75.42\% of tasks, compared with 73.33\% for exhaustive WM scoring and 65.83\% for line-of-sight geometry, using 91.81\% fewer reference-candidate evaluations than exhaustive scoring. In fixed-reference replay with a six-query cap, proposer-scaled anchored bounds raised first-action certificate availability from 36.25\% to 54.69\% and reduced mean queries per state by 13.54\% relative to unit-scale anchored bounds. A complete ground-hosted pipeline processes 2,685 self-collected 20-megapixel unmanned aerial vehicle (UAV) images from seven DJI Mavic 3M routes. Context-guided admission and feedback retain all eight independently supported target confirmations with 76.05\% fewer processed images and 70.02\% less accumulated active processing service than uncapped all-frame processing; feedback saves another 5.17\% of images relative to static context. The studies connect budgeted action selection with evidence-driven processing of recorded flight imagery.
\end{abstract}
\maketitle
\pagestyle{plain}

\section{Introduction}
An unmanned aerial vehicle (UAV) searching for a regional target must decide where to move, when to observe and which predictions to compute. Overhead imagery narrows the search region; current camera observations establish local target evidence. Cross-view benchmarks connect these perspectives~\cite{dai2024denseuav,zhu2023sues200}. Closing the sensing loop adds an operational question: which prediction is worth computing before the vehicle commits its next movement? 

Two uncertainties shape that decision. Positive and negative observations update an uncertain overhead prior. Future-view scores are uncertain too. A move may leave the target occluded but enable a useful view one step later. We score reachable views over a short horizon, execute the first move, and replan. A cheap proposer scores candidates before selective WM queries. Several paths may start north. Once north is verified as the best first action, their ranking cannot change it.

\begin{figure*}[t]
\centering\includegraphics[width=.98\textwidth]{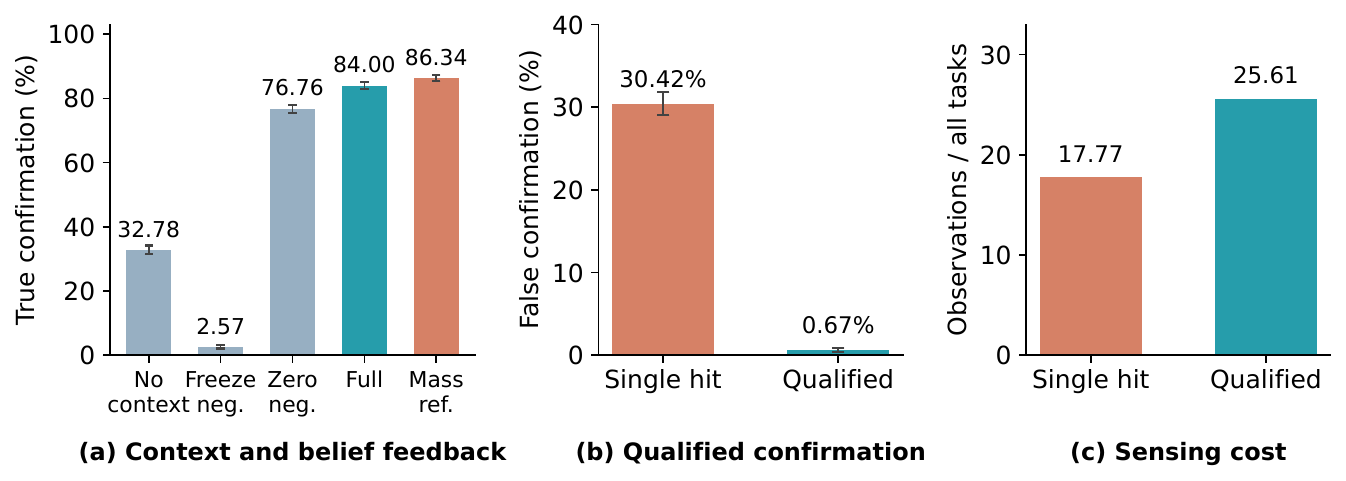}
\caption{\textbf{Context, feedback, and confirmation solve different sensing problems.} Six controllers share 4,320 tasks with known noisy-sensor likelihoods. (a) Context and likelihood-based negative updates improve true confirmation; the global-mass reference supplies a stronger baseline than full information-gain (IG) planning. (b) Qualified confirmation reduces false stops. (c) That reduction requires more observations. Intervals resample complete paired tasks; every timeout remains in the denominator.}
\Description{Context, negative updates, and qualified confirmation are tested separately. Full information gain confirms 84.00 percent of tasks; the stronger global-mass reference confirms 86.34 percent. Qualified confirmation reduces false confirmation from 30.42 to 0.67 percent at increased observation cost.}
\label{fig:motivation}
\end{figure*}

Figure~\ref{fig:motivation} isolates the roles of context, feedback and confirmation. Over 4,320 noisy-sensor tasks, likelihood-based belief updating confirms 84.00\% of targets, versus 32.78\% without context and 2.57\% with negative feedback frozen. Qualified stopping reduces false confirmation from 30.42\% to 0.67\% with more observations. A global-mass controller reaches 86.34\%, motivating strong planning baselines alongside evidence-handling controls for information-gain (IG) planning.

Information-driven trajectory and next-best-view planning connect informative views to motion~\cite{charrow2015information,bircher2016nbv}. A receding-horizon controller executes the first motion before observing again. AeroBelief makes this execution unit the verification unit: it groups viewpoints sharing a first motion and queries contenders that can change the winning group. A group's lower score bound exceeding every rival group's upper bound certifies the next action without resolving its best continuation.

\begin{figure*}[t]
\centering\includegraphics[width=\textwidth]{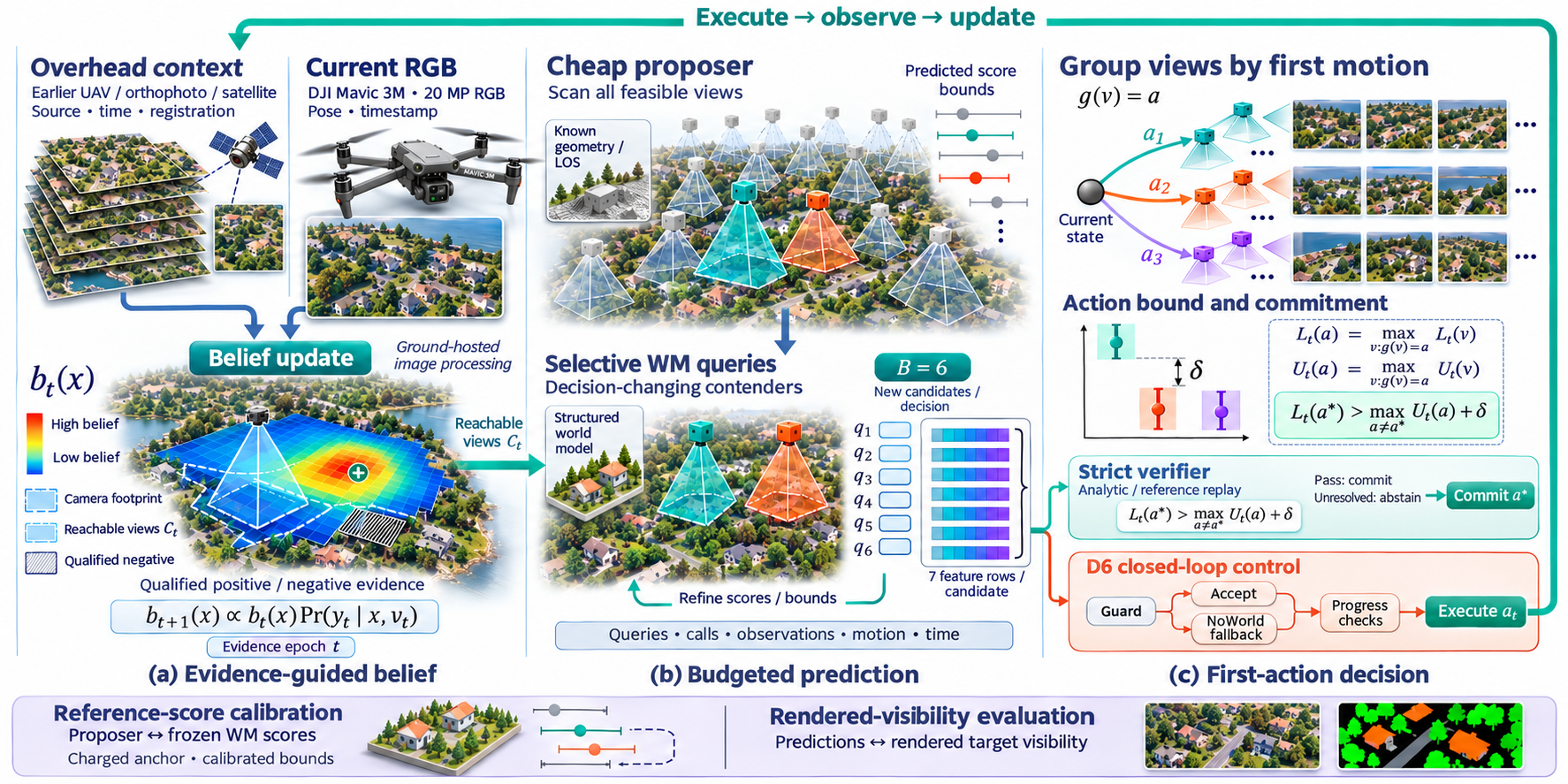}
\caption{\textbf{AeroBelief couples evidence updates with budgeted first-action decisions.} (a) Registered overhead context and qualified current RGB observations update the task belief $b_t$; DJI Mavic 3M supplies the self-collected 20-megapixel imagery. (b) A cheap proposer scans reachable views, while selective world model (WM) access evaluates at most $B=6$ new logical candidates per decision. Each engineering WM query evaluates seven feature rows; known geometry supplies line-of-sight (LOS) feasibility. (c) Grouping views by first motion supports two evaluated acceptance modes: the strict verifier commits on separated bounds or abstains; D6 combines Guard, belief-guided geometric fallback (NoWorld) and progress checks. The upper arrow closes the execution--observation loop. The offline band calibrates reference-score bounds and evaluates rendered visibility. The terrain, imagery and hardware form a conceptual illustration.}
\Description{A three-panel architecture illustration links source-qualified overhead context and DJI Mavic 3M RGB observations to a belief update, a cheap candidate proposer and selective world-model access. Six new logical candidate queries per decision are distinct from seven feature rows within each engineering query. The output branches to a strict commit-or-abstain verifier and a D6 controller using Guard, fallback and progress handling. Only executed D6 actions connect to the observation-update loop. The bottom band separates reference-score calibration from rendered-visibility evaluation.}
\label{fig:system}
\end{figure*}

Figure~\ref{fig:system} shows the resulting loop. Qualified observations update belief (a); a cheap proposer and selective reference queries allocate prediction work (b); action bounds or controller acceptance rules determine the first movement (c). The strict verifier commits on separated bounds. The three-dimensional AeroBelief controller with $B=6$ (D6) combines selective prediction with Guard, progress handling and a belief-guided geometric fallback (NoWorld). Evidence epochs invalidate obsolete predictions, and a ledger accounts separately for prediction, sensing, motion and time. The next observation then revises the belief for replanning.

This design places selective evaluation at two sensing interfaces. Before motion, the system admits reference predictions that can resolve an action comparison. Along a recorded flight route, it admits images whose footprints and observations can advance target confirmation. Both decisions use the current belief, with separate budgets for prediction and sensing. Evidence epochs connect new observations to prediction reuse, while capture and processing timestamps connect image admission to result readiness.

Table~\ref{tab:evidence} maps the evaluation populations and endpoints. Matched controls isolate grouping, query refinement and evidence feedback; common-input timing separates reference work from complete-decision latency.

The paper makes three contributions:
\begin{itemize}
\item \textbf{Verification at the execution unit.} First-motion groups avoid ranking continuations that lead to the same next action. On simultaneous score coverage, separated group bounds identify the first movement of every exhaustive reference optimum. Charged reference anchors and proposer-derived scaling increase resolution within the query budget.
\item \textbf{Prediction and sensing in one control loop.} Selective WM access, belief updates, Guard and progress handling connect the prediction budget to three-dimensional execution. Evidence epochs prevent reuse across changed beliefs, and a resource ledger distinguishes prediction, observation, motion and result readiness.
\item \textbf{Matched control and self-collected sensing evidence.} Four layout families, 40 worlds and 960 executions evaluate completion and reference work against matched controls; two public retrieval benchmarks evaluate observation budgeting. A complete pipeline processes 2,685 self-collected red--green--blue (RGB) photographs from seven DJI Mavic 3M routes. On fixed flight routes, belief-guided scheduling preserves independently supported target confirmations while processing fewer images than uncapped all-frame detection.
\end{itemize}

\section{Sensing problem and system design}
\label{sec:design}
\subsection{Context that guides, observations that establish evidence}
\label{sec:overhead-contract}
The context interface accepts prior UAV imagery, orthophotos, semantic maps or suitable satellite products. Each record supplies the source, task area, acquisition time or uncertainty, coordinate transform, registration residual and asset. Common coordinates align context with current observations. Initialization uses temporally eligible context; uncertain provenance or alignment broadens the prior.

Let $b_t(x)$ denote belief over target cells. An overhead cue acquired at $\tau$ provides a regional density, registered through offset distribution $\pi_R(\delta)$ and propagated through age kernel $K_{t-\tau}$. For an observed anchor $\tilde z_\tau$, one instance is
\begin{equation}
b_0(x)\propto\sum_{\delta\in\mathcal E_R}\pi_R(\delta)
K_{t-\tau}(x\mid\tilde z_\tau-\delta).
\label{eq:prior}
\end{equation}
Registration uncertainty enters through $\pi_R$; the scheduler receives no hidden true offset. In the flight replay, regional road/building masks supply a broad mixture with uniform support. Provenance, mask quality, registration confidence and target likelihood describe distinct parts of this interface. Sentinel-2's approximately 10\,m red--green--blue (RGB) sampling supplies regional context~\cite{copernicus2026s2}; current UAV observations establish local target evidence.

Current observations update the task belief. Pose, attitude, camera intrinsics, and height qualify a visible footprint; the known obstacle geometry used in three-dimensional control additionally qualifies line of sight (LOS). For visibility $\nu_t(x)\in[0,1]$, qualified detection probability $p_D$ and false-alarm probability $p_F$, the observation likelihood is
\begin{equation}
\begin{aligned}
\ell_1(x)&=p_F+(p_D-p_F)\nu_t(x),\\
\ell_0(x)&=1-\ell_1(x),\qquad
b_{t+1}(x)=\frac{b_t(x)\ell_y(x)}{\sum_u b_t(u)\ell_y(u)}.
\end{aligned}
\label{eq:belief}
\end{equation}
Here $y$ is the detector's binary output. A non-detection downweights locations where the sensor model predicts visibility and preserves relative support elsewhere. Observation identities prevent duplicate updates; correlated frames require spacing or conditional likelihoods. A belief, model, context, or feature-state change advances the cache epoch.

The controlled noisy-sensor experiment uses known likelihoods and binary observations. Its qualified stop requires a current positive, at least 95\% posterior mass in the observed footprint, and two positive footprints containing the maximum a posteriori cell. With footprint mass $M$, expected information gain (IG) is
\begin{equation}
I(M)=h\!\left(p_F+(p_D-p_F)M\right)-Mh(p_D)-(1-M)h(p_F),
\label{eq:ig}
\end{equation}
where $h$ is binary entropy. The flight-replay detector uses image-conditioned matching and a separate two-positive rule; its scores do not calibrate these simulated binary likelihoods. Manual-observation replay is a separate control.

\subsection{Reachable evidence and verification work}
Let $\Rset_t(h)$ contain feasible views reachable within $h$ motion steps. The analytic certificate benchmark combines IG, visible belief mass $M_t$, novel visible cells $N_t$, and distance $d_t$:
\begin{equation}
V_t(v)=I_t(v)+M_t(v)+0.012N_t(v)-0.025 d_t(v).
\label{eq:utility}
\end{equation}
The three-dimensional controller uses the composite score in Eq.~\ref{eq:engineering-utility}. Motion, sensor acquisition, and reference-score queries are counted separately. Only the first action toward a selected view is executed before replanning.

\subsection{Verifying the first executable action}
\label{sec:certificate}
For each feasible candidate $v\in\Cset_t$, a cheap proposer predicts $\widehat V_t(v)$ with interval $[L_t(v),U_t(v)]$. An analytic reference query evaluates $V_t(v)$ exactly and collapses that interval. Budget $B$ caps new logical candidate evaluations \emph{per decision}; cache hits are counted separately. A candidate certificate requires
\begin{equation}
L_t(v^*)>\max_{v\ne v^*}U_t(v)+\delta,
\label{eq:candidate}
\end{equation}
where $\delta\geq0$ is the declared separation margin. Views sharing a first motion need not be ranked against one another.

Let $g(v)$ map a viewpoint to its first feasible action. Group bounds are
\begin{equation}
L_t(a)=\max_{v:g(v)=a}L_t(v),\qquad
U_t(a)=\max_{v:g(v)=a}U_t(v).
\label{eq:group}
\end{equation}
The first-action certificate commits $a^*$ when
\begin{equation}
L_t(a^*)>\max_{a\ne a^*}U_t(a)+\delta.
\label{eq:action}
\end{equation}
\paragraph{First-action property.}
Let $V_t(a)=\max_{v:g(v)=a}V_t(v)$ and let $\mathcal E_t$ be the event that every candidate score lies in its declared interval. On $\mathcal E_t$, Eq.~\ref{eq:group} gives $L_t(a)\leq V_t(a)\leq U_t(a)$. If Eq.~\ref{eq:action} passes, then for every rival $a$,
\begin{equation}
V_t(a^*)\geq L_t(a^*)>U_t(a)+\delta\geq V_t(a)+\delta.
\label{eq:action-property}
\end{equation}
On $\mathcal E_t$, every finite-candidate maximizer therefore begins with $a^*$. Candidate certification implies action certification, while within-group rankings can remain unresolved. Inter-group ties cause abstention. Trajectory risk additionally depends on coverage along visited states.

Each query replaces an interval with its reference score, charges one unit against $B$ and retests the certificate. Certification authorizes the first action $g(v)$; continuations are replanned. Unqueried offline scores stay outside the scheduler. At an unresolved cap, the verifier abstains; enabled reactive fallback is logged separately.

\paragraph{Calibration at the decision unit.}
Calibration state $i$ uses residual $R_i=\max_v|V_i(v)-\widehat V_i(v)|$; known additive terms cancel when only IG is learned. Stratum $h$ uses $k_h=\lceil(n_h+1)(1-\alpha)\rceil$ at $\alpha=0.025$, abstaining if $k_h>n_h$. Its finite-sample rank interpretation assumes within-stratum exchangeability~\cite{angelopoulos2023conformal}. Calibration and test states have disjoint world keys and candidate records. Certificate availability, simultaneous coverage and task completion remain distinct.

\paragraph{Charged anchors and proposer-scaled bounds.}
The charged anchor separates a common prediction offset from differences in prediction error across candidates. Write $c_x(v)$ and $r_x(v)$ for cheap and reference utility. First query a stable cheap maximizer $v_0$, charging one of the $B$ queries. Let $e_x(v)=r_x(v)-c_x(v)$ and use the cheap-only scale
\begin{equation}
s(x)=\max\{1,\max_v|\widehat z_x(v)|\},
\label{eq:cheap-scale}
\end{equation}
where $\widehat z$ is the inexpensive proposer's output before known geometry terms. Calibration aggregates each world's maximum positive normalized relative residual,
\begin{equation}
E_w=\max_{x\in w,v}\left[\frac{e_x(v)-e_x(v_0)}{s(x)}\right]_+.
\label{eq:world-residual}
\end{equation}
For a supported order statistic $q_\alpha$, an unqueried candidate has upper bound
\begin{equation}
U_x(v)=r_x(v_0)+c_x(v)-c_x(v_0)+s(x)q_\alpha.
\label{eq:relative-upper}
\end{equation}
Queried scores provide group lower witnesses and collapse their own upper bounds. After the anchor, the largest observed reference score identifies the leading action. The scheduler queries the unqueried rival-action candidate with the largest upper bound, using candidate order to break ties, then updates the bounds and retests. State certificate availability means that a queried witness exceeds every rival-action upper bound by $10^{-10}$ within $B$ queries. Otherwise the verifier abstains. If calibration cannot support the required quantile rank, it abstains without querying. The replay makes no sensor acquisition or movement; unqueried reference scores cannot enter its scheduler or scale.

Calibration excludes the evaluated world and uses one residual maximum from each remaining world. The rank interpretation assumes exchangeable world scores and a fixed scorer, scale and aggregation rule. The certificate resolves the first action relative to the fixed reference scorer. Section~\ref{sec:scale-calibration} evaluates this verifier in replay.

\subsection{Structured implementation and resource accounting}
\label{sec:fidelity}
Two acceptance modes use the same first-action grouping. The strict verifier requires Eq.~\ref{eq:action} and abstains at an unresolved cap. The six-query AeroBelief controller (D6) also permits Guard acceptance with unresolved intervals, followed by shared progress handling and geometric fallback. Table~\ref{tab:evidence} distinguishes controller completion from strict certificate availability.

The structured WM predicts future observability from belief, pose and known geometry. An inexpensive candidate proposer (Student) scans candidate centers. The WM refines selected proposals using 56 belief, pose, geometry and observability features. The analytic proposer uses gradient-boosted belief/view features. D6 uses the composite score
\begin{equation}
\begin{aligned}
V_t^{\rm eng}(v)={}&z_t(v)+M_t^{\rm LOS}(v)\\
&+0.25\frac{A_t^{\rm new,LOS}(v)}{100\,{\rm m}^2}-0.05d_{\rm step}(v),
\end{aligned}
\label{eq:engineering-utility}
\end{equation}
where $M^{\rm LOS}$ is visible belief mass, $A^{\rm new,LOS}$ is novel visible area, and $d_{\rm step}$ is candidate motion depth. The fixed run configuration supplies these weights. The WM predicts $\log(1+\text{visible target pixels})$; $z$ averages seven local feature-row predictions after querying and uses the proposer's center estimate otherwise. The seven rows perturb the current pose in feature space while holding candidate paths fixed. The yaw rows equal the center under this encoding.

The following rules specify the D6 controller used in the three-dimensional closed-loop evaluation. It skips reference evaluation if the maximum proposer score is at most 0.25. Otherwise, it queries in upper-bound order, retaining the initial order among unqueried candidates. The queried witness must belong to the first-action group with the highest current composite score. Its Guard requires predicted visibility probability at least 0.25, raw-score margin at least 0.5 against rival actions, a nonnegative reported physical-score lower bound, and agreement in at least five of seven mixed raw-score action votes. Before reaching the 13\,m search phase, replacing the geometric fallback's upward action also requires margin 2.5. Each perturbation row casts one vote for the first-action group whose maximum raw candidate score is largest. Queried candidates use WM scores; unqueried candidates repeat their center proposer estimate across rows.

A low-value gate, Guard rejection, or model error selects the belief-guided geometric fallback. It ranks visible belief mass plus 0.01 times novel area (m$^2$) minus 0.025 times candidate distance (m), choosing the earliest horizon reaching 65\% of the best reachable value. Shared progress handling forces a feasible horizontal action after three consecutive nonhorizontal actions, preferring movement toward the belief mode. Queries and Guard checks occur within the decision; progress handling then selects the executed first action, followed by any scheduled acquisition and belief update. The ledger distinguishes fallback choices from progress overrides.

A WM query resolves approximation to that reference. Under proposer-to-WM and WM-to-physical error bounds with radii $q_1,q_2$, the physical-score radius is $q_1+q_2$ before querying and retains $q_2$ afterward. Rendered endpoint checks evaluate physical fidelity. Offline target visibility and counterfactual endpoints remain outside the online decision.

The ledger records attempted logical queries, seven-row evaluations, cache hits, acquisitions, movements and elapsed time. This separates reference work from the backend calls that batching combines. Cache reuse requires both the current evidence epoch and complete feature-state identity, keeping reused predictions consistent with the belief and model. Section~\ref{sec:methodology} defines complete-decision timing and simulated mission-cost measurements.

\section{Evaluation methodology}
\label{sec:methodology}
\paragraph{Endpoints and sampling units.}
Table~\ref{tab:evidence} separates sensing, control, reference-state replay, and image processing. Success denominators include every planned task: initialization errors, exceptions, collisions, timeouts, and abstentions. Analytic snapshot availability includes terminal no-operation states; reference replay requires an action certificate within the cap. Crossed-trajectory task availability denotes non-abstention, including 21 initial-sensing confirmations; completion requires the task endpoint. Cost comparisons use paired starts with valid initialization for the compared controllers, or the joint-success subset where stated. World-bootstrap 95\% confidence intervals (CIs) cluster starts, arms and timing repeats by world, stratifying by family where specified; retrieval intervals cluster by location.

\begin{table*}[t]
\centering\small
\caption{\textbf{Evaluation populations and endpoints.} Shared states and routes contribute once to their environment count. Reference-state diagnostics measure certificate resolution, not mission completion.}
\label{tab:evidence}
\begin{tabularx}{\textwidth}{p{2.7 cm}p{4.4 cm}X}
\toprule Study & Population & Endpoint \\
\midrule
Noisy-sensor controls & 4,320 matched tasks per policy; six policies & Context, negative evidence and qualified confirmation. \\
Empirical-noise transfer & 216 worlds; 6,480 evaluations & Development-route likelihoods and prior mismatch. \\
Crossed verification & 2,160 keys; 21,600 trajectories & Action grouping, refinement and budget frontier. \\
Independent calibration & 3,240 calibration / 6,480 test worlds & Simultaneous coverage and usable decisions. \\
Four-family control & 40 worlds; 960 executions; four arms & Completion, reference work and paired mission costs. \\
Two-regime policy control & 1,440 six-policy + 240 aligned Student executions; 1,680 total & Completion across difficulty regimes and proposer-only controls. \\
Strict factorial / replay & 480 executions / 10,459 common states & Reference agreement, interval cost and fallback. \\
Physical fidelity & 759-state rendering bank + 236-state endpoint bank & Endpoint prediction, action opportunity and calibration. \\
Correlated observations & 2,720 synthetic test worlds; 63,232 trajectories & Correlation-aware filtering, revisiting and fault controls. \\
Reference-state calibration & 40 reused worlds; 320 states; 15,477 candidates & Charged anchors and world-scale normalized certificate resolution. \\
Matched processor cost & 120 states; 11 cells; 3 rounds; 3,960 decisions & Native batch/serial contracts; 5,400 backend-scale tests. \\
Public-image transfer & DenseUAV 777 locations; University-1652 701 buildings & Retrieval quality versus query processing. \\
Self-collected image flow & 7 routes; 8 tasks; 2,685 photos; 64 jobs/segments & Processing, confirmation clocks, queueing and 226 review judgments. \\
\bottomrule
\end{tabularx}
\end{table*}

\paragraph{Controlled sensing and verification.}
A 25$\times$25 partially observable grid varies cue age, field of view, occlusion, and registration. Noisy-sensor controls cross $p_D=0.6/0.8/1$ and $p_F=0/0.02/0.05$ in 216 conditions, with a 55-motion cap and shared observation-index uniforms. Four primary contrasts use Holm adjustment and 10,000 paired-task draws. Crossed verification uses 1,560 seed clusters; independent calibration compares three interval families and six methods, including terminal zero-query states. Empirical-noise transfer fixes development-route likelihoods across five sensor/prior conditions and six policies. Correlated-observation tests use four generator families and development-only controller selection, with a hidden Markov model (HMM) for sensor dependence. Appendix~\ref{sec:s-current-availability} specifies the exported records and raw inputs for each study.

\paragraph{Three-dimensional control.}
ProjectAirSim~\cite{projectairsim2026} and Unreal provide six starts per world: three heights (6, 9, and 12\,m) crossed with two headings separated by $15^\circ$. The camera is $30^\circ$ from nadir; tasks have a 40-motion cap. Offline discovery requires at least 35 visible target pixels; evaluator segmentation is excluded from online scores. The four-family cohort compares four control arms: AeroBelief D6 selectively queries up to six candidates; exhaustive WM scoring F evaluates every candidate; controller-aligned proposer-only $S_{\mathrm G}$ uses Student predictions with the complete Guard; and LOS geometry G uses the belief-guided geometric policy without WM queries. Each family has ten worlds and 240 executions: layout perturbations (F1), longitudinal/side obstacles (F2), staggered transverse barriers (F3), and long/oblique barriers (F6). The 240-execution F6 extension follows inspection of F3's ceiling effect; all F3 outcomes are included. Two other families serve development only (Appendix~\ref{sec:s-fourfamilies}). Fixed policy blocks, D6/G/F/$S_{\mathrm G}$, match initial conditions but do not randomize session or host-load effects. Family-stratified paired intervals use 30,000 world draws, seed 541103, without multiplicity adjustment.

The prediction arms share the scoring, Guard, progress handling, and fallback rules of Section~\ref{sec:fidelity}. The inexpensive proposer (Student) and the structured WM have 20,777 and 24,938 parameters, respectively. The aligned proposer-only arm adds a 33-parameter visibility-probability head to the original proposer, bringing its total to 20,810 parameters. Appendix~\ref{sec:s-student-aligned} distinguishes the two proposer-only controllers. A shifted synthetic anchor and known static geometry supply context; a red-marker detector supplies observations. Mission time includes motion, stabilization, and sensing. Energy denotes simulator battery telemetry.

\paragraph{Fidelity, strictness, faults and cost.}
The rendering and endpoint banks contain 37,453 and 10,835 candidate labels. The endpoint bank contains 236 states from NoWorld trajectories in a 20-world cohort: 176 all-zero, 52 qualifying-opportunity, and eight subthreshold states. Fixed-state comparisons evaluate frozen selectors on this common bank; group opportunity is the best rendered continuation within the selected first-action group. Development probability mapping uses 33 states/1,248 candidates. The $2\times2$ strict-policy factorial crosses candidate/action certification with point/upper-bound query priority: 480 executions over 120 starts. Its same-input audit contains 10,459 states and 473,189 candidates; reference-interval adaptation replays these states. Fault evaluation has 64 executions and 25 local boundary tests.

Central processing unit (CPU) timing uses balanced common inputs and single-thread 32-bit floating-point inference on a Windows 11 Ryzen 7 6800HS host. The access cells compare full/reference-shortlist batching and selective/Student control. Complete-decision timing includes candidates, LOS, 56-feature construction, models, Guard, and fallback; it excludes image inference, simulator motion, and loading. Logical candidates, seven-row inputs, and backend invocations are counted separately.

\paragraph{Real images and processing.}
The stopping rule is frozen on SUES-200~\cite{zhu2023sues200} (120 training/80 test locations); reanalyses reuse those 80 test locations. DenseUAV~\cite{dai2024denseuav} supplies 777 query locations/2,331 UAV images and a 3,033-location/18,198-image satellite gallery. We apply the stopping rule fixed on SUES-200 to five frozen DenseUAV retrieval backbones. CPU timing uses 60 hash-selected locations, five backbones, seven rules, and two rounds (4,200 records). University-1652~\cite{zheng2020university1652} uses 631 train/70 validation buildings, five seeds, and 701 test buildings for adaptation, with frozen transfer evaluated separately; its UAV images are rendered.

Seven DJI Mavic 3M routes supply eight target-query tasks and 2,685 distinct 20-megapixel RGB photographs. Shared-route images yield 2,942 task-frame labels; two tasks reuse earlier target instances. Query templates are prepared retrospectively from the evaluated routes; five source photographs occur after the combined policy's confirmation time. Template-source frames are excluded from confirmation. Regional masks and flat-ground camera footprints supply replay context; known obstacle geometry is used by the three-dimensional controller. Independent review has 226 initial judgments and eight targeted adjudications, preserving original judgments. Four policies produce 32 complete original-cadence runs and 32 eight-times-rate segments of the first 120 capture seconds. Logs cover reading, decoding, detection, projection, belief, scheduling, and confirmation on a shared CPU host; result-ready clocks start after detector initialization and include processing and queueing, while capture timestamps index policy budgets.

\paragraph{Implementation and label provenance.}
OpenAI Codex and ChatGPT assisted setup, software implementation, replay orchestration, statistical analysis and plotting code in the analytic-verification, normalized-reference-replay and correlated-observation studies. OpenAI tools also assisted the plotting code for the table-derived physical-action summary. OpenAI generated the conceptual module illustration in Figure~\ref{fig:system}; numerical charts use recorded outcomes. Evaluation uses final human-adjudicated labels and independent review. Appendix~\ref{sec:s-current-availability} documents the 30 preliminary machine-assisted label proposals.

\section{Evidence and verification mechanisms}
\label{sec:mechanisms}
\subsection{Context and current feedback improve confirmation}
\label{sec:motivation-v46}
The controller using likelihood updates confirms 3,629 targets, versus 1,416 without context and 111 with negative feedback frozen. Paired gains are 51.23 and 81.44 percentage points, with pointwise 95\% CIs of [49.54, 52.89] and [80.25, 82.59], respectively. Zeroing an observed region after a miss confirms 3,316; retaining uncertainty through likelihood updates adds 7.25 points [6.23,8.26].

Six policies share task inputs and observation-index uniforms. Confirmation tests final-footprint target membership after traces are sealed. Figure~\ref{fig:motivation} includes all 4,320 tasks per policy, including disconnected worlds and timeouts.

Requiring a posterior threshold and two positive observations reduces false confirmations from the single-hit rule's 1,314 (30.42\%) to 29 (0.67\%). The paired reduction is 29.75 points [28.40,31.13], with mean observations increasing from 17.77 to 25.61. All four primary contrasts pass Holm adjustment. The global-mass planner confirms more targets than the information-gain (IG) planner: 3,730 (86.34\%) versus 3,629.

\subsection{Grouping and query refinement resolve different comparisons}
\label{sec:components-v46}
Combining action grouping and query refinement improves availability over either component alone. At $B=6$ per decision, candidate certification (C), candidate plus query refinement (C+Q), action grouping (A), and action grouping plus refinement (A+Q) have task-availability counts 26, 211, 153, and 453. Completion counts are 26, 211, 153, and 450. Grouping adds 5.88 availability points [4.80,7.00]; refinement after grouping adds 13.89 [12.39,15.36]. Ten arms execute 21,600 trajectories on 2,160 matched task keys from 1,560 environment-seed clusters, holding sensing, cache, fallback, and commitment semantics common (Table~\ref{tab:crossed-six}).

Grouping merges views sharing a first motion; refinement replaces selected intervals with reference scores. These analytic episodes use strict certification, with query work separate from motion and sensing.

\begin{table}[b]
\centering\small
\caption{Matched mechanism controls at $B=6$ per decision. All 2,160 planned tasks per arm remain in the denominator, including failures and abstentions. Task availability is the reported non-abstention endpoint.}
\label{tab:crossed-six}
\begin{tabular}{lrr}\toprule
Rule & \shortstack{Available\\tasks} & \shortstack{Successful\\tasks} \\\midrule
Candidate certificate & 26 & 26 \\
Candidate + query refinement & 211 & 211 \\
First-action certificate & 153 & 153 \\
First-action + query refinement & 453 & 450 \\\bottomrule
\end{tabular}
\end{table}

A+Q at $B=6$ uses 9.60 queries per task and 2.57 per visited decision; a task spanning multiple decisions can exceed six queries. One of 6,359 commitments disagrees with the canonical reference action on an uncovered state, with zero action-group value gap. Caps 12 and 24 raise task availability to 667 and 924 and completion to 660 and 877, at 22.86 and 51.92 queries per task. At $B=48$, 1,340 tasks succeed versus 1,753 under exhaustive scoring, using 138.46 versus 958.76 mean queries per task. At this cap, 24 commitments incur positive utility loss on uncovered states. More queries improve resolution but cannot repair an already erroneous commitment.

\subsection{Independent calibration improves usable resolution}
\label{sec:calibration-v46}
Fresh calibration raises A+Q state availability at $B=6$ from 3,566/6,480 with transported nine-group widths to 5,260 with fresh nine-group widths and 5,312 with fresh age-conditioned widths: 55.03\%, 81.17\%, and 81.98\%. Paired gains are 26.14 points [25.11,27.25] and a further 0.80 [0.05,1.53], both passing their two-test Holm family. The comparison uses one predefined state per world, giving 3,240 calibration states and 6,480 independent test states. Six methods under three interval families yield 116,640 decisions on identical test inputs.

Simultaneous prequery coverage on 6,348 active states is 99.89\%, 97.78\%, and 97.89\%; availability also includes 132 terminal zero-query states. Two-sided 95\% binomial lower endpoints are 99.77\%, 97.39\%, and 97.50\%. Fresh calibration increases availability with observed coverage of 97.78\% or 97.89\%, against the nominal 97.5\% target; the transported intervals cover 99.89\%. A+Q at $B=6$ uses 3.55, 2.05, and 1.66 queries per test world. Canonical action disagreements are 1, 2 and 1, respectively, all at numerical ties with maximum group-value gap $2.22\times10^{-16}$.

\subsection{Scale normalization resolves more reference decisions}
\label{sec:scale-calibration}
\begin{figure*}[t]
\centering\includegraphics[width=.98\textwidth]{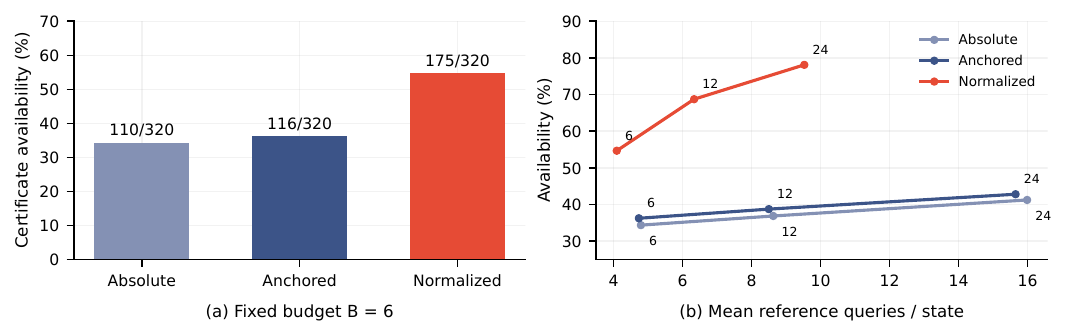}
\caption{\textbf{Proposer-derived scale normalization improves reference resolution.} (a) Three bound constructions share 320 recorded states from 40 worlds and a six-query budget, including the charged anchor. (b) Caps 6, 12 and 24 give state certificate availability against all-state mean reference queries. The frozen-reference bank is evaluated by paired replay.}
\Description{Absolute, anchored, and normalized bounds certify 110, 116, and 175 of 320 states at budget 6. Normalized counts increase to 220 and 250 at budgets 12 and 24.}
\label{fig:certificate-upgrade}
\end{figure*}

Scale normalization resolves more actions under the same fixed reference and query cap. At $B=6$, normalized anchor-relative bounds certify 175/320 states, compared with 116/320 for unit-scale anchor-relative bounds and 110/320 for absolute bounds (Fig.~\ref{fig:certificate-upgrade}). Relative to the unit-scale rule, availability rises from 36.25\% to 54.69\%, a paired gain of 18.44 percentage points (95\% world interval [12.19,24.69]), while mean reference queries fall from 4.731 to 4.091 (13.54\%). Normalization gains 66 commitments and loses seven, leaving 145 states unresolved. At caps 12 and 24, normalized availability is 220/320 and 250/320, versus 124/320 and 137/320 for the unit-scale rule.

Proposer outputs scale the bounds; charged queries supply reference scores. Replay of Eq.~\ref{eq:relative-upper} uses 320 states from 40 reused worlds, 15,477 candidates and a fixed scorer. Each evaluated world uses the other 39 worlds for interval calibration at nominal $\alpha=0.025$. The cap charges the anchor and every subsequent candidate reveal; averages include all 320 states. Repeated folds, budgets and the 32-world cross-fit controls reuse this bank.

Every certified action in this replay matches the reference action across the three bound constructions and three query caps. Appendix~\ref{app:normalization} gives calibration assumptions, unsupported-rank behavior and paired gains and losses.

Both anchored rules start by querying the candidate with the highest proposer score and use the same scheduling rule. The normalized rule uses proposer outputs to scale the bounds for unqueried candidates. The verifier stops once a queried score in the leading group exceeds every rival group's upper bound by the required margin. Unqueried continuations in the leading group share the same first action, so their ranking can remain unresolved.

\subsection{Revisit-aware search under correlated observations}
\label{sec:correlated-evidence}
\begin{figure*}[t]
\centering\includegraphics[width=.98\textwidth]{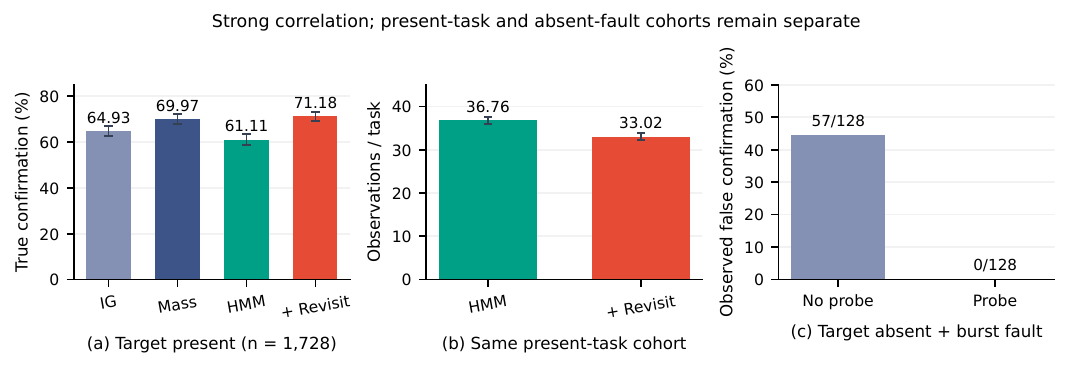}
\caption{\textbf{Inference, search and probing address distinct evidence failures.} At $\rho=0.85$, (a,b) target-present tests use 1,728 matched tasks in four families; error bars are pointwise fixed-family world-bootstrap intervals. (c) A separate 128-task target-absent burst-fault test records 57 versus zero false confirmations with probing. Present-target confirmation and sensing costs are reported alongside the fault-control result.}
\Description{Four controllers, IG, Mass, HMM, and HMM with revisit-aware search, confirm 64.93, 69.97, 61.11, and 71.18 percent of 1,728 paired tasks. Their mean observation counts are 34.35, 33.27, 36.76, and 33.02. A separate absent-target fault test records 57 versus 0 false confirmations in 128 worlds.}
\label{fig:correlation-upgrade}
\end{figure*}

Revisit-aware search improves confirmation while reducing observations under the synthetic correlated channel. At $\rho=0.85$, the hidden Markov model (HMM) with a development-selected revisit penalty confirms 1,230/1,728 target-present tasks (71.18\%), versus 1,056/1,728 (61.11\%) for the same filter without that penalty (Fig.~\ref{fig:correlation-upgrade}). The paired gain is 10.07 percentage points (95\% interval [7.93,12.15]); mean observations fall from 36.76 to 33.02, paired difference $-3.75$ [$-4.30$,$-3.18$]. Information-gain (IG) and global-mass (Mass) planning confirm 64.93\% and 69.97\%; Mass with the revisit rule reaches 70.78\%. The revisit-aware HMM's paired gain over Mass is 1.22 points [$-0.58$,3.01].

The revisit penalty discourages repeated low-value views, isolating the contribution of where to observe while HMM inference remains unchanged. The HMM maintains joint belief over target location and an observation regime. Development-route detection marginals are fixed at $p_D=0.6356$ and $p_F=0.000285$; temporal dependence is generated synthetically. The three regimes emit a forced positive, the visibility-dependent outcome, or a forced negative, with stationary probabilities $(p_F,p_D-p_F,1-p_D)$. The channel retains its regime with probability $\rho$ and otherwise redraws. Additional controls add an absence hypothesis or negative-control probe. Stopping requires posterior mass 0.95 and two positive observations within a 55-motion budget.

In this controlled study across four world families, development uses 192 worlds and 10,368 trajectories crossing controllers and channels; the held-out test has 2,720 task instances and 63,232 trajectories across target-present, target-absent and injected-burst cohorts. Policies and correlation settings are matched on task keys. 

The negative-control probe reduces observed false confirmations from 57/128 to 0/128 when the first eight observations are forced positive on 128 target-absent tasks. This comparison adds the probe to the revisit-aware HMM with an absence hypothesis. On ordinary present-target tasks, the same probe confirms 1,131/1,728 instead of 1,230/1,728 and uses more observations. Appendices~\ref{sec:s-correlated-observations}--\ref{sec:s-negative-control-probe} give all arms and their costs.

\section{Three-dimensional control and systems cost}
\label{sec:systems-results}
\subsection{Task completion and simulated resource use}
\label{sec:frozen-control-v51}
\begin{figure*}[t]
\centering\includegraphics[width=.98\textwidth]{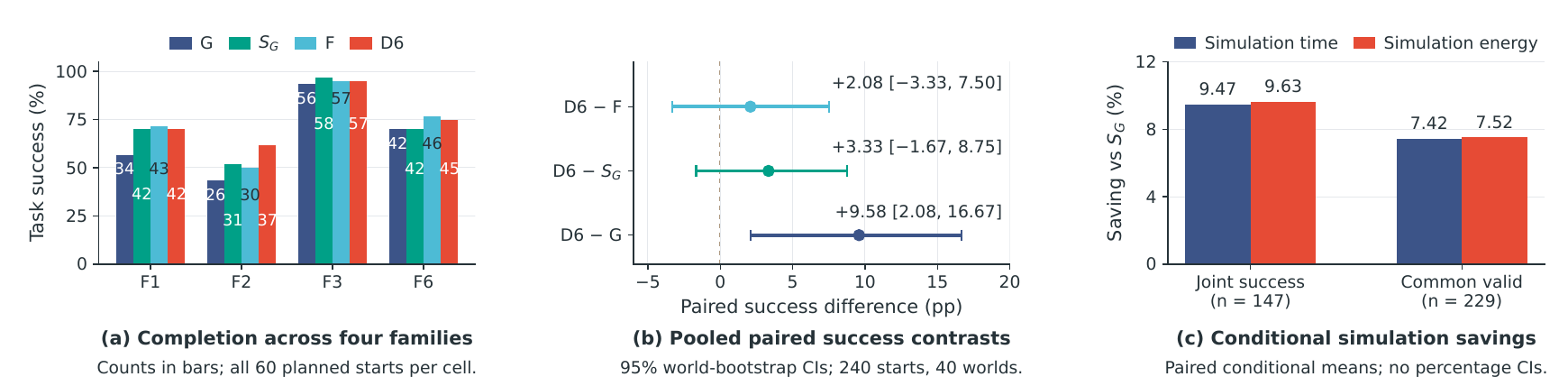}
\caption{\textbf{Four-family task completion and simulated resource use.} (a) Ten worlds per family and six starts per arm. (b) Paired completion differences with family-stratified world-bootstrap 95\% CIs; all planned starts are included. (c) D6 versus aligned proposer-only control on 147 joint-success and 229 common-valid starts, including timeouts in the latter. Costs include simulated motion, stabilization and sensing.}
\Description{D6,full reference,aligned Student and geometry complete 181, 176, 173 and 158 of 240 tasks. D6 minus geometry is 9.58 points, while D6 minus aligned Student is 3.33 points with interval spanning zero. Paired mission-time and energy costs decrease relative to aligned Student on the declared conditional sets.}
\label{fig:closed-loop-v47}
\end{figure*}

Across 40 worlds and 960 executions, D6 improves task completion over LOS geometry G by $+9.58$ percentage points, with a family-stratified world interval [2.08,16.67]. The four arms---AeroBelief D6, exhaustive WM scoring F, aligned proposer-only $S_{\mathrm G}$ and geometric G---each have 240 matched starts (Fig.~\ref{fig:closed-loop-v47}). They complete 181, 176, 173 and 158 tasks, respectively: 75.42\%, 73.33\%, 72.08\% and 65.83\%. D6's differences from F and $S_{\mathrm G}$ are $+2.08$ [$-3.33$,7.50] and $+3.33$ [$-1.67$,8.75]; both intervals span zero.

Across the closed-loop runs, D6 performs 14,453 WM candidate evaluations, compared with 176,451 for exhaustive WM scoring (F), a 91.81\% reduction in total candidate evaluations. These totals cover 5,309 decision steps for D6 and 5,525 for F along their respective trajectories. $S_{\mathrm G}$ and G make zero WM calls. Common-input latency is measured separately in Section~\ref{sec:cost-v47}.

On the 229 paired starts valid for all four arms, including timeouts, D6 uses 7.42\% less simulated mission time and 7.52\% less simulated battery energy than $S_{\mathrm G}$. Paired world intervals for D6-minus-Student cost are [$-8.66$,$-2.45$]\,s and [$-611.0$,$-176.7$]\,J. Among the 147 starts completed by both controllers, the reductions are 9.47\% and 9.63\%. Their intervals are [$-6.96$,$-2.62$]\,s and [$-492.3$,$-189.9$]\,J. D6 also has lower mean simulated mission time and energy use than F on jointly successful starts, but both intervals span zero when timeouts are included (Appendix~\ref{sec:s-fourfamilies}).

\subsection{Reference agreement and action utility}
Certifying first-action groups uses 4.39\% fewer queries than certifying individual candidates on 10,459 common states. The 480-execution strict factorial crosses these certification units with query priority based on point scores or upper bounds; completion-effect intervals span zero. Under the original empirical intervals, 33 of 2,309 submitted decisions disagree with a reference-optimal action. Wider calibration gives zero observed disagreements among 1,446 replay submissions but requires 2.86 times as many queries. Appendix~\ref{sec:s-strict-factorial} gives factorial outcomes; Appendix~\ref{sec:s-strict-risk} gives interval sensitivity.

In the fixed-state bank defined in Section~\ref{sec:methodology}, mean nominal first-step progress on 176 all-zero states is 0.235, 0.218 and 0.125\,m for exhaustive WM scoring, D6 and LOS geometry G. The paired D6--G gain is 0.092\,m under equal-world weighting (95\% interval [0.026,0.160]). On 52 states with qualifying future observations, Full and D6 each select a best first-action group on 21 states, versus 14 for G, counting ties. Across the 15 represented worlds, the equal-world gain is 10.00 percentage points [3.33,17.78]; the state-weighted difference is 13.46 points. Appendix~\ref{sec:s-action-physical} also reports the NoWorld fallback control and separates moving from nonmoving outcomes. We measure a selected first-action group's future opportunity by the largest rendered target-pixel count among its continuations and assess WM endpoint probabilities separately using the Brier score (0.0297 versus 0.0172 for a constant-absence predictor, with lower scores indicating better predictions). Depth-stratified score controls in Appendix~\ref{sec:s-prediction-controls} compare opportunity, first-step and endpoint fidelity.

We test whether residual corrections improve the frozen Student's first-action agreement with the WM. Seven learner families, including TabM~\cite{gorishniy2025tabm}, XGBoost~\cite{chen2016xgboost}, and Extra Trees~\cite{geurts2006extratrees}, share the 320 states from Section~\ref{sec:scale-calibration}. Extra Trees leads (302/320 versus 299/320 for Student), with a gain of 0.94 percentage points and an interval spanning zero [$-0.94$, 2.81]. Each of five folds for the new heads uses 24 training, 8 validation, and 8 test worlds. Validation selects 70 models from 140 fits. Appendix~\ref{sec:learned-query-controls} reports root-mean-square error (RMSE), action agreement, representations, and training scope.

\subsection{When fewer queries reduce decision time}
\label{sec:cost-v47}
D6 reduces complete-decision time when reference candidates are evaluated serially: it averages 35.074\,ms versus F's 41.683\,ms, a 15.86\% reduction, with a paired world interval for D6 minus F of [$-7.874$,$-5.319$]\,ms. With native batching, exhaustive WM scoring (F) takes 34.437\,ms. Serial-query D6 is 1.85\% slower than native-batched F, with a paired difference of 0.637\,ms [0.319, 0.948]. Full batches of 12 take 34.980\,ms, with an unresolved 0.094\,ms difference from D6. Matched timing uses 120 states, 11 access configurations and three rounds: 3,960 complete decisions and 5,400 backend-scale measurements.

\begin{table}[t]
\centering\small
\caption{\textbf{Complete-decision cost under matched access.} The same 120 states from 40 worlds are measured in three rounds. Geometry, features, models, Guard and fallback are included; camera and motion are excluded. Candidate means include gated decisions.}
\label{tab:main-cost-controls}
\begin{tabular}{lrr}\toprule
Policy/access & Time (ms) & WM candidates \\\midrule
Full, serial & 41.683 & 29.43 \\
Full, native batch & 34.437 & 29.43 \\
Top-6, native batch & 33.837 & 3.70 \\
D6, serial & 35.074 & 2.60 \\
Aligned Student, batch & 34.252 & 0 \\\bottomrule
\end{tabular}
\end{table}

Native-batched Top-6 is the fastest measured path on these 120 inputs (Table~\ref{tab:main-cost-controls}). Top-6 selects up to six candidates using proposer scores and evaluates them in one WM batch, while D6 queries candidates sequentially, updates the first-action comparison after each query, and can stop before reaching the six-query cap. On a separate 320-state scoring-core study, optimized Full takes 1.031\,ms versus 1.118\,ms for anchor-relative scoring of one batch of six candidates, including the anchor, with shared-feature and fused-row optimization applied to every path. This kernel timing excludes geometry and physical Guard; the normalized verifier is untimed. Figure~\ref{fig:kernel-upgrade} in Appendix~\ref{sec:shared-implementation} gives the implementation comparison.

On the same inputs, F and D6 evaluate an average of 29.43 and 2.60 candidates with the WM, respectively, while the aligned proposer-only baseline ($S_{\mathrm G}$) makes no WM calls and averages 34.252\,ms. Exhaustive scoring selects the same candidate and first action across the tested batch configurations. With native batching, F scores all candidates in one backend call, whereas D6 may make several sequential calls to score fewer candidates. Candidate construction, line-of-sight checks, and feature construction account for most of the decision time. Selective querying therefore saves time only when the reduction in WM evaluation time exceeds the additional cost of proposer scoring and query scheduling.

\subsection{Fault isolation and bounded recovery}
\label{sec:recovery-v51}
Across 64 executions on four worlds with two starts each, the interface rejects all 48 injected faults and records 32 healthy-query and 16 valid-observation recoveries. Twenty-five local boundary cases test result admission, stale-state handling, epoch changes and budget exhaustion (Appendix~\ref{sec:s-local-followup}). Under bad-frame injection, 2/8 tasks complete. Timeout tests use completed simulator remote procedure calls (RPCs) and local blocked-work surrogates. Appendix~\ref{sec:s-hardened-runtime} reports fault-specific mission outcomes and runtime configurations.

\section{Real-image transfer and flight replay}
\label{sec:field-results}
\begin{figure*}[t]
\centering\includegraphics[width=.98\textwidth]{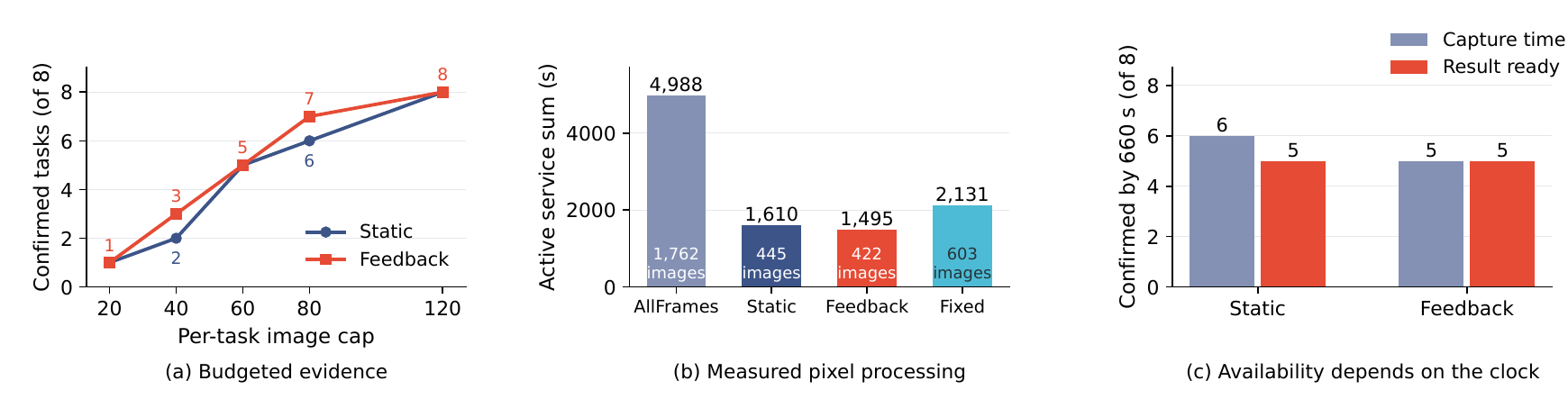}
\caption{\textbf{Complete real-image processing and confirmation availability.} (a) Static context and feedback at five per-task caps. (b) The combined system retains 8/8 confirmations with 76.05\% fewer images and 70.02\% less accumulated active service than uncapped AllFrames, on a shared ground host. (c) Capture-time and result-ready counts at 660\,s. Static context and feedback both finish with 8/8 two-full-image confirmations.}
\Description{At per-task caps 20,40,60,80,120, static and combined feedback confirm 1,2,5,6,8 and 1,3,5,7,8 of eight tasks. Their complete-job processing is 445 and 422 images, with shared-host service totals 1610.00 and 1495.23 seconds. At 660 seconds confirmation counts are 6 and 5 by capture time and 5 each by result readiness.}
\label{fig:mavic-v48}
\end{figure*}

\subsection{Image savings and retrieval accuracy}
\label{sec:dense-transfer}
In an exploratory evaluation on SUES-200 across 80 test locations and four altitudes, adaptive stopping reduced mean image use per query from 10 to 5.76 (42.40\%), with an R@1 of 71.56\% versus 72.63\% for fixed-ten aggregation. We then applied the frozen SUES-200 stopping rule to DenseUAV, where it used 2.60 images per query instead of three, a 13.33\% reduction. Across 777 locations, five backbones use frozen stopping rules and a complete gallery; mean location-task recall at rank one (R@1) is 82.059\% versus 82.394\% for fixed-three aggregation. The paired difference is $-0.335$ points (95\% location-cluster interval [$-0.746$,$+0.077$]). Fixed one- and two-image rules achieve 78.996\% and 81.338\%. Image-level R@1 and mean average precision (mAP) decrease by 0.412 and 0.315 points. Table~\ref{tab:dense-nine} compares frozen margin, entropy, matched-work and development-selected controls; Figure~\ref{fig:dense-full-v42} gives the full-gallery frontier.

The reduction in image work carries through to measured query processing on a 60-location CPU subset: 365.22\,ms for adaptive stopping versus 418.77\,ms for fixed-three, a 12.79\% reduction. Their subset R@1 values are 78.67\% and 78.33\%, respectively; the matched-work hash control achieves 77.00\% at 361.59\,ms. The 4,200 measurements cover decoding, preprocessing, feature extraction, full-gallery retrieval, fusion and stopping. Gallery construction is excluded.

\paragraph{University-1652: adaptation gain and stopping cost.}
Task-specific adaptation raises fixed-three R@1 by 29.586 points on 701 test buildings. Before adaptation, frozen fixed-three and adaptive R@1 are 52.725\% and 52.183\%, using 3 and 2.844 images. After training on 631 buildings with 70 validation buildings, the respective scores are 82.311\% and 80.228\% at 3 and 2.350 images. In the adapted system, stopping saves 21.67\% of image work at a 2.083-point accuracy cost. Table~\ref{tab:university-transfer} reports reverse-order sensitivity under the dataset's rendered UAV-view protocol.

\subsection{Complete image processing on fixed flight routes}
\label{sec:mavic-v48}
The complete ground-hosted system retains 8/8 confirmations supported by two fully visible target images while processing 422 rather than 1,762 images for uncapped all-frame processing (AllFrames), a 76.05\% reduction. Static-context scheduling also confirms 8/8 using 445 images; adding feedback saves a further 23/445 (5.17\%). Comparison with AllFrames measures the combined effect of context-guided admission, budgets and feedback; comparison with static context isolates feedback. All three policies share fixed-route images, the frozen detector and the stopping endpoint. Seven September 2025 routes captured by a DJI Mavic 3M provide eight target-query tasks and 2,685 distinct 20-megapixel RGB photographs~\cite{dji2026mavic}. The template and shared-route protocols follow Section~\ref{sec:methodology}. 

The frozen detector combines rotated templates at multiple scales, root-normalized scale-invariant feature transform (RootSIFT) descriptors~\cite{lowe2004sift,arandjelovic2012rootsift}, affine random sample consensus (RANSAC)~\cite{fischler1981ransac} and masked correlation. Its score uses image features; labels enter evaluation only. Final adjudication gives 89 fully visible, 11 partially visible and 2,842 absent task-frames. After excluding eight template-source positives, the detector identifies 58/81 full and 3/11 partial cases, with zero observed detections among the 2,842 negatives. Independent review supports target presence and center location in all 18 distinct confirming photographs. The expanded review records 218/226 initial agreements (96.46\%) and eight adjudications; final detection and review outcomes are in Figure~\ref{fig:s-mavic-review} and Appendix~\ref{sec:s-review-expanded}.

\paragraph{Processing budget and feedback controls.}
For a frame captured $t$ seconds after the route origin, the cumulative processing allowance is $q(t)=\min(120,1+\lfloor t/8\rfloor)$. Each selected frame consumes one image credit. A frame is eligible when 20 capture seconds have elapsed since the last processed frame, its footprint's belief mass is at least the corresponding uniform-prior mass, or a positive observation awaits confirmation. Template-source frames are excluded. Two processed positives within 30 capture seconds confirm; a processed negative resets the chain, while skipped frames leave it unchanged.

Positive-only and negative-only feedback process 434 and 428 images, respectively, also retaining 8/8 two-full-image confirmations. Fixed-interval processing confirms target presence in 4/8 tasks, with two-full-image support in 3/8; novelty-based processing confirms 5/8. The five random controls confirm 4, 7, 6, 3 and 4 of eight tasks. Task-presence confirmation accepts genuinely visible partial targets; two-full-image confirmation requires two full positives. The task-presence endpoint was clarified after the experiment; Appendix~\ref{sec:s-feedback-expanded} reports both endpoints for all controls.

Figure~\ref{fig:mavic-v48} reports the image-budget frontier. At caps 20, 40, 60, 80 and 120, static context confirms 1, 2, 5, 6 and 8 tasks; combined feedback confirms 1, 3, 5, 7 and 8. At 660 capture seconds, static and combined counts are 6/8 and 5/8; at 960 seconds both have 6/8. One task confirms at 679\,s with feedback versus 657\,s with static context because its footprint score falls below the admission threshold.

\paragraph{Pixel processing and queueing.}
Thirty-two original-cadence jobs execute reading, decoding, frozen detection, projection, belief updates, scheduling and confirmation; thirty-two eight-times-rate segments stress the first 120 capture seconds. All 3,963 completed detector calls preserve the frozen detector's binary decisions; Appendix~\ref{sec:s-pixel-stream} reports numerical parity and stage timings. A shared Windows 11 Ryzen 7 6800HS host (16 logical processors, 15.2\,GiB RAM) runs four policies with two OpenCV threads per process; peak resident memory is 367--369\,MiB.

Accumulated active processing service is 4,987.52\,s for AllFrames, 1,610.00\,s for static context and 1,495.23\,s for combined feedback. The complete system reduces this service sum by 70.02\% against AllFrames; feedback adds 7.13\% savings against static context. The service sum accumulates active pipeline processing; result-ready timestamps include queueing. Maximum queues fall from 244 frames for AllFrames to 15 for static context and 13 for feedback. Both policies provide 5/8 confirmations by 660 result-ready seconds. All 32 original-cadence jobs and 32 accelerated segments record zero first-in, first-out (FIFO) overload drops; post-confirmation cancellations are separate. Appendix~\ref{sec:s-pixel-stream} gives timing and stress details.

\section{Related work}\label{sec:related-work-v49}
\paragraph{Overhead context and active sensing.}
University-1652 benchmarks drone geolocalization~\cite{zheng2020university1652}, while TransGeo develops transformer-based cross-view matching~\cite{zhu2022transgeo}.
DenseUAV and SUES-200 characterize low-altitude and multi-height matching~\cite{dai2024denseuav,zhu2023sues200}.
AirLock+ registers UAV imagery to satellite maps for target geolocalization and geospatial augmentation~\cite{deng2026airlock}, while reference-only self-supervision reduces dependence on paired UAV data~\cite{amadei2026caevl}.
Retrieval selects an overhead reference, whereas registration supplies spatial correspondence. Search additionally needs a belief about where a target is currently observable.
Active perception and partially observable planning supply observation-conditioned decisions~\cite{bajcsy1988active,kaelbling1998pomdp}.
Next-best-view (NBV) chooses views that reveal useful geometry~\cite{connolly1985nbv}; frontier exploration exposes unknown space~\cite{yamauchi1997frontier}; information-based exploration incorporates localization uncertainty~\cite{bourgault2002information}.
Trajectory optimization and receding-horizon NBV connect information gain to feasible motion~\cite{charrow2015information,bircher2016nbv}, while rapid exploration and FUEL organize efficient UAV exploration~\cite{cieslewski2017rapid,zhou2021fuel}.
Sensor placement and adaptive submodularity offer selection guarantees under structural assumptions~\cite{krause2008sensor,golovin2011adaptive}; multi-robot and sampling-based information gathering account for travel-constrained observations~\cite{singh2009informative,hollinger2014sampling}.
AerialVLN and OpenFly focus on language-conditioned aerial navigation~\cite{liu2023aerialvln,gao2026openfly}; active visual-language navigation explicitly chooses informative observations~\cite{wang2023activevln}.
Language grounding specifies a navigation goal; our task belief represents uncertainty about a target's location and visibility as observations arrive.
AeroBelief retains observation-driven replanning, using source-qualified overhead cues as regional priors whose support reflects age and registration uncertainty.
Fresh positive and qualified negative observations revise target belief, separating useful map correspondence from current local evidence.
A non-detection reduces belief where the observation model predicts visibility, linking the next view to evidence already collected.
This feedback changes the value of revisiting a region after an expected detection fails to appear.

\paragraph{Budgeted prediction for the next action.}
Recurrent world models and PlaNet learn predictive representations~\cite{ha2018worldmodels,hafner2019planet}; Dreamer learns behavior through latent imagination~\cite{hafner2020dreamer,hafner2025dreamerv3}, and temporal-difference model predictive control (TD-MPC~2) combines world models with local trajectory optimization~\cite{hansen2024tdmpc2}.
DayDreamer extends model-based learning to physical robots~\cite{wu2023daydreamer}; RoboDreamer develops compositional robot imagination~\cite{zhou2024robodreamer}.
These methods develop predictive control; AeroBelief addresses the complementary question of which candidate predictions merit evaluation before motion.
Metareasoning values decision-relevant computation~\cite{russell1991metareasoning}; lazy search delays costly edge tests~\cite{dellin2016lazy}.
Fixed-confidence identification allocates samples to distinguish alternatives~\cite{kaufmann2016complexity,garivier2016optimal}, while structured identification propagates component uncertainty into action values~\cite{huang2017structured}.
The connection is structural: several candidate views can share one executable motion, and their individual scores contribute through a maximum over continuations.
Successive allocation budgets computation across partially evaluated alternatives~\cite{jamieson2016nonstochastic,li2018hyperband}.
Multi-fidelity optimization exploits cheaper approximations and hierarchical search~\cite{kandasamy2016multifidelity,kandasamy2017multifidelity,sen2018multifidelity}.
AeroBelief gives this allocation an execution-level target: queries need only resolve comparisons between rival first-action groups.
Resolving an ordering within the winning group would not change execution.
Here each query reveals a fixed scorer output, whereas bandit samples typically estimate an unknown mean; acquiring a new observation creates the next decision state.
Conformal regression supplies marginal coverage~\cite{angelopoulos2023conformal,lei2018distribution}; selective prediction trades acceptance against risk~\cite{elyaniv2010selective,geifman2019selectivenet}.
AeroBelief calibrates maximum residuals for simultaneous candidate bounds, since marginal coverage or expected-loss control~\cite{angelopoulos2024crc} alone does not certify every comparison.
State residuals maximize over candidates; world residuals additionally aggregate across recorded states.
Their rank-based interpretations require exchangeability at the corresponding sampling unit, with the scorer and aggregation rule fixed.
Covariate weighting, adaptive inference and time-series methods address specific shifts or dependence~\cite{tibshirani2019covariate,gibbs2021adaptive,xu2021dynamic}; beyond-exchangeability analysis further clarifies these conditions~\cite{barber2023beyond}.
Our simultaneous bounds address candidates at a decision, while confidence sequences address time-uniform validity under sequential inspection~\cite{howard2021confidence}.
When simultaneous candidate coverage holds, separated group bounds certify the reference's first action.
At an unresolved query cap the strict verifier abstains; D6 also permits Guard acceptance, with progress handling and geometric fallback.
Rendered observations assess the reference model's remaining error against physical visibility, separately from certified reference-action agreement.

\paragraph{Selective processing under resource constraints.}
Sensor placement and mobile sensing allocate information subject to communication and observation costs~\cite{krause2006placement,singh2006active}, and eSENSE schedules sensing activity~\cite{liu2006esense}.
The two Glimpse systems exploit recognition delay and camera-level early discard, respectively~\cite{chen2015glimpse,naderiparizi2017glimpse}.
They address when to spend recognition work on a captured stream, linking data selection to application responsiveness.
NoScope uses cascades for video queries~\cite{kang2017noscope}; DeepDecision considers inference and offloading~\cite{ran2018deepdecision}; Turbo exploits idle resources for edge analytics~\cite{lu2022turbo}.
DeepX, BranchyNet and adaptive computation modules reduce inference work through software optimization, early exits and conditional execution~\cite{lane2016deepx,teerapittayanon2016branchynet,wojcik2025adaptive}.
These mechanisms change the cost of processing admitted data; belief-based admission also changes which observations enter the task's evidence update.
Recent sensing systems couple resource choices to communication and complementary observations.
Grace dispatches remote-sensing inference between satellites and ground stations~\cite{li2026grace}.
FusionBridge exchanges cross-modal hints between independently trained camera and light detection and ranging (LiDAR) perception models through a lightweight adapter~\cite{wanniarachchige2026fusionbridge}.
EventEye merges roadside event-camera bird's-eye-view occupancy with vehicle LiDAR maps to improve the timeliness of perception~\cite{xia2026eventeye}.
These systems improve distributed perception by choosing where inference runs or which modalities contribute.
AeroBelief uses the current target belief to select images and budgets reference queries before motion. Fixed-route replay measures image-processing savings, while closed-loop experiments track sensing and motion. We report candidate evaluations, backend calls, and latency separately because batching changes the cost of reference access.

\section{Deployment implications}
Reference access and observation readiness determine the deployment configuration. For $m$ candidates, let $T_{\rm shared}$ include all common decision work: geometry/features, cheap gating and shared policy logic. Let $T_{\rm p}$ be additional selective-scheduling work and $T_{\rm r}(b)$ reference cost for a backend batch of $b$ candidates. The two access paths have the accounting form
\begin{equation}
\begin{aligned}
T_{\rm full}&=T_{\rm shared}+\sum_i T_{\rm r}(f_i),\\
T_{\rm selective}&=T_{\rm shared}+T_{\rm p}+\sum_j T_{\rm r}(b_j).
\end{aligned}
\label{eq:deployment-cost}
\end{equation}
For full scoring, backend batches partition all candidates, $\sum_i f_i=m$. Native access uses one batch; serial access uses $m$ single-candidate calls. Selective queries save time when avoided reference work exceeds scheduling and batch-fragmentation costs. Recording both candidates and backend invocations identifies this measured crossover.

The ledger records candidate work and backend invocations, allowing a deployment to budget its actual reference interface. Section~\ref{sec:cost-v47} measures complete-decision latency on matched inputs and compares the selective controller with both serial and native-batched exhaustive scoring.

In fixed-route replay, overhead context initializes the target belief, and the scheduler uses that belief and calibrated camera footprints to select frames for a frozen detector. Capture timestamps determine the image-processing budget. A confirmation counts toward a deadline if its result is ready by then, including queueing and processing delays.

\section{Conclusion}
AeroBelief verifies the next executable action before ranking every continuation. First-action grouping and proposer-scaled bounds direct a limited reference budget toward comparisons that can change that action. Observation-driven belief updates, evidence epochs and scheduled sensing connect these comparisons to a complete sensing loop.

Across four layout families, 40 worlds and 960 simulated executions, the six-query controller completes 75.42\% of tasks versus 65.83\% for LOS geometry and 73.33\% for exhaustive WM scoring, with 91.81\% fewer reference-candidate evaluations. The matched studies identify the contributions of grouping, query refinement and calibration, and connect reference-work reductions to measured access costs.

Ground-hosted fixed-route processing of seven DJI Mavic 3M routes and 2,685 self-collected images retains all eight independently supported confirmations with 76.05\% fewer processed images and 70.02\% less accumulated active processing service than uncapped AllFrames; feedback adds 5.17\% image savings over static context. Together, these results establish the value of budgeting consequential predictions before motion and admitting current image evidence according to the evolving task belief.
\label{page:body-end}
\clearpage
\bibliographystyle{ACM-Reference-Format}
\bibliography{references}
\clearpage
\appendix
\raggedbottom
\setcounter{dbltopnumber}{2}
\renewcommand{\dbltopfraction}{0.95}
\renewcommand{\dblfloatpagefraction}{0.75}
\section{Evaluation protocols and reproduction inputs}
\label{app:theme-protocols}
\label{page:appendix-start}

The Appendix is organized by evaluation purpose. Section~\ref{sec:s-fourfamilies} details four-family control, and Section~\ref{sec:s-pixel-stream} details the self-collected image pipeline. Sections~\ref{sec:s-integrated-reference} and~\ref{sec:proposer-v42} give normalized reference verification and strict certificate controls; Section~\ref{sec:s-cpu-expanded} gives complete-decision cost. Sections~\ref{sec:learned-query-controls} and~\ref{sec:s-correlated-observations} cover learned reference controls and correlated observations. Table~\ref{tab:s-input-availability} specifies reproduction inputs. Supporting cohorts and label stages retain their own populations in Sections~\ref{sec:s-populations}, \ref{sec:s-student-aligned} and~\ref{sec:s-mavic-review}.

\subsection{Experiment populations and outcome definitions}
\label{app:protocols}
\label{sec:s-populations}
Simulator discovery, certificate availability, reference agreement, retrieval and reviewed camera confirmation use separate outcome rules. All planned starts count in task denominators; resource comparisons identify their common-valid subsets. Paired starts, policy arms, timing repeats and reanalyses do not add independent environments. Source-input availability is summarized in Section~\ref{sec:s-current-availability}.

The A/B/C/D four-policy control has 960 executions on 240 starts in 40 worlds; strict policies add 480 executions on those worlds. Budget controls use 360 executions on ten reused difficult worlds. Rendered calibration uses 80 calibration and 40 test worlds. Manual-observation replay uses four routes and 1,818 frames; the automatic image pipeline uses seven routes, eight tasks and 2,685 photographs. These are distinct populations, separate from the four-family simulation cohort.

The proposer uses 61,296 candidates from 1,231 disjoint belief states. A state-calibration study has 3,216 calibration and 3,204 test decisions, one per episode.

\paragraph{Inference.}
Task contrasts use complete paired outcomes. The noisy-sensor study resamples tasks; trajectory studies resample environment-seed or world clusters. Fixed-state calibration uses independent worlds, not candidate counts as the sample size. Reported intervals are pointwise unless a multiple-comparison family is named. Equal marginal success counts do not demonstrate policy agreement or noninferiority. Canonical action mismatch is literal identity disagreement; a positive utility gap greater than $10^{-10}$ is a separately reported loss measure.

\subsection{Controlled sensing and reference-study populations}
\label{sec:s-controlled-populations}

Two controlled software studies use synthetic 2D worlds. Study I evaluates correlated-channel controls on 1,944 task worlds (1,728 target-present and 216 absent-target stress), with 52,488 main trajectories and 5,832 power-control trajectories on those worlds. Study II selects among 18 revisit/probe configurations on 192 separate development worlds (10,368 trajectories), then evaluates 2,720 synthetic test tasks: 1,728 present, 864 absent and 128 independent forced-positive fault tasks (63,232 trajectories). All tasks are included: three geometry masks repeat, giving 2,717 distinct geometries and 2,720 geometry-plus-prior fingerprints. The learner, stored-reference query and numerical-kernel studies use a separate shared bank of 40 worlds, 320 states and 15,477 candidates. Repeated methods, channels and timings reuse their respective populations.

\subsection{Evaluation populations and resource contracts}
\label{app:operations}
\label{sec:s-operating-contract}
Table~\ref{tab:s-contract-summary} links each endpoint to its population and execution interface. Reference queries, backend latency, simulated cost and image confirmation use separate measurement units.

\begin{table*}[tp]
\centering\small
\caption{Operating contracts by endpoint and population. Reference-action agreement, physical utility and measured resource use are evaluated separately.}
\label{tab:s-contract-summary}
\begin{tabularx}{\textwidth}{p{2.4cm}p{3.3cm}Xp{3.4cm}}
\toprule Contract & Population & Measured finding & Interpretation \\
\midrule
Analytic action grouping & 2,160 matched task keys & At $B=6$, grouping plus refinement makes 453 tasks available and completes 450. & A first action can be resolved without a unique best continuation. \\
Engineering D6 control & 40 worlds; 240 starts/arm & D6/F/$S_{\mathrm G}$/G complete 181/176/173/158. D6/F candidate work is 14,453/176,451. & Descriptive four-family completion and total logical work. \\
Aligned-Student cost & 229 common-valid starts & Simulated time and energy decrease 7.42\% and 7.52\%; paired cost intervals exclude zero. & Includes timeouts, excludes unavailable initialization telemetry. \\
Serial CPU interface & 120 states; three repeats & Full/D6 total latency 41.683/35.074\,ms. & 15.86\% less time when reference access is serial. \\
Native-batch CPU interface & Same 120 states & Full/D6 total latency 34.437/35.074\,ms. & D6 takes 1.85\% more time with native batching. \\
Reference-interval replay & 10,459 historical states & Submitted disagreements 33/2,309 to 0/1,446; queries 10,116 to 28,922. & A reliability--availability--work trade-off on state replay. \\
Physical prediction & 236 rendered states & Brier 0.02972, development-mapped 0.02712, constant-absence 0.01717. & Resolving a reference score leaves physical-model error. \\
Actual pixel processing & Seven routes; eight tasks & Static/feedback 445/422 images, with final 8/8 fully supported confirmations. & Fixed-route query-conditioned processing; no WM flight control. \\
Result availability & Same eight tasks; 660\,s & Static/feedback 6/5 by capture time, 5/5 when results are ready. & Confirmation is available only after required processing. \\
\bottomrule
\end{tabularx}
\end{table*}

The image-budget comparison has ten static/combined cells on the same eight tasks.

The measured serial interface saves enough reference work to offset selective scheduling overhead; native batching reverses that latency ordering. Each interface is characterized by complete-decision cost, reference-action agreement and physical utility. Image admission results report confirmation alongside budget, processed-image count and result-ready delay.

\subsection{Analysis inputs and reproduction scope}
\label{sec:s-current-availability}
Trial, query, detection and review exports, timing rows, state ledgers, model/configuration identities, derived tables and analysis code support record-level recomputation of completion, paired world intervals, access costs, physical scores and confirmation timing. Table~\ref{tab:s-input-availability} specifies the raw-input boundary. Externally indexed inputs have names, sizes and hashes; complete byte availability and a working retrieval route have not been verified.

\begin{table*}[tp]\centering\small
\caption{Input availability by component. Included exports support numerical recomputation; re-executing simulation or pixel inference requires the corresponding raw inputs.}
\label{tab:s-input-availability}
\begin{tabularx}{\textwidth}{p{2.8cm}p{3.4cm}X}\toprule
Component & Included inputs & External or unresolved inputs \\\midrule
Simulation & Trial summaries, initial-cohort state ledger, selected historical archives and Student terminal trajectories & Two full four-family raw volumes are externally indexed, including the long/oblique extension's full decision/array streams. \\
Image processing/review & Detection, schedule, judgment and coordinate exports; some image/query assets & The complete 2,685-photograph collection and review-image sources are externally indexed; included subsets do not supply the full pixel input. \\
Rendered physical evaluation & Scalar candidate labels, fixed choices and progress mappings & Full render masks, complete belief/history, and 240-state candidate probabilities are not all included. \\
Calibration and parity & Fitted maps, probability summaries and parity flags & Some candidate-level fitting arrays are missing; raw reference-calibration world maxima and underlying historical CPU selections are externally indexed. \\
Reference-state replay & Query traces and fixed-score banks & Backbone training-world membership is unknown. \\\bottomrule
\end{tabularx}\end{table*}

The 236-state prediction-correspondence records support metric recomputation; selector checks cover the recorded inputs and inspected code paths. Their upstream fixture identifier is unresolved in the provided index.

Final detector metrics use human labels: 89 full, 11 partial and 2,842 absent. Thirty source rows also contain preliminary machine-assisted label proposals (14 present, 14 partial and two unknown), separate from the final human labels; the proposal model is unspecified. For D02 vehicle frame 138, the clean native-input hash identifies the score/detection source. Annotation scores and marked-image geometry have separate provenance, and the marked geometry is excluded from geometric validation.

Reviewer-coded exports contain initial judgments and adjudications; reviewer identities and displayed review pixels have not been independently authenticated.

\subsection{Analysis scripts and archived inputs}
\label{sec:s-analysis-inputs}
Analysis scripts recompute the 240 aligned-Student and 1,440 six-policy outcomes, paired world bootstrap results, Student score-tensor comparisons, three-horizon physical-label metrics and Mavic schedules/labels. The two Student trajectory archives include terminal records, configurations and controller code. Derived tables and figure data identify their source records.

\subsection{Calibration and uncertainty}
\label{sec:s-study-uncertainty}
Pointwise bootstrap intervals resample complete worlds within the four fixed families and condition on calibration and model fits. They exclude uncertainty from backbone pretraining, sensor-marginal estimation and real-location transfer, and are unadjusted for multiple comparisons.

\section{Belief updates and observation policies}
\label{app:theme-belief}
\subsection{Noisy binary observations}
\label{sec:s-noisy-observations}
The 4,320-task experiment crosses 216 condition cells with 20 task replicates. Six policies share world keys and observation-index random uniforms. Detection probabilities are 0.6, 0.8, and 1; false-alarm probabilities are 0, 0.02, and 0.05. A positive observation supplies a bit and footprint, never the hidden target coordinate. A negative observation updates belief through its likelihood rather than removing all mass from the observed footprint. Qualified stopping requires a current positive, at least 95\% mass in its footprint, and two positive footprints containing the maximum a posteriori cell.

Table~\ref{tab:s-noise} reports true confirmations, false confirmations, unconfirmed tasks, and all-task costs. The synthetic cost is motion plus 0.25 per observation, not a physical energy measurement. Table~\ref{tab:s-noise-tests} reports the four prespecified paired contrasts. The global-mass reference is stronger than the information-gain planner, confirming 3,730 rather than 3,629 tasks. The analysis includes all 18 disconnected worlds, 259 naive-zero resets, and 15 terminal reliability failures.
\begin{table*}[tp]\centering\small
\caption{Binary-detector experiment: all 4,320 tasks in each arm. Costs use every task. Total cost is the prespecified synthetic motion cost plus 0.25 per observation; it is not joules or UAV energy.}
\label{tab:s-noise}
\begin{tabular}{lrrrrrr}\toprule
Arm & True & False & Unconfirmed & Observations & Motions & Total cost \\ \midrule
full & 3629 & 29 & 662 & 25.610 & 24.610 & 31.013 \\
no\_context & 1416 & 17 & 2887 & 47.342 & 46.342 & 58.177 \\
frozen & 111 & 1201 & 3008 & 48.866 & 47.866 & 60.082 \\
naive\_zero & 3316 & 24 & 980 & 27.637 & 26.637 & 33.547 \\
single\_hit & 2733 & 1314 & 273 & 17.770 & 16.770 & 21.213 \\
mass & 3730 & 38 & 552 & 25.136 & 24.136 & 30.420 \\
\bottomrule\end{tabular}\end{table*}
\begin{table*}[tp]\centering\small
\caption{Four prespecified paired primary contrasts. Intervals use 10,000 paired-task draws; exact tests form one four-test Holm family. Underflowing probabilities are reported as base-10 logarithms.}
\label{tab:s-noise-tests}
\begin{tabular}{lrrrr}\toprule
Contrast & Difference(pp) & Pointwise 95\% CI & Gains/losses & $\log_{10}p_{\rm Holm}$ \\ \midrule
Full--no context: true & 51.227 & [49.537,52.894] & 2331/118 & -532.436 \\
Full--frozen negative: true & 81.435 & [80.255,82.593] & 3521/3 & -1050.064 \\
Full--naive zero: true & 7.245 & [6.227,8.264] & 423/110 & -43.538 \\
Single hit--full: false & 29.745 & [28.403,31.134] & 1285/0 & -386.221 \\
\bottomrule\end{tabular}\end{table*}

\subsection{Transported detector likelihoods in controlled worlds}
\label{sec:s-transported-likelihoods}
The empirical-noise extension uses 216 simulated worlds, five sensor/prior conditions and six policies, giving 6,480 trajectories and 260,455 recorded steps. Development-route detector likelihoods are frozen before these runs. In the primary condition, full/global-mass/no-context complete 175/176/71 of 216. The 6,480 trajectories apply measured likelihood estimates to a simulated observation channel, rather than physical flights or real-image inference. Source-channel checks account for finite-sample uncertainty; zero observed false alarms is not treated as a known zero probability.

\subsection{Likelihood mismatch and target-absent sensitivity}
\label{sec:s-likelihood-sensitivity}
Likelihood mismatch and target-absent counterfactuals yield 5,184 policy evaluations on 216 reused tasks, including 1,296 normal-path reproductions. These are sensitivity calculations with assumed likelihoods, not real detector calibration or new independent environments.

\subsection{Correlated observations and revisit policy}
\label{sec:s-correlated-observations}

Temporal channels are generated synthetically from fixed development marginals $p_D=0.6355932203$ and $p_{FA}=0.0002850627$: $T=\rho I+(1-\rho)\mathbf{1}\pi^\top$, $\pi=(p_{FA},p_D-p_{FA},1-p_D)$. Channel categories emit always positive, the visibility bit, or always negative. The hidden Markov model (HMM) filters the joint target/channel distribution. Each supplied channel-persistence condition selects a separate development calibration bank containing 64 independent synthetic 256-step sequences with known visibility. Online HMM updates receive only the binary observation and footprint; the instantaneous latent channel category remains hidden. Known- and calibrated-rho HMM arms produce identical traces here; at $\rho=0$, HMM, Mass and calibrated power also coincide. The Study I power control was frozen after the main study and reuses its worlds. The 2D ``Full'' arm is the IG/coverage planner, separate from neural 3D Full-WM.

At $\rho=.85$ on Study II's matched present tasks, holding HMM inference fixed and adding a development-selected revisit penalty .5 increases true confirmation from 1056/1728 (61.11\%) to 1230/1728 (71.18\%). The difference is $+10.0694$ pp, with CI $7.9282$ to $12.1528$ pp. Mean observations decrease from 36.764468 to 33.016782; the paired difference is $-3.747685$, with CI $-4.300969$ to $-3.181134$. Hovers decrease from 16.567708 to 4.431134; the difference is $-12.136574$, with CI $-12.616334$ to $-11.649884$. Against Mass (1209/1728), the difference is $+1.2153$ pp. Its CI spans $-0.5787$ to $3.0093$ pp, with one false confirmation versus zero. This comparison isolates revisit-aware search under the synthetic channel. Temporal transfer to real video was not evaluated.

Table~\ref{tab:recovered-correlation} shows confirmation and false stops across correlation levels.

\begin{table*}[tp]
\centering\small
\caption{Present-task correlation controls. Entries are true-confirmation percentages (false-confirmation counts), with 1,728 worlds per cell. Studies I and II use different worlds and are not paired. Pointwise intervals and family-specific results accompany the data.}
\label{tab:recovered-correlation}
\begin{tabular}{llrrr}
\hline
Study & Controller & $\rho=0$ & $\rho=.5$ & $\rho=.85$ \\
\hline
Study I & IG/coverage & 77.03 (0) & 72.74 (9) & 65.16 (0) \\
Study I & Mass & 83.51 (0) & 78.30 (9) & 71.70 (0) \\
Study I & No context & 32.47 (0) & 30.56 (14) & 27.08 (1) \\
Study I & Frozen & 2.20 (1) & 2.26 (14) & 2.26 (1) \\
Study I & Naive zero & 65.10 (0) & 64.47 (9) & 61.00 (0) \\
Study I & Single hit & 77.26 (15) & 73.67 (13) & 66.03 (0) \\
Study I & Fixed power & 76.50 (0) & 74.19 (9) & 66.15 (0) \\
Study I & HMM, known rho & 83.51 (0) & 76.74 (9) & 62.04 (0) \\
Study I & HMM, calibrated rho & 83.51 (0) & 76.74 (9) & 62.04 (0) \\
Study I & Calibrated power & 83.51 (0) & 66.55 (4) & 29.22 (0) \\
Study II & IG/coverage & 78.47 (0) & 74.94 (1) & 64.93 (0) \\
Study II & Mass & 83.16 (0) & 79.63 (1) & 69.97 (0) \\
Study II & HMM & 83.16 (0) & 77.72 (2) & 61.11 (1) \\
Study II & Calibrated power & 83.16 (0) & 65.74 (1) & 27.95 (0) \\
Study II & Revisit Mass & 84.14 (0) & 80.15 (2) & 70.78 (0) \\
Study II & Revisit HMM & 84.14 (0) & 80.61 (2) & 71.18 (1) \\
Study II & Null + revisit HMM & 84.14 (0) & 80.61 (2) & 71.18 (1) \\
Study II & Probe + null + revisit HMM & 81.02 (0) & 76.68 (1) & 65.45 (0) \\
\hline
\end{tabular}
\end{table*}

\subsection{Negative-control probe: fault suppression and task cost}
\label{sec:s-negative-control-probe}

The probe chooses a viewpoint with predicted target mass at most .05 and path at most six, requires a negative observation, then a compatible positive reconfirmation within 12 observations. It shares the .95 confirmation threshold, two-hit rule and 55-motion budget (at most 56 observations). Null prior .05 is selected on development data; null alone leaves the high-correlation primary result unchanged.

In the constructed absent-target fault cohort, eight initial observations are forced positive. False confirmations decrease from 57/128 for null+revisit HMM to 0/128 with the probe: $-44.53125$ pp $[-53.125,-35.9375]$. All 128 probe runs are unconfirmed, use 56 observations and cost 69. Target absence is specified by the task metadata, so true confirmation is structurally zero in this cohort. The conditional binomial upper endpoint for 0/128 is 2.841\%. Ordinary high-correlation absent tasks give 1/864 versus 0/864, too sparse to resolve a general false-rate difference. On present tasks, true confirmations decrease from 1,230/1,728 to 1,131/1,728: $-5.7292$ pp $[-6.8287,-4.6296]$. These results characterize the constructed synthetic fault and its budget/present-task costs.

Table~\ref{tab:recovered-fault-cost} gives absent-target false stops and probe costs.

\begin{table*}[tp]
\centering\small
\caption{Study II absent-target and fault costs at $\rho=.85$. The ordinary absent cohort has $n=864$ and the independent forced-positive fault cohort $n=128$. Cost is motions plus 0.25 times observations. Observation/cost intervals and exact false-rate intervals accompany the data.}
\label{tab:recovered-fault-cost}
\begin{tabular}{lrrrrr}
\hline
Controller & Absent false/864 & Fault false/128 & Fault unconf./128 & Fault obs. & Fault cost \\
\hline
IG/coverage & 1 & 125 & 3 & 3.656 & 3.570 \\
Mass & 1 & 125 & 3 & 3.648 & 3.561 \\
HMM & 1 & 58 & 70 & 31.766 & 38.707 \\
Calibrated power & 1 & 3 & 125 & 54.875 & 67.594 \\
Revisit Mass & 1 & 125 & 3 & 3.648 & 3.561 \\
Revisit HMM & 1 & 57 & 71 & 32.141 & 39.176 \\
Null + revisit HMM & 1 & 57 & 71 & 32.180 & 39.225 \\
Probe + null + revisit HMM & 0 & 0 & 128 & 56.000 & 69.000 \\
\hline
\end{tabular}
\end{table*}

\subsection{Fallback-triggered sensing on reused worlds}
\label{sec:s-sensing-toggle}
A separate sensing-toggle study executes 192 C/D trajectories on the same eight worlds. Forced-fallback sensing off/on yields C 36/48 versus 41/48 and D 36/48 versus 40/48. World-cluster success intervals include zero; observations increase by approximately 9.77 and 9.50 per task. The off arm uses periodic and geometric sensing, so it is not a no-sensing arm. More frequent observations add time and simulated battery cost.

\section{First-action verification and calibration}
\label{app:theme-verification}
\subsection{Conditional action certificate and calibration}
\label{sec:s-state-calibration}
For viewpoint $v$, write its declared reference bounds as $L(v)$ and $U(v)$. Let $g(v)$ denote its first feasible action. The group bounds are
\[
L(a)=\max_{g(v)=a}L(v),\qquad U(a)=\max_{g(v)=a}U(v).
\]
On the simultaneous-coverage event,
\[
\max_{g(v)=a}V(v)\geq L(a),\qquad
\max_{g(v)=b}V(v)\leq U(b).
\]
Hence $L(a)>\max_{b\ne a}U(b)+\delta$ identifies the exhaustive reference's first action. The same inequality does not select a unique continuation inside the group. A candidate certificate implies this action certificate, because its witness is attainable and its rival upper maximum includes every rival group. The reverse implication is unnecessary: views inside the winning group can overlap without changing the first motion.

At the cap, the strict verifier abstains if the inequality is unresolved. An engineering Guard or reactive fallback has a separate decision reason and does not inherit this certificate. Offline audited scores are used only for evaluation.

\paragraph{Independent state-max calibration.}
The analytic residual is the maximum absolute raw-information-gain error over all candidates in one state. Known utility terms cancel. With $n_h$ states in calibration stratum $h$, the rank is $\lceil(n_h+1)(1-\alpha)\rceil$ at $\alpha=0.025$; an unsupported rank abstains. The method-independent study has 3,240 calibration worlds, including 3,176 active states, and 6,480 test worlds, including 6,348 active states. One snapshot is predefined per world. Each of 54 field-of-view (FoV), registration, and age cells contributes 60 calibration and 120 test worlds. Occlusion and snapshot slots use independent seed channels. All 18 method/band configurations are fixed before access to test truth.

The transported nine-group band, fresh nine-group band, and fresh 54-group band use the same proposer and identical test candidates. Table~\ref{tab:s-calibration} gives every method. The two primary availability contrasts use paired-world draws and a two-test Holm family. Fresh-nine minus transported-nine is 26.142 percentage points [25.108,27.253]; fresh-54 minus fresh-nine is 0.802 [0.046,1.528]. Coverage lower endpoints are reported separately in the main paper. The 1/2/1 A+Q6 canonical discrepancies are numerical ties, with maximum group gap $2.22\times10^{-16}$. Fresh-nine non-refined methods have one positive gap of 0.0294342, shown in the table.
\begin{table*}[tp]
\centering\small
\caption{All 18 methods on 6,480 shared test worlds. Coverage uses 6,348 active states; availability includes 132 terminal no-operation states. C/A denote candidate/action certificates, and Q denotes query refinement. A positive gap exceeds $10^{-10}$.}
\label{tab:s-calibration}
\begin{tabular}{llrrrrr}\toprule
Band & Method & Available & Cov (\%) & Queries/world & Mismatch & Positive gap \\ \midrule
Transported 9 & C6 & 740 & 99.89 & 5.492 & 0 & 0 \\
Transported 9 & C+Q6 & 2,462 & 99.89 & 4.748 & 1 & 0 \\
Transported 9 & A6 & 2,172 & 99.89 & 4.312 & 0 & 0 \\
Transported 9 & A+Q6 & 3,566 & 99.89 & 3.548 & 1 & 0 \\
Transported 9 & A+Q12 & 4,151 & 99.89 & 5.993 & 1 & 0 \\
Transported 9 & A+Q24 & 4,994 & 99.89 & 9.598 & 1 & 0 \\
Fresh 9 & C6 & 2,153 & 97.78 & 4.326 & 4 & 1 \\
Fresh 9 & C+Q6 & 4,603 & 97.78 & 3.158 & 2 & 0 \\
Fresh 9 & A6 & 3,893 & 97.78 & 2.804 & 4 & 1 \\
Fresh 9 & A+Q6 & 5,260 & 97.78 & 2.050 & 2 & 0 \\
Fresh 9 & A+Q12 & 5,714 & 97.78 & 2.982 & 2 & 0 \\
Fresh 9 & A+Q24 & 6,140 & 97.78 & 3.965 & 2 & 0 \\
Fresh 54 & C6 & 3,113 & 97.89 & 3.448 & 0 & 0 \\
Fresh 54 & C+Q6 & 4,964 & 97.89 & 2.414 & 1 & 0 \\
Fresh 54 & A6 & 4,372 & 97.89 & 2.198 & 0 & 0 \\
Fresh 54 & A+Q6 & 5,312 & 97.89 & 1.661 & 1 & 0 \\
Fresh 54 & A+Q12 & 5,561 & 97.89 & 2.633 & 1 & 0 \\
Fresh 54 & A+Q24 & 5,893 & 97.89 & 4.049 & 1 & 0 \\
\bottomrule\end{tabular}\end{table*}

\subsection{Transport and rule-development controls}
\label{sec:s-transport-development}
The one-state-per-episode development study has 3,216 calibration states and 3,204 independent test states. The state-max band covers 3,135/3,204 states (97.85\%); 604 commit and 2,600 abstain at six queries. Three committed first actions disagree with the reference, all on uncovered states: conditional error is 3/604 (0.50\%). Of 3,240 planned episodes per split, 24/36 lack the selected snapshot and are not replaced.

Transport to a different 540-key trajectory population yields four available tasks and three successes at cap six, and five available tasks and three successes at cap 24; exhaustive scoring succeeds in 430/540. A seed-disjoint 540-key population gives candidate/first-action availability of 6/44 and successes of 5/43; exhaustive scoring succeeds in 454. A matched 540-key follow-up gives 13, 54, and 141 available tasks for candidate, first-action, and first-action-plus-query methods at cap six, with 140 successes for the last method. At cap 24, that method is available on 240 and succeeds on 235; mean task queries rise from 9.34 to 42.22.

Action grouping and the first-action-only follow-up arm were developed after inspecting preliminary transport results; these studies are exploratory development controls. Their bands and seed populations differ from the independent state-max calibration in Section~\ref{sec:s-state-calibration} and the crossed mechanism study in Section~\ref{sec:s-crossed-mechanisms}.

\subsection{Crossed mechanisms and the query frontier}
\label{sec:s-crossed-mechanisms}
The expanded trajectory experiment has 2,160 matched keys in 1,560 environment-seed clusters, ten arms, and 21,600 trajectories. The four cap-six arms cross candidate/action certification with query refinement. Sensing, cache validity, fallback, and certificate timing are held common. Additional A+Q arms change only the cap. All-task availability includes 21 common initial-sensing successes. Table~\ref{tab:s-budget} reports all arms. Increasing the cap reaches later states and raises both available and successful tasks; it also exposes additional uncovered states. All 24 positive-gap commitments at cap 48 occur on uncovered states. Strict abstention terminates these analytic episodes; query counts measure workload at differing completion rates.
\begin{table*}[tp]\centering\small
\caption{Complete budget frontier: 2,160 matched tasks per arm in 1,560 environment-seed clusters. Available tasks include 21 initial-sensing successes. Canonical mismatch and positive utility loss are distinct. Query means use all tasks.}
\label{tab:s-budget}
\begin{tabular}{lrrrrrr}\toprule
Arm & Available & Success & Abstain & Canonical & Positive gap & Queries/task \\ \midrule
C6 & 26 & 26 & 2134 & 1 & 0 & 6.081 \\
C+Q6 & 211 & 211 & 1949 & 0 & 0 & 8.588 \\
A6 & 153 & 153 & 2007 & 1 & 0 & 6.149 \\
A+Q0 & 78 & 78 & 2082 & 0 & 0 & 0.000 \\
A+Q3 & 293 & 293 & 1867 & 1 & 0 & 4.258 \\
A+Q6 & 453 & 450 & 1707 & 1 & 0 & 9.605 \\
A+Q12 & 667 & 660 & 1493 & 1 & 0 & 22.864 \\
A+Q24 & 924 & 877 & 1236 & 1 & 0 & 51.923 \\
A+Q48 & 1586 & 1340 & 574 & 26 & 24 & 138.464 \\
Exact & 2160 & 1753 & 0 & 0 & 0 & 958.763 \\
\bottomrule\end{tabular}\end{table*}

\subsection{Analytic budget frontiers}
\label{sec:s-analytic-budget-frontiers}
\begin{figure*}[tp]
\centering\includegraphics[width=.98\textwidth]{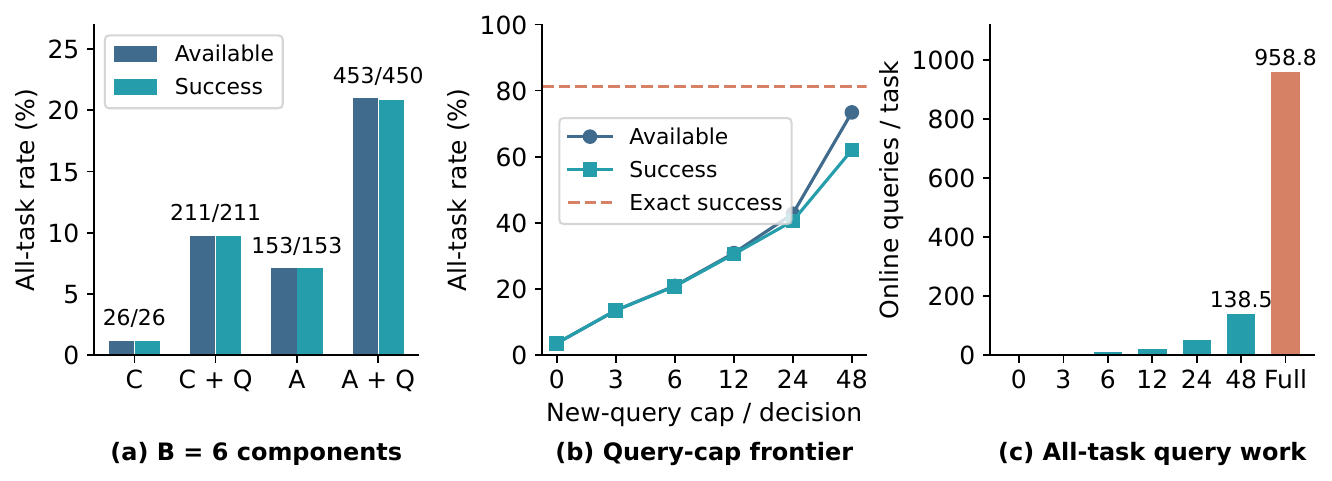}
\caption{\textbf{Analytic budget frontier.} Each arm uses 2,160 matched tasks in 1,560 seed clusters, including abstentions. (a) The $B=6$ factorial separates certificate granularity and query refinement. (b,c) Increasing the full method's cap raises availability, completion and all-task query work. Exhaustive scoring supplies the finite-candidate analytic reference.}
\Description{The crossed comparison has available-task counts 26, 211, 153, and 453. Larger budgets raise availability but use more reference queries. Availability, successful completion, and query work are separate endpoints.}
\label{fig:budget-v46}
\end{figure*}

The analytic budget sweep reports completion, availability and reference work with all timeouts and abstentions included (Fig.~\ref{fig:budget-v46}).

\subsection{Candidate and first-action verification on 18,133 states}
\label{sec:s-historical-strict}
Identical-state replay uses 18,133 states and 813,162 candidates. Both methods reject the same 9,895 states at the initial low-value gate. Table~\ref{tab:s-strict-state} gives certificate, Guard, coverage, and work endpoints. This replay uses a frozen marginal Student band, separate from the independent analytic state-max band. Every observed wrong reference certificate has a simultaneous interval miss.
\begin{table*}[tp]
\centering\small
\caption{Identical-state candidate/action comparison. All rates use 18,133 states unless the entered-state denominator is explicitly given. Gate rejection, certificate passing, Guard permission, and actual model action are distinct events.}
\label{tab:s-strict-state}
\begin{tabular}{lrr}
\toprule
Endpoint & Candidate certificate & First-action certificate \\
\midrule
Reference certificate passes & 5,904 (32.56\%) & 7,914 (43.64\%) \\
Guard allows model action & 2,795 (15.41\%) & 2,961 (16.33\%) \\
Allowed action differs from NoWorld & 2,424 & 2,559 \\
Reference-action discrepancies after Guard & 19/2,795 & 38/2,961 \\
New logical queries & 23,035 & 14,321 \\
Cap hits / entered states & 2,405/8,238 & 432/8,238 \\
Prequery simultaneous WM coverage & 66.83\% & 66.83\% \\
Postquery simultaneous WM coverage & 72.59\% & 71.97\% \\
\bottomrule
\end{tabular}
\end{table*}

SC and SA execute a separate 480 trajectories, succeeding on 199/240 and 202/240. Logical queries decrease from 5,989 to 3,653 (39.00\%). The paired table has 190 common successes, 12 SA rescues, nine reverse degradations, and 29 common failures; the success difference is 1.25 points [\,$-3.33$,5.42\,]. SC/SA certificate passes are 1,635/2,122, Guard-permitted passes are 789/815, and actual model actions after progress handling are 710/735. A reference certificate concerns only that branch and reference; fallback, progress replacement, and physical visibility are separate.

\subsection{Budget refinement on the 18,133-state bank}
\label{sec:s-fixed-state-budget}
On this 18,133-state bank, D6 averages 4.142 queries per entered state, with 61.68\% at the cap; D12 permits 15 additional suggestions. The 360 budget-control trajectories use the ten-world population in Section~\ref{sec:s-aligned-budget}; the 20-world cohort provides no executed D0/D2/D12 budget comparison.

\subsection{Strict factorial and complete same-input reference audit}
\label{sec:s-strict-factorial}
Candidate/first-action certification is crossed with point/upper-bound priority at six queries. Each executed arm has 120 starts on the same 20 worlds. The primary denominator includes 13 initialization failures; the common-start-valid sensitivity set has 116 starts. Table~\ref{tab:s-new-strict} separates executed work from identical-input replay.

\begin{table*}[tp]\centering\small
\caption{Strict factorial: separately executed trajectories and four-rule replay on the same 10,459 states. CP/CU use candidate certificates; AP/AU use first-action certificates.}
\label{tab:s-new-strict}

\begin{tabular}{lrrrrrr}
\toprule

Arm & Success & Run queries & Queries/entered & Replay queries & Certificates & Permitted \\
\midrule

CP & 91/120 & 2662 & 1.755 & 10352 & 5777 & 2318 \\

CU & 93/120 & 2618 & 1.758 & 10329 & 5793 & 2318 \\

AP & 95/120 & 2395 & 1.615 & 9902 & 5789 & 2302 \\

AU & 95/120 & 2441 & 1.634 & 9871 & 5796 & 2302 \\

\bottomrule
\end{tabular}
\end{table*}

First-action grouping changes completion by $+2.50$ points, with a 95\% interval of [$-2.08$,7.92]. Upper priority changes it by $+0.83$ points [$-2.50$,4.58]. Grouping changes world-equal queries per entered decision by $-0.094$ [$-0.137$,$-0.052$]. These executed comparisons include trajectory feedback.

The identical-input cache covers 10,459 states and 473,189 candidates. Averaging the two query priorities, first-action certificates reduce queries by 4.39\%. The corresponding world-equal effect is $-0.03350$ queries per state [$-0.04742$,$-0.02129$]; upper priority contributes $-0.00453$ [$-0.00830$,$-0.00145$]. Grouping slightly reduces permitted suggestions in this population, so it is not an unconditional availability improvement.

Of 5,788 certificate-passing decisions, 196 select a first action outside the reference optimum. Guard and progress handling leave 2,309 model submissions: 2,276 match the WM action-optimum set and 33 disagree. No reference contradiction occurs on a state satisfying simultaneous interval coverage; the deployed empirical bands do not always cover. Of 5,700 pre-cap stops, 3,177 are Guard rejections and 2,523 are pre-progress acceptances. Section~\ref{sec:s-strict-risk} reports disagreement at decision and trajectory units.

The 10,459-state factorial replay and 18,133-state comparison (Table~\ref{tab:s-strict-state}) use separate populations. Candidate-level strict-reference vectors are unavailable and their bank-level identity is unresolved in the index; per-state summaries do not supply these inputs (Section~\ref{sec:s-current-availability}).

\subsection{Strict certification crossed with query order}
\label{app:action-details}
\label{sec:proposer-v42}
\begin{figure*}[!t]
\centering\includegraphics[width=.98\textwidth]{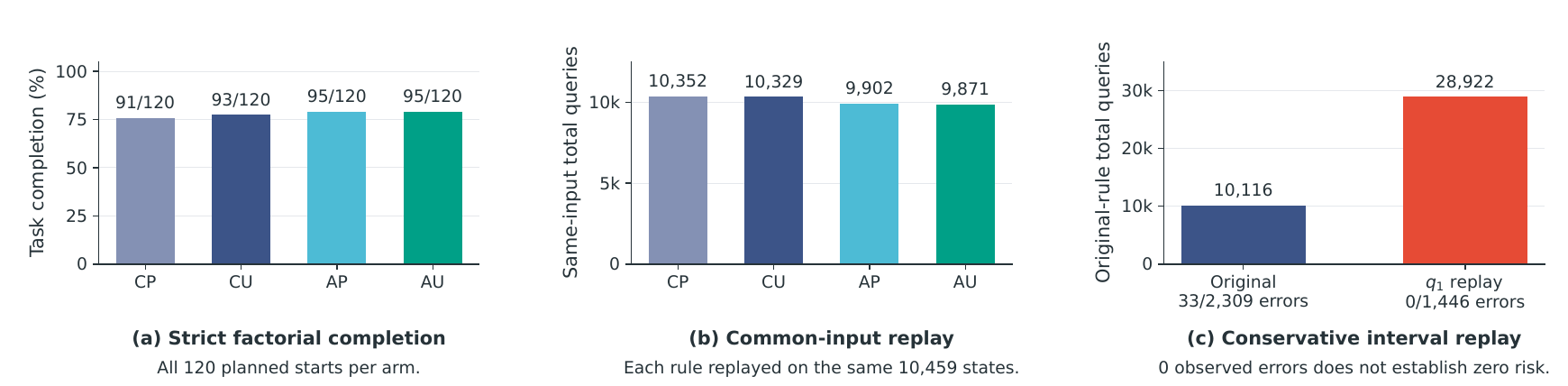}
\caption{\textbf{Strict certification and interval-width sensitivity.} (a) The four-cell trajectory factorial has 120 planned starts per arm. (b) Common-input replay isolates grouping and query-priority effects. (c) A wider reference interval changes observed submitted-action disagreements from 33/2,309 to 0/1,446 on the same 10,459 states, while queries increase from 10,116 to 28,922. The wider-interval result is fixed-state replay.}
\Description{Strict CP, CU, AP and AU succeed 91, 93, 95 and 95 of 120. Same-input query totals are 10,352, 10,329, 9,902 and 9,871. Conservative reference calibration changes 33 observed disagreements among 2,309 submissions to zero among 1,446, with 2.86 times as many queries.}
\label{fig:certificate-v47}
\end{figure*}

The 480-execution strict factorial crosses candidate/first-action certificates (C/A) with initial-point/upper-bound query priority (P/U). CP/CU/AP/AU complete 91/93/95/95 of 120 tasks (Fig.~\ref{fig:certificate-v47}). The world-equal first-action main effect is $+2.50$ percentage points [$-2.08$,7.92]; upper-bound priority contributes $+0.83$ [$-2.50$,4.58]. Both completion intervals span zero.

First-action arms average approximately 1.62 queries per entered decision versus 1.76 for candidate arms. The mechanism ledger records 5,788 certificates, 2,524 pre-progress accepts and 2,309 final model submissions. Among 5,700 pre-cap stops, 3,177 are certificate-triggered Guard rejections and 2,523 accept before progress handling. Early termination therefore includes reverting to the fallback.

Common-input replay covers 10,459 states with 473,189 candidate views. First-action rules use 4.39\% fewer queries, with a world-equal difference of $-0.0335$ queries per state (95\% interval $-0.0474$ to $-0.0213$). Upper-bound priority changes work by $-0.0045$ queries per state (interval $-0.0083$ to $-0.0014$). These effects use this factorial's state population.

Under the deployed empirical intervals, 33 of 2,309 submitted decisions disagree with Full-WM; 2,276 (98.57\%) belong to its co-optimal action set. Disagreements affect 32 of 467 observed trajectories (6.85\%); 13 of 480 planned runs enter no state. The theorem applies when simultaneous coverage holds, a condition the empirical intervals do not always satisfy. These observed disagreements concern the learned reference action, separately from physical-action correctness.

A separate replay uses $q_1=2.3873806$ from an 80-world/480-state calibration bank (finite-sample rank 80), holding each state's source rule and progress mapping fixed. The physical-error width is fixed at $q_2=6.3777392$. Submissions decrease from 2,309 to 1,446 and disagreements from 33 to zero, while queries increase from 10,116 to 28,922 (2.86$\times$). Across the same 5,983 entered decisions, mean queries rise from 1.69 to 4.83 and cap hits from 283 to 3,952 (66.05\%). Initial coverage is 10,457/10,459 states: two uncovered states occur in one world, leaving 19/20 worlds fully covered. This width is evaluated on fixed states and is not deployed in the four-family trajectories; raw calibration maxima are externally indexed (Section~\ref{sec:s-current-availability}).

\subsection{Empirical strict risk and model correction}
\label{sec:s-strict-risk}
In the strict factorial (Section~\ref{sec:s-strict-factorial}), submitted model actions disagree with a frozen-reference optimum in 33/2,309 decisions (1.429\%) and in at least one decision in 32/467 observed trajectories (6.852\%). Thirteen of 480 planned executions never enter a state. These reference-action disagreement rates use distinct decision/trajectory denominators and do not measure physical failure.

First-action grouping removes within-action comparisons conditional on simultaneous score coverage; it cannot repair an invalid interval that passes early. The reference-width replay in Section~\ref{sec:proposer-v42} evaluates different intervals on these same states; the rates above describe the deployed intervals. On a distinct reused 320-state bank, reference agreement is 308 for the student, 317 for D6 and 319 for fixed Top-6; D6 makes ten corrections and one regression. This is a fixed-state diagnostic.

\subsection{Learned reference-score controls}
\label{sec:learned-query-controls}
\label{app:learned}

Seven residual learner families share the same fixed Student output: TabM~\cite{gorishniy2025tabm}, XGBoost~\cite{chen2016xgboost}, Extra Trees~\cite{geurts2006extratrees}, a multilayer perceptron (MLP), LightGBM, histogram gradient boosting (HistGB) and ridge regression. Inputs are raw features plus Student (57 dimensions) or common-prefix features plus Student (105 dimensions). Each of five folds uses 24/8/8 world-disjoint train/validation/test worlds for the newly trained heads. Backbone training-world membership is unknown because its manifests are unavailable. Two configurations per family/representation give 140 fits and 70 selected models. Validation selects configuration, early stopping and shrinkage $\lambda\in\{0,.25,.5,.75,1\}$; $\lambda=0$ reproduces the common Student. Agreement uses co-optimal-action tolerance $10^{-7}$. Thirteen of fourteen agreement-difference intervals include zero and the remaining interval is negative, despite lower score root-mean-square error (RMSE) for several heads.

Table~\ref{tab:recovered-learners} reports action agreement, score RMSE and validation-selected shrinkage.

\begin{table*}[tp]
\centering\small
\caption{Validation-selected residual heads on 320 states (40 worlds). $\Delta$ is agreement minus the common Student in percentage points; paired 40-world bootstrap 95\% intervals condition on fixed out-of-fold fits. The last column counts folds that reproduce Student through zero shrinkage.}
\label{tab:recovered-learners}
\begin{tabular}{llrrlr}
\hline
Representation & Learner & Agree./320 & RMSE & $\Delta$ pp [95\% CI] & $\lambda=0$/5 \\
\hline
Shared & Student & 299 & 0.078214 & -- & 5 \\
Raw+S & TabM-small2025 & 296 & 0.072965 & -0.938 [-2.500,+0.625] & 1 \\
Raw+S & MLP & 300 & 0.077809 & +0.312 [-1.562,+2.500] & 0 \\
Raw+S & LightGBM & 299 & 0.073037 & +0.000 [-1.562,+1.875] & 0 \\
Raw+S & XGBoost & 299 & 0.074822 & +0.000 [-1.250,+1.250] & 0 \\
Raw+S & ExtraTrees & 302 & 0.072326 & +0.938 [-0.938,+2.812] & 0 \\
Raw+S & HistGB & 300 & 0.074219 & +0.312 [-1.875,+2.188] & 0 \\
Raw+S & Ridge & 299 & 0.078140 & +0.000 [+0.000,+0.000] & 3 \\
Prefix+S & TabM-small2025 & 297 & 0.066967 & -0.625 [-2.188,+0.938] & 0 \\
Prefix+S & MLP & 300 & 0.067544 & +0.312 [-1.875,+2.500] & 0 \\
Prefix+S & LightGBM & 298 & 0.068011 & -0.312 [-2.188,+1.875] & 0 \\
Prefix+S & XGBoost & 297 & 0.071722 & -0.625 [-2.500,+1.250] & 0 \\
Prefix+S & ExtraTrees & 298 & 0.069404 & -0.312 [-2.188,+1.562] & 0 \\
Prefix+S & HistGB & 299 & 0.070507 & +0.000 [-1.875,+1.875] & 0 \\
Prefix+S & Ridge & 295 & 0.076701 & -1.250 [-2.500,-0.312] & 0 \\
\hline
\end{tabular}
\end{table*}

\subsection{Stored-reference queries and charged anchors}
\label{sec:s-stored-reference-queries}

Queries return exact stored WM scores for charged candidates, measuring first-action certification and reference work. With 39-world leave-one-out calibration, nominal one-sided risk .025 and budget six, absolute-bound leader/rival certifies 110/320 states versus Top-K 103/320: $+2.1875$ pp, paired 95\% CI $[0.9375,3.7500]$. Mean queries are 4.790625 versus 4.859375, difference $-0.06875$ $[-0.115625,-0.031250]$. One submitted state has an uncovered initial bound; covered-subset counts are 109 versus 102. A post hoc fixed-bound certificate-cost ceiling is 110 overall and 109 on that subset. Leader/rival and optimistic-leader share 5,745/5,760 query sequences. Rank-39 coverage of 39/40 worlds follows largely from the rank construction. With 32 calibration worlds and risk .025, rank 33 is required (Section~\ref{sec:s-integrated-reference}): the radius is infinite, all 5,760 associated decisions abstain without querying, and finite-bound coverage is undefined.

The anchored analysis charges an initial cheap-selected anchor and calibrates relative residual bounds over every candidate/action. At the same calibration/risk/budget setting, relative leader/rival reaches 116/320 versus absolute 110/320: $+1.875$ pp $[0,4.0625]$. Mean queries fall from 4.790625 to 4.731250, difference $-0.059375$ $[-0.112500,-0.012500]$. The common-mode component contributes 8.784\% of squared residual energy; relative events widen in 169/320 states.

Table~\ref{tab:recovered-query} compares the charged query schedules.

\begin{table*}[tp]
\centering\small
\caption{Stored-reference query schedules on the same 320 states/40 worlds. Errors compare submitted actions with WM. The 32-world cross-fit has infinite radius, zero availability and no submissions; finite-bound coverage is N/A.}
\label{tab:recovered-query}
\begin{tabular}{llrrrr}
\hline
Bounds & Schedule & Certified/320 & Mean queries & Mean calls & Ref. errors \\
\hline
Absolute & action round robin & 87 & 4.8750 & 4.8750 & 0 \\
Absolute & leader rival & 110 & 4.7906 & 4.7906 & 0 \\
Absolute & leader rival round2 & 108 & 5.0250 & 2.5125 & 0 \\
Absolute & optimistic leader & 110 & 4.7906 & 4.7906 & 0 \\
Absolute & random & 5 & 5.9469 & 5.9469 & 0 \\
Absolute & topk & 103 & 4.8594 & 4.8594 & 0 \\
Anchored-study absolute & batch2 & 109 & 4.9281 & 3.3094 & 0 \\
Anchored-study absolute & leader rival & 110 & 4.7906 & 4.7906 & 0 \\
Anchored-study absolute & topk & 103 & 4.8594 & 4.8594 & 0 \\
Relative & batch2 & 113 & 4.9062 & 3.2906 & 0 \\
Relative & leader rival & 116 & 4.7313 & 4.7313 & 0 \\
Relative & topk & 110 & 4.7906 & 4.7906 & 0 \\
32-world cross-fit & all (infinite radius) & 0 & 0 & 0 & N/A \\
\hline
\end{tabular}
\end{table*}

\subsection{Scale-normalized reference verification}
\label{app:normalization}
\label{sec:s-integrated-reference}
The normalized-reference replay uses 40 worlds, 320 states and 15,477 candidates. The scale is $s(x)=\max\{1,\max_j|z_j^{\mathrm S}(x)|\}$ from raw Student outputs before geometry additions. The charged-anchor construction uses queried reference scores and radius $q\,s(x)$; the anchor consumes a query. This fixed-score replay is untimed. Repeated rules, budgets and folds reuse the same states.

Table~\ref{tab:s-normalized} gives all supported bound and budget combinations.

\begin{table}[tp]\centering\small
\caption{Complete supported reference-bound replay. All counts use 320 states; query means include abstentions. No submitted reference-action disagreement is observed in these nine cells.}
\label{tab:s-normalized}
\begin{tabular}{lrrr}\toprule
Bound & Cap & Certified & Queries \\\midrule
Absolute & 6 & 110 & 4.7906 \\
Relative & 6 & 116 & 4.7313 \\
Normalized & 6 & 175 & 4.0906 \\
Absolute & 12 & 118 & 8.6281 \\
Relative & 12 & 124 & 8.5000 \\
Normalized & 12 & 220 & 6.3344 \\
Absolute & 24 & 132 & 15.9875 \\
Relative & 24 & 137 & 15.6531 \\
Normalized & 24 & 250 & 9.5281 \\\bottomrule
\end{tabular}\end{table}

Each replay world is excluded from its 39-world calibration set. The one-sided rank interpretation assumes a fixed scorer, scale and aggregation rule and exchangeable world scores. Folds overlap and use fixed pretrained scores; backbone training-world membership is unknown. Reported intervals describe paired differences on this bank. Query traces record witnesses, rival bounds, anchor charges, rankings and final decisions.

At budget six, normalization gains 66 certified states and loses seven relative to unit-scale anchor-relative bounds. A paired world bootstrap uses 40 world means, seed 10061 and 2,000 draws with folds and radii fixed. The availability gain is 18.4375 points [12.1875,24.6875]; the query difference is $-0.640625$ [$-0.837578$,$-0.431250$]. The intervals condition on the fixed fits and sampled bank.

For $n$ calibration worlds the rank is $\lceil(n+1)(1-\alpha)\rceil$. If it exceeds $n$, the radius is infinite and the rule abstains before querying. With $n=32$ and $\alpha=0.025$, the required rank is 33; finite-bound coverage is undefined. The associated all-abstain results appear in Section~\ref{sec:s-stored-reference-queries}. Query work, certificate availability, conditional reference agreement and physical utility have distinct denominators.

\section{Three-dimensional control and physical utility}
\label{app:theme-control}
\subsection{Four-family control and paired costs}
\label{app:cross-family}
\label{sec:s-fourfamilies}
The four-family cohort contains 40 generated worlds, four policies and six starts per world, giving 960 planned executions. Each ten-world family uses layout perturbations, longitudinal/side obstacles, staggered transverse barriers, or long longitudinal/oblique barriers. Initial heights are 6/9/12\,m with two headings. A fixed-height horizontal breadth-first-search descriptor assigns 30 hard and ten very-hard worlds; this is a geometric descriptor, not a measured equality of control difficulty.

The initial three-family cohort has 720 executions. The final cohort retains 26 geometries and includes four regenerated geometries: layout-perturbation world 07, longitudinal/side worlds 07 and 08, and transverse-barrier world 07. The archived protocol records the final difficulty quotas. The complete selection history is unavailable. The transverse-barrier arms completed 56--58 of 60 tasks, motivating the long/oblique-barrier extension after inspection of those results. Pooled results include the transverse-barrier family. Two intermediate layout-development families and interrupted preliminary jobs are outside the declared cohort. The extension uses static geometry acceptance according to the protocol, reuses the simulator/task assets, and is reported as a descriptive evaluation designed after the initial results.

Initial LOS is unobstructed in the transverse-barrier family; target-pixel inclusion in the first camera frame is a separate camera-geometry condition.

Policies were executed in blocks ordered D6, G, F and $S_{\mathrm G}$ on matched worlds, altitudes, headings and seeds. Session and load drift were not controlled by randomized interleaving. The WM, proposer and dual-output Student are the same as in the CPU study. The probability map, wider reference interval and relaxed Guard are evaluated separately from these four deployed arms. The six-policy comparison and aligned-Student extension are reported in Section~\ref{sec:s-student-aligned}.

Table~\ref{tab:s-fourfamily-success} gives completion by family and collection stage.

\begin{table*}[tp]
\centering\small
\caption{Four-family completion uses all planned runs; the initial three-family cohort and the later extension are shown separately.}
\label{tab:s-fourfamily-success}
\begin{tabular}{lrrrr}
\toprule Population & D6 & Full F & Aligned Student $S_{\mathrm G}$ & Geometry G \\
\midrule
Layout perturbations & 42/60 & 43/60 & 42/60 & 34/60 \\
Longitudinal/side & 37/60 & 30/60 & 31/60 & 26/60 \\
Transverse barriers & 57/60 & 57/60 & 58/60 & 56/60 \\
Long/oblique extension & 45/60 & 46/60 & 42/60 & 42/60 \\
Initial three families & 136/180 & 130/180 & 131/180 & 116/180 \\
All four families & 181/240 & 176/240 & 173/240 & 158/240 \\
\bottomrule
\end{tabular}
\end{table*}

The 960-run denominator includes 35 initialization failures: D6/F/Student/G have 10/8/8/9, respectively; the other failures are timeouts. Excluding each arm's initialization failures gives 181/230, 176/232, 173/232 and 158/231. These arm-specific sensitivity denominators differ from the 229 starts valid for all four arms and the 107 on which all four succeed.

Thirty thousand paired world draws within family (seed 541103) yield D6-minus-F/Student/G completion differences $+2.08$ [$-3.33$,7.50], $+3.33$ [$-1.67$,8.75] and $+9.58$ [2.08,16.67] percentage points. Corresponding rescue/harm counts are 28/23, 34/26 and 46/23. These descriptive contrasts are unadjusted for multiple comparisons; the F and Student intervals span zero.

\begin{table*}[tp]
\centering\small
\caption{D6 versus aligned Student on matched conditional populations. Time is simulated mission time, energy simulated battery difference. Intervals are D6 minus Student in original units.}
\label{tab:s-fourfamily-cost}
\begin{tabular}{lrrrrr}
\toprule Population & Starts/worlds & D6/Student time (s) & Time difference 95\% & D6/Student energy (J) & Energy difference 95\% \\
\midrule
Joint success & 147/39 & 45.34/50.08 & [$-6.96$,$-2.62$] & 3,169.3/3,507.2 & [$-492.3$,$-189.9$] \\
Common valid & 229/40 & 68.53/74.02 & [$-8.66$,$-2.45$] & 4,790.2/5,179.6 & [$-611.0$,$-176.7$] \\
\bottomrule
\end{tabular}
\end{table*}

Joint-success costs condition on the outcome. Their bootstrap resamples worlds with at least one eligible pair, giving 39 worlds for the Student comparison. Resampling all 40 planned worlds instead gives time/energy intervals [$-6.98$,$-2.63$]\,s and [$-494.1$,$-190.4$]\,J. Common-valid summaries include 229 matched starts and timeouts; initialization failures lack mission telemetry. Time and battery difference are simulator outputs, with correlation approximately 0.99999 across 925 valid arm records. Table~\ref{tab:s-fourfamily-cost} therefore reports two related simulated costs on explicitly conditional populations.

Against F, the 153 joint-success starts average 45.79/49.24\,s and 3,201.2/3,444.9\,J for D6/F. The time interval is [$-6.59$, $-0.30$]\,s; the energy interval is [$-463.4$, $-23.9$]\,J. On 230 common-valid starts the time/energy intervals include zero. Against G, 135 joint-success starts favor D6, while timeout-inclusive time uncertainty includes zero. Paths do not uniformly decrease: D6's common-valid mean is 22.67\,m versus Student's 22.05\,m, with a difference interval spanning zero. Observations decrease from 12.61 to 11.89 on the 229-pair population.

D6 uses 14,453 reference-candidate evaluations over 5,309 decisions, versus F's 176,451 over 5,525. D6's 14,453 backend calls exceed F's 3,944 batched calls. The Student's 177,407 score evaluations use its own backend. The available initial-cohort state ledger records at most six D6 queries per decision; the long/oblique extension's full decision/array streams are externally indexed (Section~\ref{sec:s-current-availability}).

\subsection{Three-dimensional A/B/C/D control on 40 worlds}
\subsubsection{Worlds, policies, and discovery}
ProjectAirSim and Unreal supply the simulated views. The A/B/C/D cohort has 40 worlds comprising 15 medium, 20 hard, and five very-hard layouts. Each has six starts: 6/9/12\,m altitude and two headings separated by $15^\circ$, with a $30^\circ$ off-nadir camera and 40-motion cap. A 1.5\,m shifted synthetic anchor and declared static geometry supply context. The endpoint evaluator requires at least 35 target-visible pixels from the current rendered camera. The target's offline truth is not a planning input.

A is NoWorld; B uses the structured WM without Guard; C scans all reference candidates with Guard; D uses the student, selective queries, Guard, and progress handling. The reference/student have 24,938/20,777 parameters. The 56 input features require known-obstacle line of sight (LOS). Seven perturbation rows are evaluated per new logical query; yaw perturbations duplicate the center in the frozen encoding. Table~\ref{tab:s-main-success} reports all task outcomes and difficulty strata. Table~\ref{tab:s-main-pairs} reports batchwise paired effects; Table~\ref{tab:s-main-failures} reports every failure type.
\begin{table*}[tp]
\centering\small
\caption{Primary task success and difficulty slices, including all planned starts. The separate strict comparison uses the same 40 worlds. The aligned-score control is a later 60-run comparison on reused worlds.}
\label{tab:s-main-success}
\begin{tabular}{lrrrrrr}
\toprule
Population & Worlds & Tasks/arm & A & B & C & D \\
\midrule
First batch & 20 & 120 & 93 & 34 & 105 & 107 \\
Second batch & 20 & 120 & 73 & 26 & 97 & 93 \\
Pooled descriptive & 40 & 240 & 166 & 60 & 202 & 200 \\
Medium & 15 & 90 & 87 & 40 & 87 & 87 \\
Hard & 20 & 120 & 74 & 11 & 93 & 94 \\
Very hard & 5 & 30 & 5 & 9 & 22 & 19 \\
\midrule
Strict policies & 40 & 240 & \multicolumn{4}{l}{SC: 199; SA: 202} \\
Aligned-score control & 10 reused & 60 & \multicolumn{4}{l}{C2: 50; historical A/C/D: 35/47/49} \\
\bottomrule
\end{tabular}
\end{table*}
\begin{table*}[tp]
\centering\small
\caption{Paired success differences in percentage points. Pointwise intervals resample whole worlds. World sign-flip tests are exploratory and unadjusted.}
\label{tab:s-main-pairs}
\begin{tabular}{llrrr}
\toprule
Population & Contrast & Difference & 95\% interval & World sign-flip $p$ \\
\midrule
First batch & C--A & 10.00 & [3.33,18.33] & 0.0625 \\
First batch & D--A & 11.67 & [3.33,22.50] & 0.0625 \\
First batch & D--C & 1.67 & [$-1.67$,5.00] & 0.6250 \\
Second batch & C--A & 20.00 & [6.67,34.17] & 0.0156 \\
Second batch & D--A & 16.67 & [4.17,30.83] & 0.0391 \\
Second batch & D--C & $-3.33$ & [$-9.17$,1.67] & 0.4063 \\
Pooled & C--A & 15.00 & [7.08,23.33] & 0.000854 \\
Pooled & D--A & 14.17 & [6.25,22.92] & 0.001953 \\
Pooled & D--C & $-0.83$ & [$-4.17$,2.08] & 0.8086 \\
\bottomrule
\end{tabular}
\end{table*}
\begin{table*}[tp]
\centering\small
\caption{Complete outcomes for the original 960 executions. Initialization errors and policy exceptions remain in the primary task denominator.}
\label{tab:s-main-failures}
\begin{tabular}{lrrrrr}
\toprule
Arm & Success & Timeout & Collision & Initialization error & Policy error \\
\midrule
A & 166 & 68 & 0 & 6 & 0 \\
B & 60 & 118 & 19 & 7 & 36 \\
C & 202 & 31 & 0 & 7 & 0 \\
D & 200 & 34 & 0 & 6 & 0 \\
\bottomrule
\end{tabular}
\end{table*}

C versus A has 162 common successes, 40 rescues, four reverse degradations, and 34 common failures. D versus A has 159, 41, seven, and 33. D versus C has 186, 14, 16, and 24. On the five very-hard worlds, D has 19/30 successes versus C's 22/30, with three C-relative rescues and six reverse degradations. This difficulty slice complements the complete-population gain.

On 233 common C/D start-valid tasks, mean wall time is 62.28 and 60.68\,s and simulated battery energy 3,295.89 and 3,194.91\,J. Mean observations are 8.412 and 8.554 and path lengths 16.426 and 16.600\,m. Equal-world intervals for wall-time and simulated-energy changes span zero. The original trajectories evaluate 174,622 and 8,993 logical WM candidates; their decision costs also differ in score and feature construction. Matched costs in Section~\ref{sec:s-matched-access} isolate these factors.

\subsection{First-layout six-policy cohort}
\label{sec:s-first-layout}
Twenty procedural layouts (15 hard, five very hard), six starts per layout and six policies give 720 executions. The strict and fault studies reuse these layouts, generator family and frozen models. They test variation within one procedural generator; planned denominators include initialization failures.

Tables~\ref{tab:s-new-six} and~\ref{tab:s-new-pairs} report complete outcomes, work and paired effects. Pointwise intervals use 20,000 whole-world resamples, retaining all starts and paired policy outcomes; they are not multiplicity-adjusted.

\begin{table*}[tp]\centering\small
\caption{Complete frozen six-policy outcomes. A: NoWorld; G: LOS geometry; S: student only; F: full reference; T6: fixed shortlist; D6: selective.}
\label{tab:s-new-six}

\begin{tabular}{lrrrrr}
\toprule

Policy & Success & Rate (\%) & Timeout & Initialization & Queries \\
\midrule

A & 67/120 & 55.83 & 51 & 2 & 0 \\

G & 76/120 & 63.33 & 42 & 2 & 0 \\

S & 92/120 & 76.67 & 25 & 3 & 0 \\

F & 90/120 & 75.00 & 28 & 2 & 65,599 \\

T6 & 90/120 & 75.00 & 27 & 3 & 8,754 \\

D6 & 97/120 & 80.83 & 21 & 2 & 5,739 \\

\bottomrule
\end{tabular}
\end{table*}

\begin{table*}[tp]\centering\small
\caption{Paired completion effects on the same 120 starts, clustered by 20 worlds.}
\label{tab:s-new-pairs}

\begin{tabular}{lrrrr}
\toprule

Contrast & Rescues & Harms & Difference (pp) & 95\% interval \\
\midrule

D6 vs A & 35 & 5 & +25.00 & [11.67, 40.00] \\

D6 vs G & 26 & 5 & +17.50 & [7.50, 28.33] \\

D6 vs S & 14 & 9 & +4.17 & [-5.83, 14.17] \\

D6 vs F & 11 & 4 & +5.83 & [0.00, 12.50] \\

D6 vs T6 & 12 & 5 & +5.83 & [-1.67, 13.33] \\

\bottomrule
\end{tabular}
\end{table*}

D6 averages 3.912 logical queries over 1,467 entered decisions; 56.92\% hit six. Task-total work falls 91.25\% versus full scanning and 34.44\% versus Top-6 on policy-dependent visited states. Returned-decision and logged-stage timings have different boundaries; common-input timing is reported in Section~\ref{sec:s-matched-access}. On 117 common start-valid D6/G pairs, mean paths are 21.54/19.84\,m and observations 10.78/9.84: fewer motions can coexist with longer paths and more sensing. Success rates use all planned runs.

\subsection{Cross-layout control and paired effects}
\label{app:validation}
\label{sec:s-cross-layout}
Two procedural layout regimes share simulation assets but change obstacle/layout generation rules. Each has 20 worlds, six starts per world and six policies. A 40-run full-policy engineering pilot reuses second-regime maps and YAML definitions; it is excluded from that regime's 720 formal executions. These cohorts test procedural-layout transfer within the simulator.
Success denominators include all planned starts, including 31 pre-decision initialization failures in the second regime and 45 across both formal cohorts. Failures are not rerun or replaced. Resource comparisons use common-valid telemetry without imputing missing mission time or battery values.

\begin{table*}[tp]
\centering\small
\caption{Complete six-policy outcomes and reference work. Regime 1/2 are separate 20-world cohorts. Queries count logical candidates; backend calls retain batching.}
\label{tab:regime-complete}
\begin{tabular}{lrrrrrr}
\toprule
Arm & Regime 1 & Regime 2 & Combined & Init. errors & Queries & Backend calls \\
\midrule
A & 67/120 & 46/120 & 113/240 & 7 & 0 & 0 \\
G & 76/120 & 46/120 & 122/240 & 7 & 0 & 0 \\
S & 92/120 & 80/120 & 172/240 & 10 & 0 & 0 \\
F & 90/120 & 77/120 & 167/240 & 7 & 185,418 & 4,206 \\
T6 & 90/120 & 77/120 & 167/240 & 7 & 25,098 & 4,183 \\
D6 & 97/120 & 76/120 & 173/240 & 7 & 16,158 & 16,158 \\
\bottomrule
\end{tabular}
\end{table*}

\begin{table*}[tp]
\centering\small
\caption{Paired D6 task effects. Intervals use 100,000 whole-world bootstrap draws, stratified by regime, seed 20261026. They are descriptive after pilot and historical-data exposure.}
\label{tab:paired-complete}
\begin{tabular}{lrrrr}
\toprule
Contrast & Rescues & Harms & Gain(pp) & 95\% interval \\
\midrule
D6-A & 76 & 16 & +25.00 & [17.08,33.33] \\
D6-G & 66 & 15 & +21.25 & [13.75,28.75] \\
D6-S & 38 & 37 & +0.42 & [-6.25,7.08] \\
D6-F & 29 & 23 & +2.50 & [-2.08,7.08] \\
D6-T6 & 29 & 23 & +2.50 & [-2.92,7.92] \\
\bottomrule
\end{tabular}
\end{table*}

D6 exceeds S by one success in 240 planned starts (Tables~\ref{tab:regime-complete} and~\ref{tab:paired-complete}). Resource comparisons use separate common-valid and success-only denominators. On the 60 very-hard starts, D6/S complete 38/31 tasks; this is an exploratory difficulty slice.

\subsection{Controller-aligned Student comparison}
\label{app:followup}
\label{sec:s-student-aligned}
The six-policy comparison has 1,440 executions in two 20-world regimes. Evaluating an aligned Student on the same 240 world/start keys gives 1,680 records (Table~\ref{tab:s-seven-policy}). The aligned $S_{\mathrm{G}}$ has 20,810 parameters: the original 20,777 parameters and regression-score tensors are unchanged, and 33 parameters add a probability-logit row. Both regimes use the same checkpoint. Fitting and selection use the original train/development folds; the added probability head has no independent physical calibration.

The aligned backend uses the Student with the full-reference controller's seven-row aggregation, Guard, progress replacement, fallback and sensing triggers. Its 183,389 seven-point candidate evaluations use the Student backend. The log field \texttt{wm} counts these Student evaluations; the arm makes zero actual WM queries. The 240 terminal outcomes include nine initialization errors; the other arms have 45, giving 54 across 1,680 records.

\begin{table*}[tp]\centering\small
\caption{Seven-policy control population. The follow-up $S_{\mathrm{G}}$ reuses the same world/start keys; it is not an additional 240 independent environments.}
\label{tab:s-seven-policy}
\begin{tabular}{lrrrrrr}\toprule
Arm & Layout I & Layout II & All & Timeout & Init. error & Actual WM queries\\\midrule
A &67/120&46/120&113/240&120&7&0\\
G &76/120&46/120&122/240&111&7&0\\
S &92/120&80/120&172/240&58&10&0\\
$S_{\mathrm{G}}$ &91/120&61/120&152/240&79&9&0\\
F &90/120&77/120&167/240&66&7&185,418\\
T6 &90/120&77/120&167/240&66&7&25,098\\
D6 &97/120&76/120&173/240&60&7&16,158\\\bottomrule
\end{tabular}\end{table*}

Thirty thousand world-cluster resamples (seed 103030), stratified by regime and retaining all six starts, give D6 minus $S_{\mathrm{G}}$ $+8.75$ percentage points [2.08,15.42]. The pair table has 126 joint successes, 47 D6-only, 26 Student-only and 41 joint failures. The 231 common-initialization-valid pairs give $+8.66$ points [1.75,15.52]. Layout-specific gains are 5.00 and 12.50 points. Ten very-hard worlds give 38/60 versus 30/60, a descriptive 13.33-point difference [$-5.00$,30.00]. The original S completes 172/240; D6 minus S is 0.42 points [$-6.25$,7.08]. These follow-up comparisons are not a prespecified family-wise superiority test.

On the 6,590 $S_{\mathrm{G}}$-visited states, replay before common progress handling admits 1,945 suggestions under the original Student rules and 1,463 under the aligned rules. All 482 additional rejections arise from the probability threshold; center/seven-row group leaders match on all 4,199 model-entered inputs. This state replay identifies the gate difference but cannot determine counterfactual task outcomes. The aligned rule uses the WM threshold without a separate Student operating-point calibration. Task completions are 172/240 for S, 152/240 for $S_{\mathrm{G}}$ and 173/240 for D6.

The 231 common-valid pairs have cumulative decision times 1.996/2.007\,s, simulated mission times 80.517/82.467\,s, simulated battery values 5,624.53/5,767.20\,J and 13.957/14.580 observations for D6/$S_{\mathrm{G}}$. These are exploratory paired estimates; joint-success summaries condition on the outcome. The aligned Student runs were collected at a different time from the six-policy runs, leaving session drift uncontrolled despite matched seeds.

\subsection{Seven-policy control comparison}
\label{sec:s-sevenpolicy-figure}
\begin{figure*}[tp]
\centering\includegraphics[width=.98\textwidth]{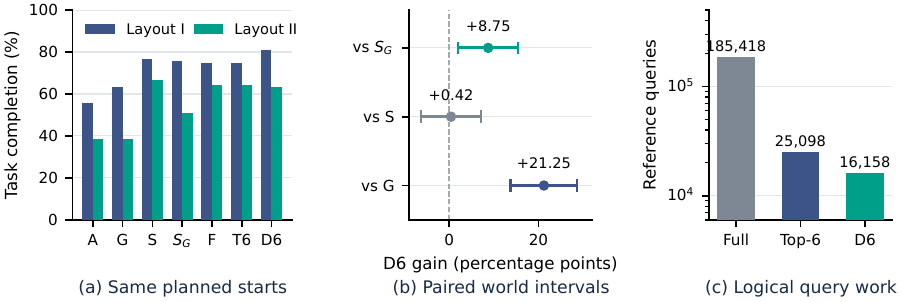}
\caption{Seven-policy control on two layout regimes. Student S completes 172/240 tasks, D6 173/240 and aligned Student 152/240. Layout-specific results, paired intervals and logical reference work use this cohort, separately from the four-family evaluation.}
\Description{Seven policies share 240 starts across two layout regimes. Both Student variants, paired world intervals and logical reference work are shown.}
\label{fig:s-historical-seven}
\end{figure*}

Figure~\ref{fig:s-historical-seven} shows the 1,680 executions, both Student variants and paired world intervals described in Section~\ref{sec:s-student-aligned}.

\subsection{Repeated executions on eight reused worlds}
\label{sec:s-execution-repeats}
Two repeated A/D rounds over eight reused worlds yield 51/96 and 72/96 successes. Including the original corresponding round gives 78/144 and 107/144. The new-round difference is 21.88 points [3.13,43.75]. These are repeated executions, not 144 independent maps. Two D executions diverge in action in 25/48 conditions versus 14/48 for A, illustrating simulator/execution variability alongside average benefit.

\subsection{Aligned six-policy budgets on ten reused worlds}
\label{sec:s-aligned-budget}
The ten reused difficult worlds contribute 60 matched conditions per policy. F, T6, and D2/D6/D12 share the student, WM, interval files, composite ranking, Guard, progress, and sensing rules. F uses the common low-value gate then batches every seven-row candidate query. T6 batches the initial point-score Top-6. D follows sequential upper-bound-priority queries and can stop early at budgets two, six, and twelve. Original C uses different ranking and is not substituted for aligned F.

Table~\ref{tab:s-aligned} reports all 360 executions; Table~\ref{tab:s-aligned-pairs} reports whole-world paired effects. D6 uses 3,295 queries versus F's 40,617 and T6's 6,054, reductions of 91.89\% and 45.57\%. Similar success totals do not establish noninferiority. D2 and F each succeed 43 times, but each rescues six failures of the other. D12 and F each rescue eight of the other's failures. Budget effects are consequently not monotonic across tasks.
\begin{table*}[tp]
\centering\small
\caption{Aligned policies on all 60 planned conditions each. Work accumulates over the states visited by each policy. Entered-decision means exclude prequery gate rejection.}
\label{tab:s-aligned}
\begin{tabular}{lrrrrrr}
\toprule
Policy & Success & Hard/36 & Very hard/24 & Decisions & Total queries & Entered mean \\
\midrule
A & 24/60 & 23 & 1 & 1,710 & 0 & -- \\
F & 43/60 & 30 & 13 & 1,464 & 40,617 & 45.231 \\
T6 & 36/60 & 26 & 10 & 1,576 & 6,054 & 6.000 \\
D2 & 43/60 & 29 & 14 & 1,480 & 1,479 & 1.585 \\
D6 & 42/60 & 26 & 16 & 1,482 & 3,295 & 3.719 \\
D12 & 43/60 & 27 & 16 & 1,501 & 6,609 & 7.114 \\
\bottomrule
\end{tabular}
\end{table*}
\begin{table*}[tp]
\centering\small
\caption{Exploratory paired success effects. Intervals resample ten whole-world clusters. Rescue and harm counts include every planned condition; no prospective noninferiority margin is tested.}
\label{tab:s-aligned-pairs}
\begin{tabular}{lrrrr}
\toprule
Contrast & Rescues & Harms & Difference (pp) & 95\% interval (pp) \\
\midrule
F--A & 23 & 4 & 31.67 & [11.67,53.33] \\
T6--A & 16 & 4 & 20.00 & [3.29,36.67] \\
D2--A & 22 & 3 & 31.67 & [11.67,51.67] \\
D6--A & 21 & 3 & 30.00 & [8.33,51.67] \\
D12--A & 23 & 4 & 31.67 & [5.00,56.67] \\
D6--T6 & 9 & 3 & 10.00 & [$-3.33$,25.00] \\
D6--F & 6 & 7 & $-1.67$ & [$-15.00$,13.33] \\
\bottomrule
\end{tabular}
\end{table*}
\begin{table*}[tp]
\centering\small
\caption{Resource means on the secondary 58-condition common-start-valid population. Wall time covers the mission. Battery energy is simulated telemetry, not measured processor energy.}
\label{tab:s-aligned-resources}
\begin{tabular}{lrrrrrr}
\toprule
 & A & F & T6 & D2 & D6 & D12 \\
\midrule
Task wall time (s) & 103.89 & 97.61 & 100.87 & 96.56 & 97.42 & 99.32 \\
Simulated battery (J) & 5,585.04 & 5,149.10 & 5,363.50 & 5,131.83 & 5,086.20 & 5,210.81 \\
\bottomrule
\end{tabular}
\end{table*}

The 360 executions contain 231 successes, 122 timeouts, and seven zero-action initialization errors. Those errors affect two starts in one world; removing both starts from every arm gives the 58-condition resource sensitivity in Table~\ref{tab:s-aligned-resources}. Success uses all 60 planned conditions. D6--F wall-time and simulated-energy intervals include zero.

The implementation follows a fixed initial upper-bound order and adapts only its stopping point: unqueried bounds remain fixed, and all 2,748 entered D decisions follow a prefix of that order. Entered-decision cap-hit rates are 58.52/53.05/54.90\% for D2/D6/D12. D2 has 387 pre-cap stops and 27 decisions resolved at query two. D6 has 416 interval/Guard stops, 38 overridden by progress handling, and 470 unresolved exhausted attempts that fall back. Query totals include gate rejection, resolved decisions and unresolved fallback.

\subsection{Rendered-view fidelity and physical-score intervals}
\label{sec:s-rendered-physical}
The rendered collection has 480 calibration states from 80 worlds, 240 fixed test states from 40 new worlds, and 39 selected visited states. One planned visited snapshot is not reached because its task has already succeeded; it is not replaced. Table~\ref{tab:s-rendered} gives the complete 759-state, 37,453-candidate population. Pixel labels come from target segmentation at counterfactual Unreal endpoints, distinct from static-geometry visibility proxies.
\begin{table*}[tp]
\centering\small
\caption{Independent rendered collection. A positive state has a candidate reaching the 35-pixel discovery threshold; subthreshold states have positive pixels below that threshold.}
\label{tab:s-rendered}
\begin{tabular}{lrrrrrr}
\toprule
Split & Worlds & States & Candidates & Pixel-positive & All zero & Subthreshold \\
\midrule
Calibration & 80 & 480 & 23,588 & 139 & 324 & 17 \\
Fixed test & 40 & 240 & 11,884 & 66 & 164 & 10 \\
Visited & 39 & 39 & 1,981 & 6 & 32 & 1 \\
\bottomrule
\end{tabular}
\end{table*}

For each calibration world, the maximum residual spans all six states and their candidates. Student-to-WM and WM-to-rendered $\log(1+\mathrm{pixels})$ widths are $q_1=2.3873806$ and $q_2=6.3777392$, with 80 worlds and per-layer $\alpha=0.0125$ (rank 80). Their sum is 8.7651198. A WM query leaves the physical width $q_2$. These are diagnostic widths; executed policies use their frozen reference bands.

\begin{figure*}[tp]
\centering\includegraphics[width=.97\textwidth]{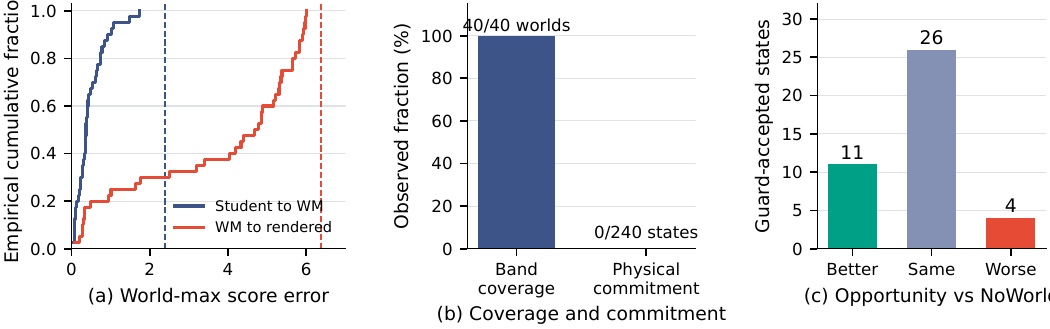}
\caption{Rendered-view diagnostics. (a) Test world-max errors versus calibration maxima. (b) Observed coverage and strict physical commitment have different units. (c) Rendered two-step opportunity for 41 engineering-accepted states versus the same-state NoWorld action.}
\label{fig:s-rendered}
\Description{Rendered-view diagnostics. (a) Test world-max errors versus calibration maxima. (b) Observed coverage and strict physical commitment have different units. (c) Rendered two-step opportunity for 41 engineering-accepted states versus the same-state NoWorld action.}
\end{figure*}

Both layers jointly cover 40/40 test worlds, with an exact two-sided 95\% interval of [91.19,100]\% and a one-sided lower bound of 92.78\%; the sample lower bound is below 97.5\%. At six queries, neither physical-score certificate commits on any test state (0/240 for both candidate and action certificates). Figure~\ref{fig:s-rendered} separates coverage from commitment and shows 11 better, 26 equal, and four worse rendered opportunities among 41 engineering-accepted states.
\begin{table*}[tp]
\centering\small
\caption{Acceptance-conditioned diagnostic widths fitted on calibration worlds. Selected-candidate-only residuals do not certify unselected competitors.}
\label{tab:s-conditioned}
\begin{tabular}{lrr}
\toprule
Calibration scope & $q_1$ & $q_2$ \\
\midrule
All states / all candidates & 2.3874 & 6.3777 \\
D Guard accepted / all candidates & 1.8825 & 5.8382 \\
Full-WM Guard accepted / all candidates & 1.5046 & 5.8382 \\
D accepted / selected candidate only & 1.0131 & 5.7313 \\
\bottomrule
\end{tabular}
\end{table*}

Student/WM first-action sets intersect on 215/240 fixed test states and all 41 engineering-accepted states. WM/rendered mean absolute error is 0.3426 over all test candidates, 0.5229 over accepted-state candidates, and 3.0148 for the selected candidate in accepted states. Fourteen accepted states have no visible candidate. Only 43 calibration worlds contain selective-accepted states and 41 contain full-accepted states. Table~\ref{tab:s-conditioned} treats empty worlds as zero maxima; dropping them would make the requested ranks unsupported. Acceptance-conditioned state-level guarantees do not follow from the unconditional world calculation.

\subsection{Nominal rendered endpoints and physical diagnostics}
\label{sec:s-physical-endpoints}
Nominal-endpoint rendering collects 236/240 planned slots on the same 20 worlds, with 10,835 candidate labels. Two source initialization failures leave four slots missing and unreplaced. There are 176 all-zero states, 52 with a candidate reaching 35 visible pixels, and eight with nonzero subthreshold visibility; 186 candidates reach 35 pixels. Table~\ref{tab:s-new-render} separates reference approximation from physical prediction.

\begin{table*}[tp]\centering\small
\caption{Nominal-endpoint diagnostics on 236 states from the 20-world layout cohort, using the deployed calibration half-widths.}
\label{tab:s-new-render}

\begin{tabular}{lr}
\toprule

Measure & Value \\
\midrule

Student--WM mean absolute error & 0.02137 \\

WM--rendered log-pixel mean absolute error & 0.35586 \\

Probability-head Brier / absent baseline & 0.02972 / 0.01717 \\

Probability-head area under ROC / average precision & 0.4691 / 0.01791 \\

Probability threshold 0.25: TP / FP / FN / TN & 13 / 378 / 173 / 10,271 \\

35-pixel prediction: TP / FP / FN / TN & 3 / 166 / 183 / 10,483 \\

D6 agreement at F-accepted first actions & 35 / 36 \\

D6 two-step visibility improves / ties / harms NoWorld (A) & 4 / 231 / 1 \\

\bottomrule
\end{tabular}
\end{table*}

Threshold diagnostics use correlated candidates. Local opportunity measures the best continuation inside a selected first-action group, separately from immediate visibility and mission completion. World-max physical certificates are evaluated in Section~\ref{sec:s-rendered-physical}; raw-WM rankings and executed endpoints are analyzed in Section~\ref{sec:s-physical-ranking}.

\subsection{Physical reference diagnostics and sparse visibility}
\label{sec:s-physical-ranking}
For the 236-snapshot/10,835-candidate population in Table~\ref{tab:s-new-render}, raw-WM physical-optimum top-1/top-3/top-6 counts are 0/4/6 among the 60 non-all-zero snapshots. This ranks raw candidate predictions separately from composite guarded control and executed first motion.

The raw-score threshold diagnostic in Table~\ref{tab:physical-threshold} separates log-pixel predictions from Guard probabilities.

\begin{table}[tp]
\centering\small
\caption{Raw WM log-pixel threshold diagnostic: predicted positive when $z\geq\log(36)$, rendered positive at 35 pixels. It is not the probability-head or controller Guard confusion matrix.}
\label{tab:physical-threshold}
\begin{tabular}{lr}
\toprule
Quantity & Count \\
\midrule
True positive & 3 \\
False positive & 166 \\
False negative & 183 \\
True negative & 10483 \\
Predicted positive, actual zero & 150 \\
Actual-zero FP in all-zero states & 123 \\
\bottomrule
\end{tabular}
\end{table}

Candidate-weighted student-to-WM absolute error is 0.021374 and WM-to-rendered log-pixel error is 0.355856. The probability head has a higher Brier score than constant absence (0.02972 versus 0.01717), separately from composite ranking. The original endpoint controls compare D6 with NoWorld (A): each selects a 35-pixel-visible endpoint in 11/236 states. On the 52 qualifying states, D6/NoWorld end within 2\,m of a best-view pose in 35/33 cases, with a world-cluster distance-gain interval crossing zero. Each reaches 90\% of the best pixels after its nominal first action in only 1/52 cases. Pose proximity and best-continuation opportunity do not establish first-action visibility equivalence. Table~\ref{tab:s-new-render} retains the D6--NoWorld two-step visibility counts; Section~\ref{sec:s-action-physical} compares action-group utility with both G and NoWorld.

\subsection{Development probability calibration}
\label{sec:s-dev-reference}
Two-parameter probability maps use 33 validation states and 1,248 declared candidates while preserving regression-score weights and seven-row ordering. Train/validation/calibration/test partitions have 149/33/32/51 state or cell identifiers; these are not independent world splits. Candidate-level fitting and evaluation input gaps are listed in Section~\ref{sec:s-current-availability}.

\begin{figure*}[tp]
\centering\includegraphics[width=.98\textwidth]{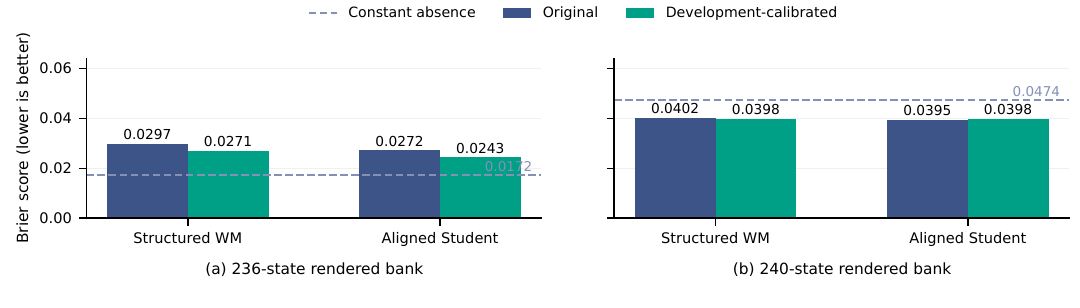}
\caption{Probability calibration across two rendered-state banks. Grouped bars compare original and development-calibrated WM and Student Brier scores for the qualifying-visibility event: (a) 236 states and (b) 240 states. Each panel includes its constant-absence baseline; lower Brier is better. The maps preserve regression-score ordering and are not deployed in the four-family control experiment.}
\Description{Panel a, 236 states: original/calibrated Brier is 0.029723/0.027123 for WM and 0.027180/0.024346 for Student, versus 0.017167 for constant absence. Panel b, 240 states: WM is 0.040181/0.039777 and Student 0.039485/0.039762, versus 0.047375 for constant absence. Student calibration increases Brier in the 240-state bank.}
\label{fig:s-physical-calibration}
\end{figure*}

Figure~\ref{fig:s-physical-calibration} compares the two rendered-state banks.

On the 236-state/10,835-candidate bank, the qualifying event has 186 positives. The map reduces WM Brier from 0.029723 to 0.027123, while constant absence gives 0.017167. At threshold 0.25, true positives fall from 13 to 11, recall from 6.99\% to 5.91\%, and false positives from 378 to 268. In the 240-state bank, WM Brier falls from 0.040181 to 0.039777 versus 0.047375 for constant absence, but mapped log loss increases. Student Brier decreases from 0.027180 to 0.024346 in the 236-state bank and increases from 0.039485 to 0.039762 in the 240-state bank. The control experiments use the original probability branches; action-ranking and horizon-specific results appear in Sections~\ref{sec:s-action-physical} and~\ref{sec:s-prediction-controls}.

A 40-cell development search over probability and margin thresholds selects zero/zero for WM, Student and D6. Each selects 19/33 qualifying endpoints with approximately 0.60 endpoint utility. This endpoint-proxy analysis uses incomplete final-progress history. The resulting relaxed gates and probability maps were not deployed in the four-family executions.

The strict factorial's reference-width replay (Section~\ref{sec:proposer-v42}) fixes the physical-error width and evaluates states offline, separately from closed-loop control.

\subsection{Action-group utility and invisible-state progress}
\label{sec:s-action-physical}
The bank and missing-slot accounting are in Section~\ref{sec:s-physical-endpoints}. The 60 nonzero states include all eight subthreshold states, whose maxima are 32,28,25,2,12,12,28,31 pixels. A first-action group's physical opportunity is the maximum rendered pixels among its existing continuations; all exact ties count as best. This evaluator's best-continuation value need not be attained by the executed first action.

Table~\ref{tab:s-action-physical} compares frozen selectors on states sampled from NoWorld trajectories. G is the four-family LOS geometry baseline, distinct from NoWorld (A). Its replay protocol scores saved LOS belief mass plus 0.01 times novel LOS area (m$^2$) minus 0.025 times candidate distance (m), selects the earliest horizon reaching 65\% of the best reachable value, and applies the common progress constraint: after three consecutive nonhorizontal actions, choose feasible horizontal movement, preferring the belief mode. Target coordinates and rendered pixels enter evaluation only. The comparison is conditional on the NoWorld collection policy and does not establish closed-loop physical gains for G or D6.

\begin{table*}[tp]\centering\small
\caption{Recorded group opportunity and nominal progress on fixed NoWorld-trajectory states. Best groups include tied maxima. Group columns use 52 qualifying or 60 nonzero states; distance and movement columns use all 176 all-zero states, with zero progress for nonmoving actions.}
\label{tab:s-action-physical}
\begin{tabular}{lrrrrrr}\toprule
Rule & Best group /52 & Group$\geq 35$/52 & Best group /60 & Distance-best /176 & Mean progress(m) & Moving states\\\midrule
LOS geometry (G) &14&24&15&27&0.124871&84\\
NoWorld (A) &19&30&20&28&0.142363&82\\
Full + Guard &21&32&22&39&0.235326&99\\
Top-6 &21&32&22&38&0.225961&98\\
D6 &21&32&22&38&0.218087&98\\
Raw WM &14&28&16&33&$-0.237370$&149\\\bottomrule
\end{tabular}\end{table*}

On the 52 qualifying states from 15 worlds, D6 selects a best first-action group (ties count) on 21 states versus 14 for G: seven D6-only selections and none G-only. The state-weighted difference is 13.46 percentage points; averaging within-world paired rates equally over the 15 worlds gives 10.00 points (95\% world-bootstrap interval [3.33,17.78]). Against NoWorld, D6 has three D6-only and one NoWorld-only best-group selections, with 48 agreements: 3.85 points by states and 3.56 points by worlds (world interval [$-4.44$,11.56]). F and T6 each match D6's best-group count. Raw WM selects zero-pixel endpoints on 43/60 nonzero states and 35-pixel-visible endpoints on eight; its exact-optimum top-1/top-3/top-6 counts are 0/4/6 (Section~\ref{sec:s-physical-ranking}). Applying progress mapping changes the selector.

For every all-zero state, the safe candidate pool permits positive nominal target-distance reduction. D6 minus G averages 0.093216\,m over states and 0.091900\,m with equal-world weighting (95\% interval [0.025630,0.160177]\,m). The D6--G best-group and all-zero progress intervals use 5,000 world resamples (seed 20261031). The 38 states on which only one moves account for 87.52\% of the summed progress difference: 26 D6-only and 12 G-only moves. On the 72 states where both move, the state mean difference is 0.028444\,m; the equal-world difference is 0.083879\,m, with interval [$-0.043417$,0.251120]\,m, spanning zero.

The original NoWorld progress control gives a D6-minus-NoWorld difference of 0.075724\,m by states and 0.064729\,m by worlds (5,000 world draws, seed 20261031; interval [0.020953, 0.115586]\,m). D6 is nearer/tied/farther on 12/155/9 states. Sixteen D6-only moves contribute 12.7688\,m of the summed 13.3275\,m gain (95.81\%). On the 82 states where both move, D6 and NoWorld means are 0.312372 and 0.305559\,m (difference 0.0068\,m), with a world interval spanning zero. In this NoWorld-control record, 43 of D6's 98 translations reduce and 55 increase Euclidean target distance; a single-step distance increase alone does not identify a navigation error. Both comparisons are exploratory nominal single-step outcomes on the shared bank; the both-moving intervals leave movement-direction gains unresolved.

The separate 240-state rendered bank has 164 all-zero, 66 qualifying and ten subthreshold states, with 11,884 candidate labels. Rescoring holds Guard choices, progress mappings and nominal primitive positions fixed; input availability is in Section~\ref{sec:s-current-availability}.

\subsection{Physical endpoints, first-action opportunities and local progress}
\label{sec:s-physical-action-summary}
\begin{figure*}[!t]
\centering\includegraphics[width=.98\textwidth]{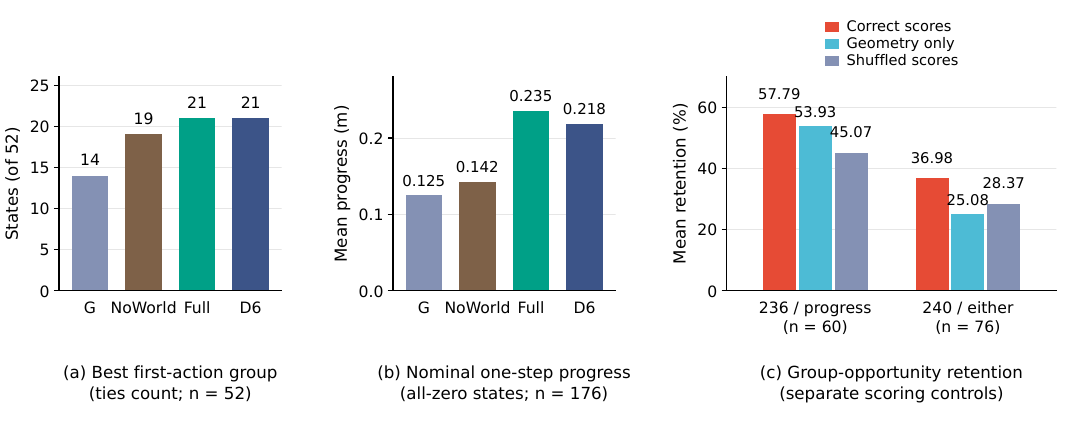}
\caption{\textbf{Physical opportunity and local progress.} (a) Best first-action groups (ties count) on 52 qualifying states. (b) Mean nominal one-step target-distance reduction over all 176 all-zero states, including turns and hover at zero progress. G, NoWorld, Full + Guard and D6 share the fixed NoWorld-trajectory state bank; these are conditional selector comparisons. (c) Correct composite, geometry-only and depth-stratified shuffled scores use the same progress mapping; bars average oracle-normalized group-opportunity retention over 60/76 nonzero states in the 236/240 banks. These scoring controls are distinct from G and frozen D6.}
\Description{Panels a and b use G, NoWorld, Full + Guard and D6 in that order: best-group counts 14,19,21,21 of 52, and mean nominal progress 0.124871,0.142363,0.235326,0.218087 metres over 176 all-zero states. Panel c shows correct/geometry-only/shuffled group-opportunity means 57.79/53.93/45.07 percent for 236 states with progress and 36.98/25.08/28.37 percent for the 240-state bank, as in Table~\ref{tab:s-prediction-horizons}.}
\label{fig:physical-actions}
\end{figure*}

At query budget six, strict physical bounds certify actions in 0/240 rendered test states. Physical-certificate availability, reference-action agreement, first-action visibility and group opportunity are distinct outcomes. Figure~\ref{fig:physical-actions} plots the state-weighted counts and means in Table~\ref{tab:s-action-physical} alongside the separate score-correspondence controls in Table~\ref{tab:s-prediction-horizons}.

\subsection{Prediction correspondence controls at three horizons}
\label{sec:s-prediction-controls}
The exploratory selectors use composite scores, geometry alone, or depth-stratified shuffled scores. There are 512 permutations per state (seed 103031); world uncertainty uses 5,000 draws (seed 103032). Geometry and the score distribution at each depth remain fixed. The 243,712 shuffled rankings reuse 476 states. Progress mapping is identity throughout the 240-state bank, so both wrapper analyses yield identical choices.

\begin{table*}[tp]\centering\small
\caption{Oracle-normalized utility retention at three horizons. Each percentage averages per-state utility ratios over nonzero-oracle states. These scoring controls are not the frozen D6 controller.}
\label{tab:s-prediction-horizons}
\begin{tabular}{llrrrr}\toprule
Bank / wrapper & Horizon & Nonzero states & Correct scores & Geometry only & Shuffled scores\\\midrule
236 / no progress & Endpoint &60&11.72&12.01&13.52\\
236 / no progress & First step &25&55.18&47.83&48.59\\
236 / no progress & Group opportunity &60&51.16&45.41&40.06\\
236 / progress & Endpoint &60&17.17&17.65&15.52\\
236 / progress & First step &25&47.83&43.83&44.28\\
236 / progress & Group opportunity &60&57.79&53.93&45.07\\
240 / either & Endpoint &76&14.55&12.58&10.12\\
240 / either & First step &27&35.22&56.86&38.83\\
240 / either & Group opportunity &76&36.98&25.08&28.37\\\bottomrule
\end{tabular}\end{table*}

Table~\ref{tab:s-prediction-horizons} reports state-weighted retention; the separate recorded world-bootstrap summaries depend on horizon. With progress, composite-minus-shuffled group-opportunity contrasts are 12.44 points [3.96,21.80] and 9.73 points [0.21,20.79] in the two banks, whereas differences of the table means are 12.72 and 8.61 points. The recorded world-bootstrap endpoint contrast without progress in the 236-state bank is $-2.90$ points [$-6.69$,$-0.12$] versus shuffled scores; the first-step contrast in the 240-state bank is $-18.47$ points [$-35.76$,$-4.05$] versus geometry alone. These exploratory world intervals are unadjusted for multiple comparisons; the permutations are descriptive controls without an inferential permutation $p$-value. Geometry comparisons vary by horizon; group opportunity is separate from physical probability calibration and the frozen D6 controller.

Local random, novelty and belief/LOS criteria give group retentions of 32.27\%, 36.80\% and 53.83\% in the 236-state bank. The respective 240-state values are 23.19\%, 24.30\% and 36.95\%. Composite-score values are 57.79/36.98\%; actual F/D6 opportunity retention is 52.46\% in the 236 bank. The composite's world contrast against belief/LOS spans zero in both banks. These locally adapted criteria are not complete reproductions of published planners.

The recorded selector protocol reserves target pixels and distance for evaluation and fixes cached upstream choices and progress mappings. Full histories and the upstream fixture identity remain unresolved (Section~\ref{sec:s-current-availability}); these selector paths do not establish the provenance of the cached inputs.

\subsection{Auxiliary reachable-horizon diagnostics}
\label{sec:s-horizon-diagnostics}
Define reachable value $\mathcal V_t(h)=\max_{v\in\Rset_t(h)}V_t(v)$. For $H_{\max}=5$ and development-selected $\eta=0.65$, the sufficient horizon and evidence debt are
\begin{equation}
\begin{aligned}
H_t^*&=\min\{h:\mathcal V_t(h)\geq\eta\mathcal V_t(H_{\max})\},\\
D_t^e&=1-\frac{\mathcal V_t(1)}{\mathcal V_t(H_{\max})+\epsilon}.
\end{aligned}
\label{eq:horizon}
\end{equation}
Nested reachable sets make $\mathcal V_t(h)$ nondecreasing. Evidence debt describes additional reachable sensing opportunity; verification debt concerns unresolved reference-score comparisons. The five-step setting is auxiliary: the frozen-WM control integration limits candidate depth to two.

\section{Computation costs and implementation}
\label{app:theme-computation}
\subsection{Complete-decision CPU cost}
\label{sec:s-cpu-expanded}
The 120-state CPU bank covers 40 worlds with identical candidates, geometry, feature inputs, models and Guard semantics across 11 timed cells. Three rounds give 3,960 complete decisions; repetitions are averaged within state before paired world resampling. D6 takes 15.86\% less time than serial Full, but 1.85\% more than native-batched Full (Table~\ref{tab:s-expanded-cpu}). A separate 45-combination candidate-count/batch study has 5,400 backend-scale rows.

\begin{table*}[tp]
\centering\small
\caption{Mean complete-decision CPU cost and per-decision work on common inputs. Full/reference-shortlist numeric caps are candidates per backend call; all denotes native batching. Student tail evaluations are not WM work.}
\label{tab:s-expanded-cpu}
\begin{tabular}{lrrrrr}
\toprule Cell & Total (ms) & Logical WM candidates & Backend calls & Model rows & Model accepts \\
\midrule
F, cap 1 &41.683&29.43&29.43&206.03&20/120\\
F, cap 2 &38.025&29.43&14.97&206.03&20/120\\
F, cap 6 &36.025&29.43&5.21&206.03&20/120\\
F, cap 12&34.980&29.43&2.89&206.03&20/120\\
F, all&34.437&29.43&0.62&206.03&20/120\\
Top 6, cap 1&35.082&3.70&3.70&25.90&20/120\\
Top 6, cap 2&34.463&3.70&1.85&25.90&20/120\\
Top 6, cap 3&34.270&3.70&1.23&25.90&20/120\\
Top 6, cap 6&33.837&3.70&0.62&25.90&20/120\\
D6, cap 1&35.074&2.60&2.60&18.20&21/120\\
Student, all&34.252&0&0.62&206.03&14/120\\
\bottomrule
\end{tabular}
\end{table*}

The Student row's calls and rows belong to its own backend. Full-reference batching preserves action/candidate selections in 840 cross-batch comparisons; another 240 comparisons are reflexive. Recorded parity flags agree; Section~\ref{sec:s-current-availability} specifies raw-input availability.

Candidate construction takes approximately 8.6\,ms and LOS/other features 23.5--23.7\,ms. Serial F reference work takes 8.185\,ms versus 0.690\,ms with native batching. D6 reference work takes 0.960\,ms plus 1.173\,ms for scheduling, Guard and progress handling. The affine serial-cost crossover calculation uses these serial-interface components.

The backend-scale study covers 12/24/48/96/192 candidates with batch caps 1/2/6/12/24/48/96/192/all, excluding LOS and policy work. The Windows 11/Python 3.12.4/PyTorch 2.4.1 environment uses one Torch thread, CPU inference and the control models. Memory peaks cover whole processes. Timed windows exclude warmup, loading and preparation; device power is unmeasured. The separate 320-state kernel measurements and untimed normalized-certificate replay have different boundaries (Sections~\ref{sec:shared-implementation} and~\ref{sec:s-integrated-reference}).

\subsection{Matched timing and scaling interfaces}
\label{sec:s-matched-access}
The matched session uses 120 states, 40 worlds and three rounds with single-thread float32 CPU inference, balanced order and identical feature reuse. It includes candidate, NoWorld/geometry and 56-feature construction; image processing, simulator physics, model loading and output I/O are excluded.

Table~\ref{tab:matched-serial} separates complete-decision latency, logical queries and backend invocations.

\begin{table*}[tp]
\centering\small
\caption{Same-session original-student timing and work. Work columns are per-decision means over all 120 states repeated three times, including gate declines; time is mean complete-decision latency.}
\label{tab:matched-serial}
\begin{tabular}{lrrrrr}
\toprule
Policy & Access & Time(ms) & Queries & Invocations & Seven-row inputs \\
\midrule
D6 & Batch & 37.018 & 2.600 & 2.600 & 18.200 \\
D6 & Per candidate & 36.911 & 2.600 & 2.600 & 18.200 \\
F & Batch & 36.028 & 29.433 & 0.617 & 206.033 \\
F & Per candidate & 43.958 & 29.433 & 29.433 & 206.033 \\
T6 & Batch & 35.964 & 3.700 & 0.617 & 25.900 \\
T6 & Per candidate & 36.541 & 3.700 & 3.700 & 25.900 \\
\bottomrule
\end{tabular}
\end{table*}

Per-candidate D6 is 16.03\% faster than per-candidate F, while native-batch D6 is 2.75\% slower than native-batch F. Per-candidate D6 is 2.45\% slower than native-batch F. The factorial and optimized-host sessions in Sections~\ref{sec:s-factorial-timing} and~\ref{sec:s-fixed-state-followup} use separate timing estimates and intervals. A larger-ensemble test matches reference quality in 7/19 fixed-six and 6/19 Guard/reference comparisons despite favorable timing.

Native scaling uses 16, 32 and the original 33--51 safe candidates on 120 inputs. A local 360-decision stress test checks whether cache and state size remain bounded. These tests do not cover 128 candidates, multiple workers, remote transport or onboard hardware.

\subsection{Complete-decision and image-processing timing}
\label{sec:s-cost-summary}
\begin{figure*}[!t]
\centering\includegraphics[width=.98\textwidth]{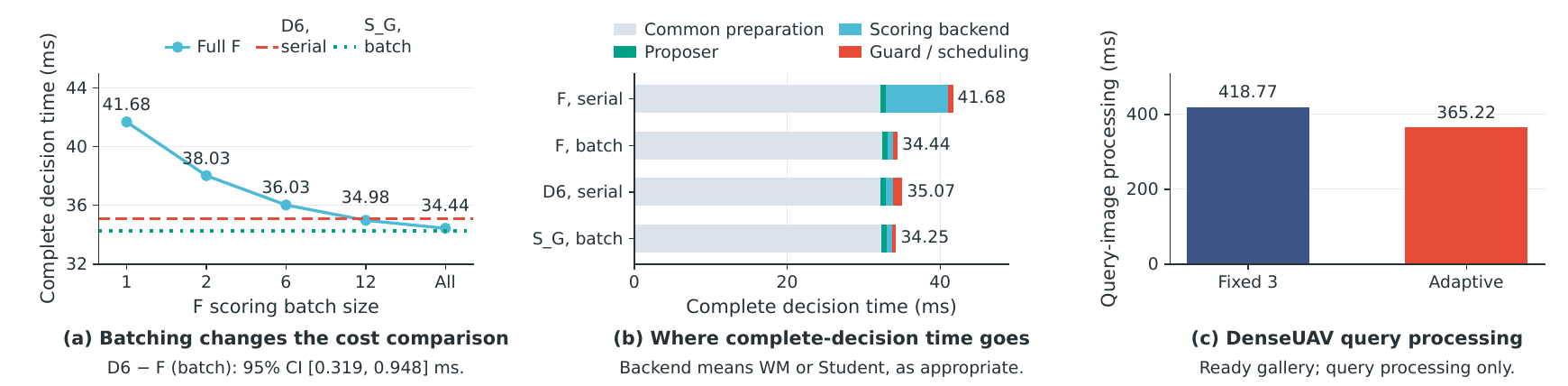}
\caption{\textbf{Complete-decision cost depends on batching and shared preparation.} (a) Identical 120-state inputs compare native batch caps and selective control. (b) Component costs include candidate construction, LOS/features, proposer, scoring backend and Guard; the Student backend makes no WM calls. (c) A separate 60-location DenseUAV subset measures actual query-image processing with a ready gallery. Timing repetitions do not add independent worlds.}
\Description{Serial full-reference and D6 complete decisions take 41.683 and 35.074 ms. Full batching takes 34.437 ms. Shared candidate and geometry preparation dominates these small models. On the DenseUAV subset, fixed-three and adaptive processing take 418.77 and 365.22 ms.}
\label{fig:cost-v48}
\end{figure*}

The matched-access CPU session reports Full/D6 43.958/36.911\,ms with serial access and 36.028/37.018\,ms with batching, separately from the 120-state study in Section~\ref{sec:s-cpu-expanded}. The 41.5-million-parameter ensemble's timing is accompanied by its reference-parity failures. CPU time, simulated mission cost and image service time are separate endpoints.

\subsection{Software cost and the query--latency boundary}
\label{sec:s-factorial-timing}
The factorial experiment uses identical 320 states, three rounds, and four crossed full/selective and reuse-off/on paths, plus a gated-full path in a separate timing window. It records 4,800 timings. A concurrent six-path student comparison records another 5,760. Six central processing unit (CPU) time quotas over 120 states and three paths give 2,160 (Table~\ref{tab:s-cpu}), for 12,720 formal timing rows in total. Input loading, warmup, verification, and disk output are excluded; candidate construction, LOS/56-dimensional features, models, scheduling, Guard, and progress handling are included.

With identical feature reuse, selective evaluation reduces query work by 94.75\% but increases mean latency by 1.67\% (world-cluster interval [0.86,2.51]\%). Full/selective means are 35.784/36.382\,ms (Table~\ref{tab:s-cost}). The reuse toggle preserves selected candidates, acceptance and query counts, with feature/WM parity tolerance $2\times10^{-5}$. Shared geometry dominates this structured backend. Separate optimized timing sessions are reported in Sections~\ref{sec:s-fixed-state-followup} and~\ref{sec:s-matched-access}.
\begin{table*}[tp]
\centering\small
\caption{Common-input factorial timing. The gated-full row is a separate timing window. Logical queries, constructed feature rows, and elapsed time measure different resources.}
\label{tab:s-cost}
\begin{tabular}{lrrrrr}
\toprule
Path & Mean ms & P95 ms & P99 ms & Queries/state & Feature rows/state \\
\midrule
Full, reuse off & 180.136 & 220.598 & 236.912 & 48.366 & 338.559 \\
Full, reuse on & 35.784 & 42.983 & 46.844 & 48.366 & 48.366 \\
Selective, reuse off & 48.800 & 74.867 & 82.262 & 2.538 & 63.591 \\
Selective, reuse on & 36.382 & 43.956 & 46.433 & 2.538 & 48.366 \\
Gated full, reuse on & 34.509 & 41.479 & 43.534 & 29.072 & 48.366 \\
\bottomrule
\end{tabular}
\end{table*}
\begin{table*}[tp]
\centering\small
\caption{Batch wall time divided by 120 states under six single-core-equivalent CPU-time quotas. CPU-time allocation does not reproduce a different processor architecture.}
\label{tab:s-cpu}
\begin{tabular}{lrrr}
\toprule
Quota & Full (ms) & 20,777 (ms) & 14,121 (ms) \\
\midrule
Normal & 33.286 & 33.627 & 33.861 \\
50\% & 62.477 & 64.719 & 64.295 \\
25\% & 127.558 & 132.743 & 129.004 \\
10\% & 306.972 & 321.743 & 316.547 \\
5\% & 618.400 & 628.366 & 633.365 \\
2\% & 1,019.425 & 1,385.534 & 1,222.096 \\
\bottomrule
\end{tabular}
\end{table*}

The concurrent optimized full/20,777-student/14,121-student paths take 34.674/35.370/35.665\,ms. The smaller student retains its own calibration, so parameter reduction is not the sole changed factor. Shared candidate/feature work costs approximately 32.5\,ms, while full backend/encoding work costs 1.56\,ms. For common work $C$, backend work $W_N/W_K$, proposal $S$, and added scheduling $H$,
\[
T_F=C+W_N,\qquad T_D=C+S+H+W_K.
\]
A timing gain requires $W_N-W_K>S+H$. Crossover calculations for more expensive backends are extrapolated scenarios.

The implementation-path measurement records 179.99/46.36\,ms and CPU package energy 5.185/1.360\,J for full/selective paths with different construction and ranking. The common-input factorial isolates reuse and has no accompanying power reading. CPU package energy is separate from UAV battery energy.

\subsection{Optimized-host timing and repeatability}
\label{sec:s-fixed-state-followup}
A separate optimized host session has five paths, 120 states from 40 reused worlds and three rounds (1,800 timing rows). Full/selective means are 34.757/35.541\,ms: selective evaluation is 2.25\% slower [1.40,3.11]\%. Hardware deployment and power are not measured. Fixed-input repeats agree in 360/360 rows, while controlled perturbations change some actions; fixed-input numerical repeatability does not control simulator-state drift.

\subsection{Decision logs and budget accounting}
\label{sec:s-decision-accounting}
State caches and decision, query, state-context, trajectory and sensor logs record 23,851 decisions, including 10,419 D6 query events in the second regime. Recorded D6 charges, caps, input epochs, candidate identities and queried witnesses are consistent. Checks cover epoch-plus-candidate key construction, without full-array reconstruction of every internal key. Source-input availability is specified in Section~\ref{sec:s-current-availability}.

Engineering D6 records low-value gating, Guard rejection, progress control, capped acceptance and reactive fallback separately from strict certification. Exhausting a budget does not satisfy a certificate. Caps apply per decision; a candidate's seven perturbation rows are model inputs, not additional candidates or observations. F can evaluate many candidates in one native batch, so candidate work and invocation count are reported separately. Executed-policy query totals also reflect different visited states; common-input latency is evaluated in Section~\ref{sec:s-matched-access}.

\subsection{Shared computation reduces implementation overhead}
\label{sec:shared-implementation}
The kernel study applies shared feature preparation, fused seven-row evaluation and equivalent preprocessing to Full, fixed-six and sequential leader--rival rules on 320 states. Nine implementations and three repeats yield 8,640 timed rows. Reference scores agree within $7.87\times10^{-6}$ and all tested queried action outcomes are preserved.

\begin{figure*}[tp]
\centering\includegraphics[width=.98\textwidth]{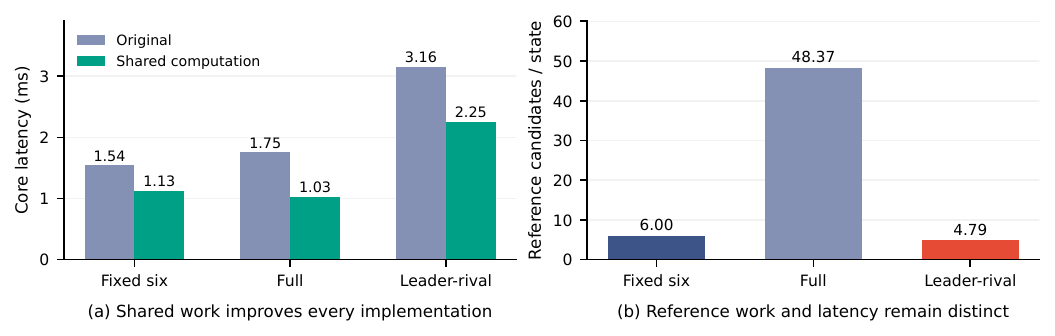}
\caption{\textbf{Shared computation and reference work.} (a) Equivalent feature/preprocessing optimization on 320 states, with three repeats per path. Bars are same-host session means. (b) Logical reference work and latency have different orderings. The measured boundary is the scoring/verification core; complete-decision tests also include geometry and Guard. The normalized certificate is untimed.}
\Description{Original and optimized core latency is 1.542 and 1.129 ms for fixed-six, 1.754 and 1.031 ms for Full, and 3.159 and 2.248 ms for sequential leader-rival. Reference-query means are six, 48.366 and 4.791.}
\label{fig:kernel-upgrade}
\end{figure*}

Figure~\ref{fig:kernel-upgrade} compares the implementations. Full latency decreases from 1.754 to 1.031\,ms, fixed-six from 1.542 to 1.129\,ms, and sequential leader--rival from 3.159 to 2.248\,ms. A one-batch relative-anchor rule takes 1.118\,ms versus native Full's 1.031\,ms. These core timings exclude candidate/LOS construction, camera processing, physical Guard and flight. Relative-rule timing intervals appear in Section~\ref{sec:s-anchored-kernel}; the normalized certificate is untimed.

\subsection{Numerical-kernel cost}
\label{sec:s-anchored-kernel}

The complete-decision CPU measurements use 120 states, 40 worlds, 11 cells and three repeats (Section~\ref{sec:s-cpu-expanded}). Each of the two kernel studies instead has 8,640 records: 320 states, nine methods and three same-host repeats. Kernels use prebuilt 56-dimensional features and exclude candidate/LOS construction, physical Guard/progress, image decoding/detection, flight, loading and I/O. The anchored study applies the same $4\times10^{-5}$ numerical bound expansion to bounded methods. Optimized Full-WM takes 1.030583\,ms, relative leader/rival 2.059321\,ms and relative one-batch-six 1.118337\,ms. Relative leader/rival minus Full is $+1.028738$\,ms $[0.931846,1.123394]$; one-batch-six minus Full is $+0.087754$\,ms $[0.052150,0.127070]$. Both selective kernels are slower than native Full. The normalized certificate in Section~\ref{sec:s-integrated-reference} is untimed.

\section{Fault isolation and recovery}
\label{app:theme-faults}
\subsection{Query faults and same-epoch cache reuse}
\label{sec:s-historical-runtime}
Forty-eight injected query faults on eight states from one world exercise timeout exceptions, ordinary exceptions, malformed outputs, nonfinite values and missing fields. Each charges the attempted query and falls back; a seventh call is refused after six attempts. The baseline runtime passes 62/63 checks: an in-place configuration change can reuse a stale same-epoch cache entry. Section~\ref{sec:s-hardened-runtime} evaluates a separately identified content-sensitive runtime.

Four hundred consecutive callback-thrown timeout events produce one-attempt fallback and zero observed net resident-set memory growth. This exception-handling test leaves blocked remote-call cancellation, flight recovery and sensor-chain safety untested.

\subsection{Content-sensitive fault runtime and software checks}
\label{sec:s-hardened-runtime}
The hardened runtime uses content-sensitive identity, response qualification and revised query/sensor handling under an identity separate from the baseline runtime in Section~\ref{sec:s-historical-runtime}. Its 64 executions (Table~\ref{tab:s-new-fault}) record 48 fault triggers and rejections, 32 subsequent healthy-query recoveries and 16 valid-observation recoveries. The run ledger contains 2,174 decisions, 5,907 charged attempts and 41,237 reported model-input rows. Faults that prevent evaluation separate charged attempts from evaluated rows.

\begin{table}[tp]\centering\small
\caption{Single-fault closed-loop settings on four reused worlds and two starts.}
\label{tab:s-new-fault}

\begin{tabular}{lrrrr}
\toprule

Setting & Runs & Successes & Injected & Rejected \\
\midrule

baseline normal & 8 & 5 & 0 & 0 \\

hardened normal & 8 & 5 & 0 & 0 \\

cache content & 8 & 7 & 8 & 8 \\

model nonfinite & 8 & 8 & 8 & 8 \\

response identity & 8 & 6 & 8 & 8 \\

query timeout & 8 & 5 & 8 & 8 \\

sensor bad & 8 & 2 & 8 & 8 \\

sensor stale & 8 & 6 & 8 & 8 \\

\bottomrule
\end{tabular}
\end{table}

The local suite for this runtime comparison contains 1,332 test records plus ten additional checks, excluding development duplicates. A regression check reproduces the stale-cache failure of the 62/63 runtime; the hardened identity preserves normal-path policy parity.

The simulator remote procedure call (RPC) completes before the 0.3005--0.3006\,s deadline in these timeout tests. Cancellation of blocked work is tested separately with local surrogates; interruption of a genuinely blocked live simulator connection and long-duration network robustness remain unverified.

\subsection{Local cancellation and recovery boundaries}
\label{sec:s-local-fault-boundaries}
The local fault study has 60 request cases: 55 faults and five healthy controls. All pass the tested recovery boundary, with no late-result cache mutation. Eight exceed the nominal 80\,ms point; none exceeds the declared 80+50\,ms tolerance. Eleven production-policy blocked-query cases fall back correctly. Table~\ref{tab:s-new-fault} reports fault-specific mission outcomes for the separate 64 simulator executions.

\subsection{Local boundary-case tests}
\label{sec:s-local-followup}
Five synthetic boundary-case families have five repeats each: invalid probability then valid; stale/nonfinite then valid; six failures at a six-attempt cap; epoch change with a late response; and identical-content reuse followed by content change. The 25 cases cover local boundary handling. Detailed traces contain 65 event rows for the first three families; the other two provide assertion summaries. These local cases do not test full deployed-validator integration and use a separate cohort from the 64 simulator fault missions.

\section{Public-image retrieval and transfer}
\label{app:theme-retrieval}
\subsection{DenseUAV complete-gallery transfer protocol}
\label{sec:s-public-transfer}
The official DenseUAV protocol uses 777 query locations, 2,331 UAV query images, 3,033 gallery locations, and 18,198 satellite images. Five frozen backbones use the stopping policy and thresholds fixed on SUES-200. Fixed-three and adaptive location-level recall at rank one average 82.394\% and 82.059\%, a paired difference of $-0.335$ points [\,$-0.746$,0.077\,]. Mean query frames decrease from three to 2.6 (13.33\% fewer query feature extractions). If gallery rebuilding is included, the combined extraction reduction is 1.514\%. Official image-level recall and mean average precision decrease by 0.412 and 0.315 points, with paired intervals [\,$-0.824$,$-0.026$\,] and [\,$-0.505$,$-0.129$\,].

\begin{figure*}[tp]
\centering\includegraphics[width=.97\textwidth]{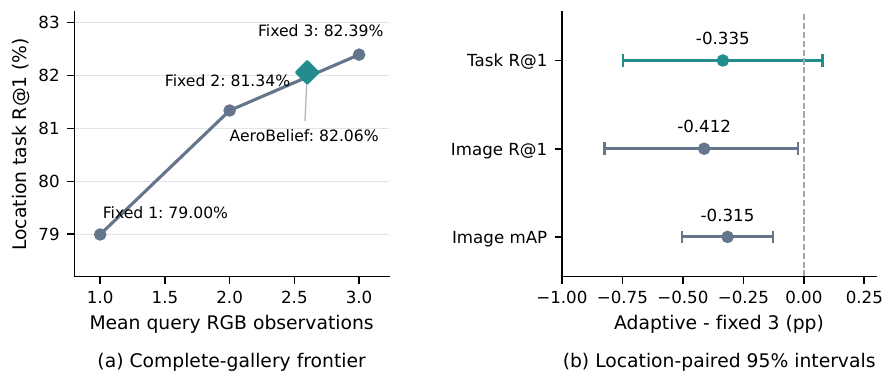}
\caption{Complete-gallery DenseUAV transfer. Frozen backbones share the same 777 query and 3,033 gallery locations. Location and official image metrics have separate definitions, with paired location-cluster uncertainty.}
\label{fig:s-transfer}
\Description{Complete-gallery DenseUAV transfer. Frozen backbones share the same 777 query and 3,033 gallery locations. Location and official image metrics have separate definitions, with paired location-cluster uncertainty.}
\end{figure*}

Figure~\ref{fig:s-transfer} reports official-gallery metrics; the 877-location custom gallery is a separate evaluation population. SUES-200 uses the official split of 120 training and 80 test locations. Reanalyses of those test locations share the same evaluation population.

\subsection{DenseUAV stopping frontiers and query processing}
\label{sec:s-public-retrieval-frontiers}
\label{app:frontiers}
DenseUAV has 777 independent query locations across five backbones. Each rule has 3,885 location/model records, not 3,885 independent locations. Location-task retrieval pools satellite image scores within each location before frame aggregation, whereas official image-level R@1 and mean average precision (mAP) retain image ranks.

\begin{table}[tp]
\centering\small
\caption{DenseUAV complete-gallery location-task accuracy and query work. Five backbones repeat the same 777 locations.}
\label{tab:dense-nine}
\begin{tabular}{lrr}
\toprule
Rule & Task R@1(\%) & Mean images \\
\midrule
Fixed 1 & 78.996 & 1.000 \\
Fixed 2 & 81.338 & 2.000 \\
Fixed 3 & 82.394 & 3.000 \\
AeroBelief & 82.059 & 2.600 \\
Frozen margin & 81.390 & 2.004 \\
Frozen entropy & 81.725 & 2.560 \\
Matched hash & 81.725 & 2.600 \\
Dev. margin & 82.394 & 2.980 \\
Dev. entropy & 82.394 & 3.000 \\
\bottomrule
\end{tabular}
\end{table}

Table~\ref{tab:dense-cpu} reports measured query-processing time and image counts.

\begin{table}[tp]
\centering\small
\caption{Actual CPU query processing on 60 hash-selected locations, five backbones and two rounds; 600 measurements per rule. Query timing excludes gallery construction and reuses one ready gallery.}
\label{tab:dense-cpu}
\begin{tabular}{lrr}
\toprule
Rule & Mean time(ms) & Mean images \\
\midrule
Fixed 1 & 139.46 & 1.000 \\
Fixed 2 & 279.96 & 2.000 \\
Fixed 3 & 418.77 & 3.000 \\
AeroBelief & 365.22 & 2.603 \\
Dev. margin & 415.21 & 2.967 \\
Dev. entropy & 419.31 & 3.000 \\
Matched hash & 361.59 & 2.593 \\
\bottomrule
\end{tabular}
\end{table}

The CPU subset has 4,200 records and is a separate accuracy population from the full 777-location cohort. Fourteen repeated prediction rows (seven unique seed/location/rule keys) differ between CPU processing and historical records. CPU accuracy, timing and frame counts use the same recorded outputs. Graphics processing unit (GPU) gallery timings and CPU query timings come from separate device sessions and do not measure a same-device end-to-end pipeline.

\subsection{Frozen stopping-rule frontier}
\label{sec:s-frozen-retrieval-frontier}
\begin{figure}[!htbp]
\centering\includegraphics[width=.98\columnwidth]{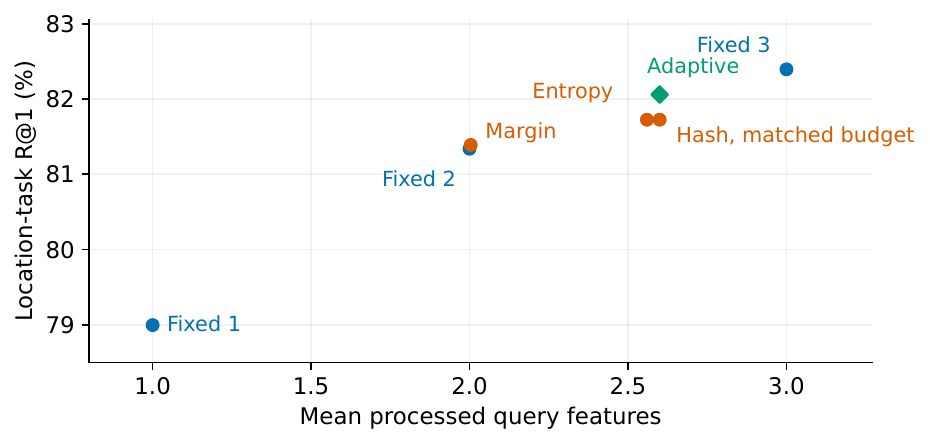}
\caption{Frozen stopping-rule frontier on the same 777 locations and five backbones. Points are measured prefix-output reanalyses; no interpolated accuracy curve or newly tuned test threshold is introduced.}
\Description{Fixed one, two and three frames, adaptive, margin, entropy and matched-budget hash stopping have distinct feature costs and location-task accuracies. Adaptive uses 2.6 query features with 82.059 percent task recall.}
\label{fig:s-frontier-v51}
\end{figure}

Figure~\ref{fig:s-frontier-v51} compares frozen margin-only, entropy-only and matched-mean-budget hash rules on the existing five-backbone DenseUAV prefix outputs, without new training, feature extraction or locations. Section~\ref{sec:s-public-transfer} gives official-gallery accuracy losses and work denominators; Table~\ref{tab:dense-nine} includes validation-selected alternatives.

\begin{figure*}[!t]
\centering\includegraphics[width=.98\textwidth]{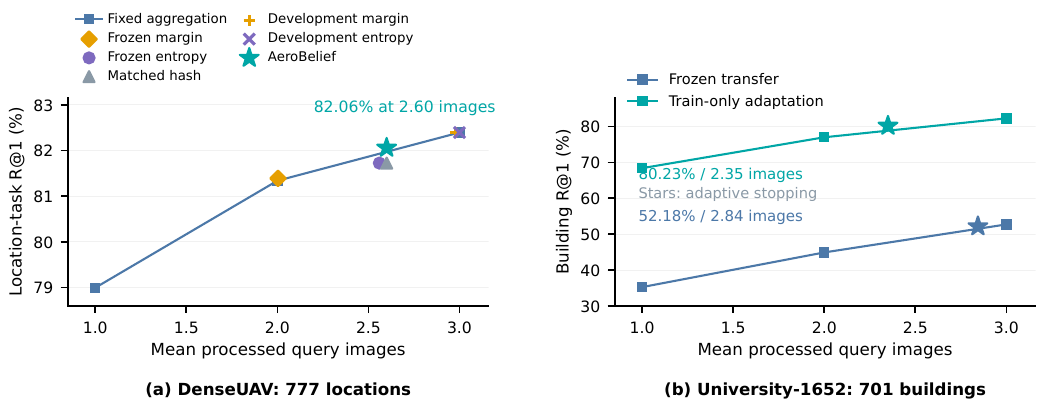}
\caption{\textbf{Retrieval quality versus query work.} (a) DenseUAV compares frozen stopping controls on 777 locations and five backbones. (b) University-1652 separates frozen transfer from train-only adaptation on 701 test buildings. Stars mark adaptive stopping. Query-image count is a work measure; actual DenseUAV CPU timing is shown separately in Fig.~\ref{fig:cost-v48}.}
\Description{Dense adaptive stopping attains 82.059 percent task R@1 at 2.60 images. University frozen/adapted adaptive values are 52.183/80.228 percent at 2.844/2.350 images, versus fixed-three 52.725/82.311 percent.}
\label{fig:dense-full-v42}
\end{figure*}

\subsection{University-1652 prescribed-view transfer and adaptation}
\label{sec:s-university-transfer}
University-1652 uses 701 test buildings, a 951-building gallery and three prescribed views per test building, separately from the official full-query-image leaderboard. Gallery-ID pairs 0239/0240 and 0761/0768 have identical image hashes, leaving 949 distinct hashes for 951 IDs. Metrics retain recorded tie ordering because exact-score ties can change canonical rank and average precision across sorting implementations. Train-only adaptation uses 631 training and 70 validation buildings disjoint from test, five seeds and 12 epochs. Frozen transfer and adapted checkpoints keep separate model identities; repeated test evaluation does not add a confirmatory population.

\begin{table}[!htbp]
\centering\small
\caption{University-1652 primary prescribed-view outcomes. Adaptation and stopping are different factors. Processed-image count is not measured latency.}
\label{tab:university-transfer}
\begin{tabular}{lrrr}
\toprule
Encoder & Rule & R@1(\%) & Mean images \\
\midrule
Frozen & Fixed 1 & 35.235 & 1.000 \\
Frozen & Fixed 2 & 44.907 & 2.000 \\
Frozen & Fixed 3 & 52.725 & 3.000 \\
Frozen & AeroBelief & 52.183 & 2.844 \\
Frozen & Dev. margin & 52.725 & 2.991 \\
Frozen & Dev. entropy & 52.725 & 3.000 \\
Frozen & hash 40 pct & 49.900 & 2.601 \\
Adapted & Fixed 1 & 68.417 & 1.000 \\
Adapted & Fixed 2 & 77.033 & 2.000 \\
Adapted & Fixed 3 & 82.311 & 3.000 \\
Adapted & AeroBelief & 80.228 & 2.350 \\
Adapted & Dev. margin & 82.311 & 2.925 \\
Adapted & Dev. entropy & 82.311 & 2.995 \\
Adapted & hash 40 pct & 80.143 & 2.601 \\
\bottomrule
\end{tabular}
\end{table}

Under the primary prescribed-view order, adaptive R@1 loses 0.542 points for frozen transfer and 2.083 points after adaptation versus fixed three (Table~\ref{tab:university-transfer}). The adapted paired building-cluster interval is [$-2.739$,$-1.484$] points, with 84 harms and 11 rescues. On DenseUAV, adaptive minus equal-work hash is $+0.335$ points [$-0.129$,$+0.798$], leaving superiority unresolved.

Reverse-order adaptation is an order-sensitivity analysis: adaptive R@1 is 82.996\% with 24.45\% fewer images than fixed three. Fixed two is both more accurate (83.538\%) and uses fewer images (2) than reverse-order adaptive stopping (82.996\%, 2.266 images). The exported risk value is $\exp(-700)$ on every University-1652 decision trace, so this subgate does not affect stopping in these experiments; the other stopping gates are active.

\subsection{Public-image retrieval frontiers}
\label{sec:s-retrieval-frontier-summary}
Retrieval uses fixed location sets and train-only adaptation; query-image counts measure feature work, with ready-gallery processing timed separately.

Figure~\ref{fig:dense-full-v42} reports frozen and train-only-adapted retrieval frontiers.

\subsection{SUES-200 stopping-rule evaluation}
\label{sec:s-sues-efficiency}
The SUES-200 efficiency study replays 1,600 query episodes on the official 80 test locations at 150, 200, 250 and 300\,m. Each location supplies five ten-image episodes per altitude. Fixed-ten aggregation and adaptive stopping use the same frozen SE-ResNet retrieval scores and a 200-location gallery. We fit the stopping gate using the 120-location calibration split and freeze the altitude-specific configurations before the v4.1 replay. Earlier SUES-200 test results informed the design, so this comparison is exploratory.

\smallskip
\noindent\textbf{Recall and image use by altitude.}
\par\smallskip
{\small\centering
\begin{tabular}{@{}rrrrr@{}}
\toprule
Altitude & \multicolumn{2}{c}{R@1 (\%)} & \multicolumn{2}{c}{Mean images} \\
(m) & Fixed 10 & Adaptive & Fixed 10 & Adaptive \\
\midrule
150 & 60.25 & 60.00 & 10 & 8.47 \\
200 & 73.25 & 71.00 & 10 & 4.16 \\
250 & 78.50 & 77.25 & 10 & 5.29 \\
300 & 78.50 & 78.00 & 10 & 5.13 \\
\midrule
All & 72.625 & 71.5625 & 10 & 5.76 \\
\bottomrule
\end{tabular}\par}
\smallskip

Fixed ten and adaptive stopping rank the correct location first in 1,162 and 1,145 of 1,600 episodes, respectively. The paired difference is $-1.0625$ percentage points (95\% location-cluster interval [$-2.0625$,$-0.0625$]). Adaptive stopping uses 9,216 of the 16,000 images processed by fixed ten, a reduction of 42.40\% (95\% interval [38.36,46.27]). Both intervals use 5,000 paired resamples of the 80 location IDs (seed 20271910), keeping all heights and episodes from each sampled location together. Image counts measure query work.

The accompanying \texttt{sues200\_evidence} source directory includes the episode and stopping traces, per-altitude calibration and checkpoint identities, and code to recompute the reported results. The historical E2 Code/Data archive also contains the calibration and test score matrices.

\FloatBarrier

\section{Self-collected Mavic evaluation}
\label{app:theme-mavic}
\subsection{Complete image pipeline and confirmation clocks}
\label{sec:s-pixel-stream}
Thirty-two original-cadence jobs cross eight recorded tasks with AllFrames, static context, combined feedback and fixed interval. Jobs stop at confirmation or route end; four unconfirmed fixed-interval jobs reach route end. Another 32 segments replay each task's first 120 capture seconds at eight-times rate. AllFrames bypasses the processing cap. These fixed-route jobs execute pixel processing with a classical query-conditioned detector, separately from the three-dimensional structured WM.

Each process creates a detector and executes reading, decoding, detection, flat-ground footprint projection, belief updates, admission and confirmation on a Windows 11 Ryzen 7 6800HS machine with 16 logical processors and 16,363,622,400 bytes of RAM. Four policy processes share the host; OpenCV uses two threads per process and operating-system-managed disk caching. Startup/template initialization is logged separately. The measurements use CPU inference and shared-host service time; cold-disk loading and device power were not measured.

Table~\ref{tab:s-pixel-jobs} summarizes the original-cadence processing jobs.

\begin{table*}[tp]
\centering\small
\caption{Original-cadence actual-image pipeline. Service sums count active processing seconds under shared-host load; result-ready timestamps track elapsed wall-clock time. Confirmations use final reviewed labels.}
\label{tab:s-pixel-jobs}
\begin{tabular}{lrrrrrr}
\toprule Method & Jobs & Target confirmed & Two-full support & Images & Service (s) & Max queue \\
\midrule
AllFrames (uncapped)&8&8/8&8/8&1,762&4,987.52&244\\
Static context&8&8/8&8/8&445&1,610.00&15\\
Combined feedback&8&8/8&8/8&422&1,495.23&13\\
Fixed interval&8&4/8&3/8&603&2,130.76&9\\
\bottomrule
\end{tabular}
\end{table*}

The execution records contain 3,963 detection calls: 3,232 in original-cadence jobs and 731 in stress segments. Binary detections, scores and original-cadence selection/confirmation order agree with the fixed reference records. Service distributions, read/decode stages, queues, RSS, input/processing/cancellation counts and result-ready timestamps are recorded; raw-image availability is stated in Section~\ref{sec:s-current-availability}.

Capture time is the relative photograph timestamp. Result readiness is elapsed time from the stream origin after detector initialization; arrival-to-ready latency includes queue waiting and processing. Confirmation becomes ready when its last required inference completes. At 660\,s, static/combined counts are 6/8 and 5/8 by capture time, and 5/8 each by result-ready time. Final counts are 8/8. The e0928 task's 22-second difference is measured in capture time.

At eight-times rate, static/combined/fixed/AllFrames segments process 94/94/123/420 images, with maximum queues of 51/51/51/58 frames. All 32 original-cadence and 32 stress jobs record zero overload drops within these jobs; inputs canceled after a stopping decision are counted separately. The stress measurements cover short prefixes; whole-task eight-times-rate completion was not measured.

A separate four-task/three-route image-processing batch has 199 static calls/141.68\,s and 195 combined calls/139.18\,s, with 4/4 confirmations each. Its 1.77\% single-pass time difference, the all-route shared-host totals and the partial-label 659.09/159.53\,s measurements use separate processing cohorts.

\subsection{Expanded independent review and target identity}
\label{sec:s-review-expanded}
The expanded review selects 265 task-image cases: 100 full/partial cases, 160 negative-target judgments from 140 photographs sampled at 20 per route, and five identity-conflict cases. G01 has two designated targets, both required absent for a photograph to enter its negative pool. Thirty-nine cases reuse independent judgments; 226 additional initial judgments cover 204 photographs.

Initial agreement is 218/226 (96.46\%, $\kappa=0.9074$): 159/159 random-negative judgments, 54/62 positive-completion judgments, and 5/5 identity-conflict judgments agree. A third reviewer adjudicates the eight disagreements: seven e0928 frames are full and frame 251 of the D02 farm-machine task is partial. The exports distinguish initial absent judgments from adjudicated labels. Independent coordinate-valid boxes cover all 100 visible cases; detector-box agreement is measured separately on the 18 confirming photographs (Section~\ref{sec:s-mavic-review}).

\begin{figure*}[tp]\centering
\includegraphics[width=.96\textwidth]{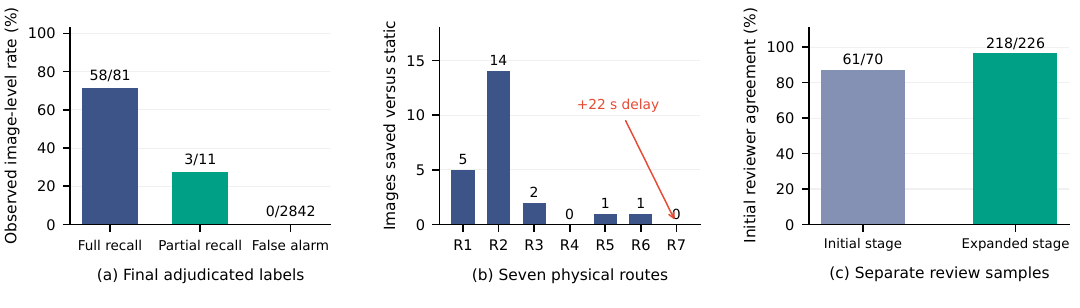}
\caption{\textbf{Final-label detection, route-level processing and separate review stages.} (a) Under final human labels, detections are 58/81 full, 3/11 partial and 0/2,842 absent cases after source-template exclusion. (b) Combined feedback saves 5, 14, 2, 0, 1, 1 and 0 images across seven physical routes, totaling 23; the two G01 target tasks share one route. (c) Initial agreement is 61/70 in the 70-judgment review and 218/226 in the expanded review. These are different samples, not a paired before/after comparison. The resolved-label subset of the 70-judgment review, 61/66, is reported separately in Section~\ref{sec:s-mavic-review}.}
\Description{Final-label full and partial recall are 58 of 81 and 3 of 11, with zero detections on 2,842 absent cases. Seven route-level savings sum to 23 images. The two distinct initial-review cohorts have agreement 61 of 70 and 218 of 226.}
\label{fig:s-mavic-review}
\end{figure*}

Across 2,942 task-image cases, final human labels contain 89 full, 11 partial and 2,842 absent cases, with no unresolved label. Excluding eight source templates gives full detection 58/81, partial detection 3/11, total visible detection 61/92 and absent-target detections 0/2,842. All main confirming pairs have full-image support. Figure~\ref{fig:s-mavic-review} uses these final detector denominators and separates the two review samples; initial agreement describes each sampling scheme rather than whole-dataset annotation accuracy.

Task-presence confirmation accepts partial-positive support for the correct designated target; the strict endpoint requires two full images. The emphasis on task presence was specified after the experiments. Fixed interval achieves 4/8 and 3/8 under the respective definitions; Table~\ref{tab:s-strong-stopping} also reports both endpoints for random controls.

The main evaluation uses retrospectively prepared within-route templates (Section~\ref{sec:s-mavic-review}). Separate same-instance queries captured on earlier routes yield 65 processed images and confirmation at 522\,s for Route 008 (query captured 428\,s earlier), and 72 images without confirmation for Route 009 (query captured 1,264\,s earlier). Under final labels, source-neighbor exclusion radii 0/1/3/5 yield task-presence counts 8/6/5/4 and two-full counts 8/5/5/4 for static, combined and AllFrames. These counts use per-run final-label support flags. At radius one, the D02 farm-machine pair includes partial frame 251. Exclusion settings reuse the same routes and tasks.

For the D02 vehicle task, frame 138, detector analyses use the clean 5,280$\times$3,956 native-input log/cache score 0.28447015; the annotation export stores 0.48175245. Both are below the detection threshold, leaving the binary outcome negative. Clean-input fields supply only the score and detection status; marked-image geometry is excluded from geometric validation. Source-representation and label-field boundaries are specified in Section~\ref{sec:s-current-availability}.

\subsection{Adjudicated feedback, baselines and deadline sensitivity}
\label{sec:s-feedback-expanded}
Capped controls share an image quota cap of 120, one added quota every 8\,s, a 20\,s periodic trigger and two processed positives within 30\,s for confirmation. Positive/negative updates use fixed engineering settings, including 0.8 attenuation of newly negative cells. Static/positive-only/negative-only/combined feedback each achieve 8/8 task-presence and strict confirmations using 445/434/428/422 images. Combined saves 23 images (5.1685\%); five of eight tasks save images and three tie. The two G01 tasks share a route cluster. Positive and negative updates separately save 11 and 17 images, while their combination saves 23 rather than 28.

\begin{table*}[tp]
\centering\small
\caption{Stopping controls at the shared operating point under final human labels. Task-presence and two-full-image endpoints are reported separately. AllFrames is budget exempt; every random seed is shown.}
\label{tab:s-strong-stopping}
\begin{tabular}{lrrrr}
\toprule Method & Target confirmed/8 & Strict full/8 & False confirmed & Images \\
\midrule
AllFrames&8&8&0&1,762\\
Static context&8&8&0&445\\
Positive only&8&8&0&434\\
Negative only&8&8&0&428\\
Combined&8&8&0&422\\
One-shot reserve variant&8&8&0&422\\
Fixed interval&4&3&0&603\\
Novelty&5&5&0&526\\
Random 0&4&4&0&618\\
Random 1&7&5&0&507\\
Random 2&6&6&0&533\\
Random 3&3&2&0&619\\
Random 4&4&3&0&596\\
Uniform, no updates&4&3&0&603\\
Uniform, negative&5&5&0&306\\
Uniform, positive&6&5&0&568\\
Uniform, both&5&5&0&310\\
\bottomrule
\end{tabular}
\end{table*}

\begin{figure*}[tp]
\centering\includegraphics[width=.98\textwidth]{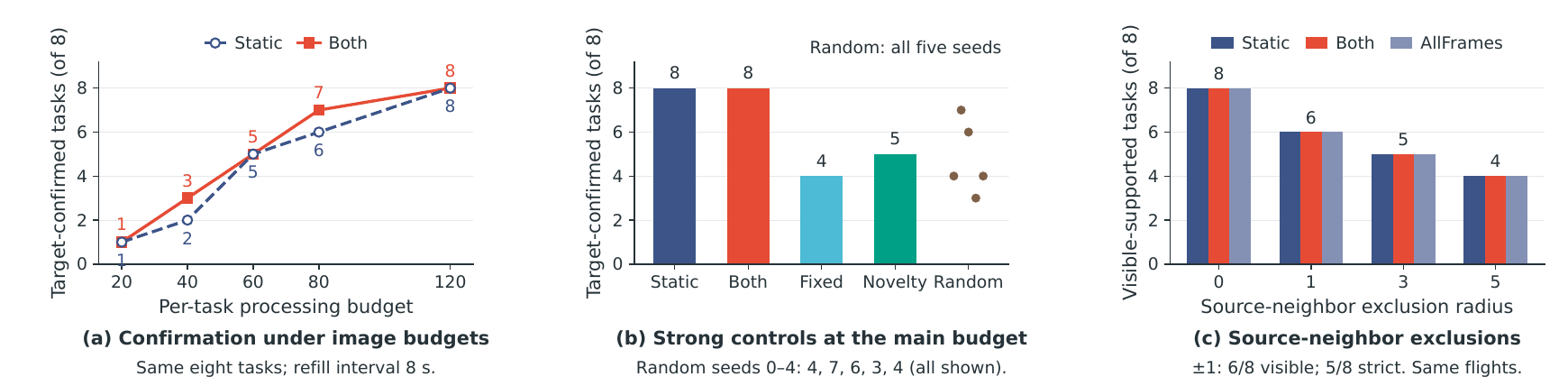}
\caption{Budget, strong-baseline and source-neighbor sensitivity on the same eight tasks. (a) Static/combined target confirmations versus cap. (b) Strong sampling controls include every random seed and use the same target-presence endpoint. (c) Source-neighbor exclusions reduce the number of available qualifying positive pairs. Repeated settings do not increase the physical route/task denominator.}
\Description{At caps 20,40,60,80,120 static/combined confirmations are 1/1,2/3,5/5,6/7,8/8. Fixed interval confirms 4 and novelty 5 of 8, while random seeds span 3 to 7. Source-neighbor exclusion radii 0,1,3,5 leave 8,6,5,4 target-presence and 8,5,5,4 strict confirmations.}
\label{fig:s-feedback-budget}
\end{figure*}

Figure~\ref{fig:s-feedback-budget} shows budget and sampling sensitivity; result-ready timing is defined in Section~\ref{sec:s-pixel-stream}.

At cap 40, combined confirms three tasks versus static two with 288 versus 289 images; at cap 80, seven versus six with 419 versus 427 images. Caps 20/60/120 give equal final counts. Capture-deadline and equal-processed-count curves use distinct endpoints. Seven task pairs confirm at the same capture time; e0928 changes from 657 to 679\,s with no image saving. At 960\,s both methods confirm 6/8, and at 1920\,s both confirm 8/8; the 660\,s two-clock comparison is in Section~\ref{sec:s-pixel-stream}. The e0928 footprint scores 0.9983/0.9773 miss the 1.0 threshold at 655/657\,s, so the next qualifying pair is processed at 677/679\,s. An exploratory threshold 0.99 selected among 1.0/0.99/0.95/0.90 on development routes retains 422 images/eight confirmations and removes this delay; the factorial uses the original threshold.

Under injected wrong priors, static/combined each achieve 6/8 task-presence confirmations and 5/8 versus 6/8 strict confirmations, using 510/491 images. Injected burst misses give 5/5 under both definitions and 518/489 images. A separate 42-task designated-target-absent replay cohort comprises 33 route--instance combinations and 168 four-policy replays, with no observed false confirmation. Static and negative-feedback processing totals are 3,764 and 3,603 images (4.28\% fewer). All these conditions reuse recorded images.

The 1,288 replay records, 128 controlled-stress conditions and 80 development variants reuse the eight main tasks. Prefix/suffix checks cover software causality. Task/route/target mappings, budget/refill/seed results and confirmation-support flags are included with the analysis exports.

\subsection{Independent 70-judgment Mavic review}
\label{sec:s-mavic-review}
The risk-stratified review has 70 task-image judgments over 69 distinct image hashes. It includes all 18 main-comparison confirming photographs, partial/unknown cases, selected missed positives and difficult/random negatives. The review protocol presents the query instance and clean image, with labels, detector scores, identity metadata and guide boxes hidden. Thirty-seven judgments use 5280$\times$3956 images and 33 use 1920$\times$1439 images. The expanded review's 226 additional initial judgments are a separate study (Section~\ref{sec:s-review-expanded}).

Initial agreement in the 70-judgment study is 61/70 (87.14\%, $\kappa=0.78644$), or 61/66 (92.42\%, $\kappa=0.86847$) excluding four previously unknown labels. The nine disagreements comprise those four unknowns and five resolved-label disagreements. A second reviewer agrees with the first on five of the nine and differs on four; adjudication changes eight original labels and leaves one. This adjudication yields 90 full, 10 partial and 2,842 absent cases, with no unknown. Template exclusion leaves 82 full and 10 partial cases, of which 59 and two are detected. Before that adjudication, eligible full/partial denominators were 81/7. These label counts describe the 70-judgment study; the final label set is reported in Section~\ref{sec:s-review-expanded}.

All 18 confirming photographs have independent full-presence and predicted-center-in-box support. Eleven meet IoU$\geq 0.5$ and mean IoU is 0.56374. The 32 primary feedback replays reuse these photographs in 64 confirmation-support positions. Geometric validation uses the independent boxes on these 18 photographs; the score-source discrepancy for a separate D02 frame is specified in Section~\ref{sec:s-review-expanded}.

Schedule reconstruction compares 48 replay schedules, 40 policy-output pairs and 144 truncation checks. Belief values agree within $5.69\times 10^{-14}$; floating-point reduction order can change bitwise hashes. Human labels enter evaluation only. Label-only re-evaluation of these 48 schedules, 1,000 historical-control records and 240 sensitivity records changes no confirmation-quality judgment.

The 70-judgment adjudication identifies earlier first-complete frames on the D02 vehicle task and G01 target 1, increasing absolute visibility-to-confirmation delays by 189 and 108\,s without changing policy-to-policy timing differences. Paired processing and deadline results are reported in Sections~\ref{sec:s-pixel-stream} and~\ref{sec:s-feedback-expanded}.

The seven-route cluster interval for the 5.1685\% processing saving is [1.2887,10.1167]\%.

The query-template evaluation uses within-route sources prepared retrospectively; five source frames were captured after the combined-feedback confirmation. Preflight template availability is unverified after self-match exclusion. Earlier-route same-instance sensitivity cases are reported separately in Section~\ref{sec:s-review-expanded}.

\begin{figure*}[tp]
\centering\includegraphics[width=.96\textwidth]{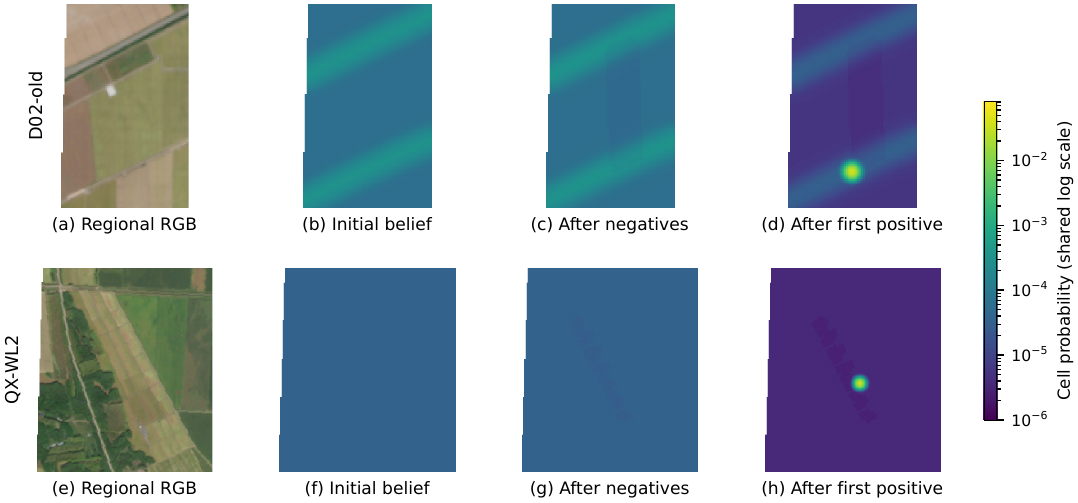}
\caption{\textbf{Regional context and observation updates.} Regional RGB and belief arrays show initial, post-negative and first-positive states for contextual and uniform priors. A shared logarithmic scale shows belief redistribution after observations; processing counts are measured by the replay experiments.}
\Description{Two regional RGB context crops and corresponding belief maps show updates after negative and positive observations.}
\label{fig:s-belief-context}
\end{figure*}

Figure~\ref{fig:s-belief-context} illustrates regional context and belief updates.

\subsection{Pre-adjudication eight-task Mavic replay}
\label{sec:s-preadjudication-mavic}
This section uses the pre-adjudication label version: 2,942 task-frame rows over eight tasks and 2,685 photographs, with 89 full positives, seven partial positives, 2,842 explicit negatives and four unknowns. Excluding eight template-source positives leaves 81 full positives; detection is 59/81 full and 2/7 partial, with zero observed false positives on 2,842 explicit negatives. Unknown and partial rows remain separate from absence. The first independent review is in Section~\ref{sec:s-mavic-review}; final adjudicated labels are in Section~\ref{sec:s-review-expanded}.

Table~\ref{tab:mavic-complete} gives the pre-adjudication endpoint counts on their original label population.

\begin{table*}[tp]
\centering\small
\caption{Pre-adjudication eight-task replay at the fixed causal quota. Full/visible support uses full versus full-or-partial positive images under that label version. Final adjudicated endpoints are reported in Section~\ref{sec:s-review-expanded}.}
\label{tab:mavic-complete}
\begin{tabular}{lrrr}
\toprule
Policy & Image events & Full support & Visible support \\
\midrule
Belief & 434 & 8/8 & 8/8 \\
Belief + negative & 422 & 8/8 & 8/8 \\
Static prior & 445 & 8/8 & 8/8 \\
No context & 568 & 5/8 & 6/8 \\
Uniform & 603 & 3/8 & 4/8 \\
Novelty & 526 & 5/8 & 5/8 \\
Fixed phase & 603 & 3/8 & 4/8 \\
Random & 618 & 4/8 & 4/8 \\
All frames & 1762 & 8/8 & 8/8 \\
\bottomrule
\end{tabular}
\end{table*}

Static prior, positive updates and combined positive/negative updates use 445, 434 and 422 events with 8/8 full-supported confirmations. The four additional tasks use 199/197/195 events with 4/4 full support. A 1,000-condition replay produces 408 automatic confirmations, 394 full-supported and 14 partial-supported, while reusing 90 confirming pairs and 56 images. Those conditions and the 240 prior-perturbation runs reuse the same eight tasks.

\paragraph{Annotation and image provenance.}
The pre-adjudication review export is not two blinded independent initial label sets. In the coordinate system defined by each record's image width and height, 170/174 nonempty boxes reproduce prefilled guides and 115/174 exceed the recorded bounds. These guide boxes are not independent ground truth and are excluded from IoU/mAP validation; the results above use image-level labels. Final independent review and box evidence are reported separately in Section~\ref{sec:s-review-expanded}.

For D02-new frame 138, a clean-image recheck changes the score from 0.481752 to 0.284470 while detection remains negative. Only score and detection status are recomputed from the clean image; inlier, coverage and box fields originate from the marked input and are excluded from geometric validation. This recheck leaves binary detections and confirmations in the pre-adjudication analysis unchanged. A separate cross-flight template test covers one same-instance case. Original-image availability is specified in Section~\ref{sec:s-current-availability}.

\subsection{Partial-label Mavic detector protocol and timing}
\label{sec:s-partial-mavic}
The partial-label population is the basis for the timing runs in this section; final adjudicated image-label results use the separate protocol in Section~\ref{sec:s-review-expanded}.

Seven physical routes contain 2,685 photographs; two G01 tasks share 257 photographs, giving 2,942 task-label rows. Table~\ref{tab:s-mavic-labels} reports fully visible positives under the partial-label version after task-specific template-source exclusion.

\begin{table*}[tp]\centering\small
\caption{Partial-label detector results. Strict overlap additionally requires at least half the predicted-box area inside the annotation. Unboxed and partial frames are not known negatives; final labels are reported in Section~\ref{sec:s-review-expanded}.}
\label{tab:s-mavic-labels}

\begin{tabular}{lrrrr}
\toprule

Task & Frames & Eligible positives & Center inside & Strict overlap \\
\midrule

D02 old & 356 & 5 & 5 & 5 \\

D02 new & 362 & 8 & 3 & 3 \\

QX WL2 & 678 & 9 & 9 & 9 \\

e0928 & 422 & 7 & 7 & 7 \\

G01 target 1 & 257 & 3 & 3 & 3 \\

G01 target 2 & 257 & 5 & 4 & 3 \\

Route 008 & 299 & 9 & 9 & 9 \\

Route 009 & 311 & 6 & 5 & 5 \\

\bottomrule
\end{tabular}
\end{table*}

The detector uses rotated/multiscale proposals, RootSIFT, affine RANSAC and masked enhanced-correlation alignment. Admission requires at least five independent inliers, reference coverage at least 0.015 and gray/color score at least 0.60. Templates come from the evaluated flights; source-frame exclusion prevents self-matching, with temporal correlation and preflight availability untested. These partial labels lack independent second annotations; final independent judgments and adjudications use the separate protocol in Section~\ref{sec:s-review-expanded}.

The causal allocator uses quota $\min(120,1+\lfloor t/8\rfloor)$, current time, pose, footprint, prior and processed observations; future route length and terminal duration are excluded. Processing requires a periodic, belief-density or pending-confirmation trigger. Two consecutive processed positives within 30\,s confirm; skipped frames do not reset. The 48 cached policy replays and 72 prefix checks reproduce selections, quota, confirmation and belief hashes. Table~\ref{tab:s-mavic-causal} separates development routes from the four additional tasks.

\begin{table*}[tp]\centering\small
\caption{Automatic replay under the partial-label protocol. Supported means both confirming observations have fully visible annotations; unresolved partial/identity cases remain separate.}
\label{tab:s-mavic-causal}

\begin{tabular}{llrrr}
\toprule

Cases & Policy & Events & Supported & Pending \\
\midrule

Development & All frames & 940 & 4/4 & 0 \\

Development & Causal, no context & 361 & 2/4 & 1 \\

Development & Causal context & 237 & 4/4 & 0 \\

Development & Causal context + diagnostic negatives & 227 & 4/4 & 0 \\

Additional tasks & All frames & 822 & 4/4 & 0 \\

Additional tasks & Causal, no context & 207 & 2/4 & 1 \\

Additional tasks & Causal context & 197 & 4/4 & 0 \\

Additional tasks & Causal context + diagnostic negatives & 195 & 4/4 & 0 \\

\bottomrule
\end{tabular}
\end{table*}

The partial-label timing experiment has 40 records: four tasks, five policies and two passes on three physical routes. Each pass reads/decodes clean exports, runs the frozen detector on selected frames and updates belief. Results are reaggregated from recorded outputs; the original photographs, worker logs and cached/live parity traces needed to rerun inference and timing are unavailable (Section~\ref{sec:s-current-availability}). The 16-job image collection is a separate processing cohort.
Table~\ref{tab:s-mavic-timing} gives four-task sums averaged over two passes. The 75.80\% time reduction uses all-frame stop-on-confirmation as its denominator. At the same causal quota, context reduces time by 6.64\% while supported confirmation changes from 2/4 to 4/4; that contrast has different outcome quality.

Peak resident set size (RSS) measures the memory footprint of the recorded processing pass.

\begin{table*}[tp]\centering\small
\caption{40-record timing cohort under partial labels. Time sums four tasks; ranges are two observed passes, not confidence intervals over flights. Final label denominators do not reweight these timings.}
\label{tab:s-mavic-timing}

\begin{tabular}{lrrrrr}
\toprule

Policy & Events & Supported & Time (s) & Pass range (s) & Peak RSS (MiB) \\
\midrule

All frames & 822 & 4/4 & 659.09 & 650.92--667.25 & 146.57 \\

No context, recorded horizon & 356 & 3/4 & 281.86 & 271.59--292.13 & 145.69 \\

Context, recorded horizon & 243 & 4/4 & 191.82 & 191.76--191.88 & 145.52 \\

No context, causal & 207 & 2/4 & 170.87 & 170.60--171.14 & 145.20 \\

Context, causal & 197 & 4/4 & 159.53 & 157.72--161.34 & 145.03 \\

\bottomrule
\end{tabular}
\end{table*}

Peak process RSS is similar across policies. Timing passes and shared-flight tasks do not add independent routes. This timing covers image processing and belief updates, excluding motion, flight time, battery, three-dimensional LOS and onboard WM control.

\subsection{Four-route manual-observation replay}
\label{sec:s-manual-replay}
A DJI Mavic 3M with a 20-megapixel red--green--blue (RGB) camera records 5,045 photographs on 15 routes, typically near 90\,m altitude; approximately 95.8\% look vertically downward. The selected manual-observation cohort has four routes and 1,818 frames: D02-old 356, D02-new 362, QX-WL2 678, and e0928 422. Two tasks concern farm machinery and two vehicles. Its single annotation set marks 24 fully and one partially visible frame; other frames have no marked target. The final eight-task label review uses the separate protocol in Section~\ref{sec:s-review-expanded}.

\begin{figure*}[tp]
\centering\includegraphics[width=.94\textwidth]{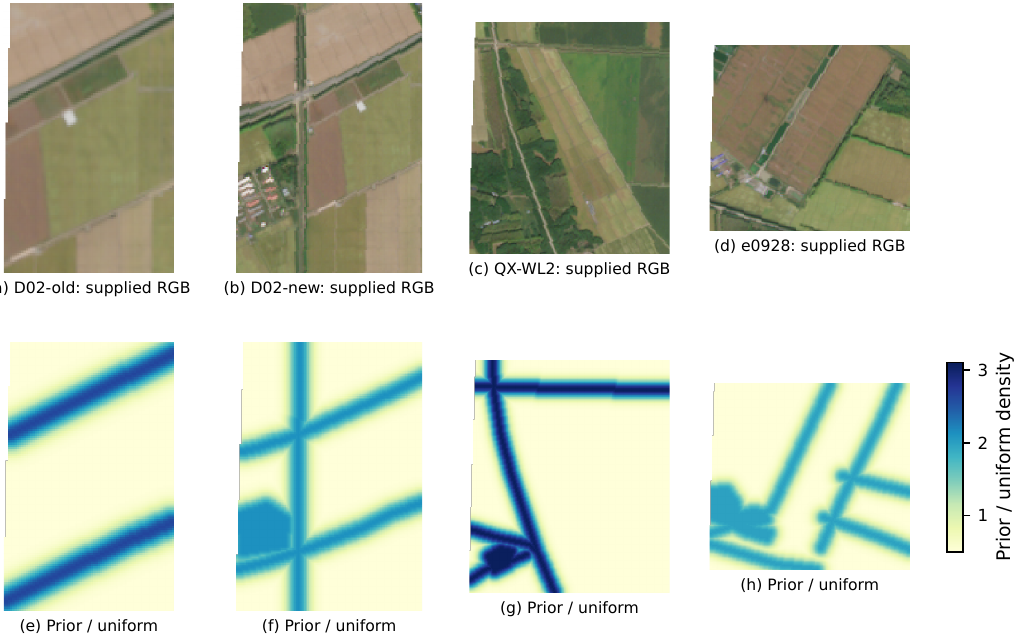}
\caption{Regional context for the four-route manual-observation replay. Top: approximately 10\,m RGB arrays. Bottom: prior density divided by uniform density. These regional cues provide neither individual-target labels nor independent source-product authentication.}
\label{fig:s-context}
\Description{Four historical route-context arrays and their road/building priors. Density is relative to uniform support; product linkage and registration remain separate validation questions.}
\end{figure*}

Four overhead RGB arrays at approximately 10\,m sampling define manually outlined road/building regions and priors (Figure~\ref{fig:s-context}). Their declared export date precedes the flights; sensing time is unverified. These historical RGB exports lack product-linked GeoTIFF metadata and independent registration residuals. Section~\ref{sec:s-satellite-provenance} reports the separate Sentinel catalogue/SCL evidence and the unresolved RGB-to-product linkage. Replay uses regional context and metadata/flat-ground footprints, without the known-obstacle 56-dimensional WM path.

For cell $x$, the prior uses road/building distance with a common 50\,m scale and mixes 50\% uniform support. The route is processed chronologically. Only a selected current frame's label enters the observation interface. The allocator also uses the recorded route duration, an offline scheduling input. A task confirms after two consecutive processed fully visible observations of its class within 30\,s. A processed negative or partial resets the counter; skipped frames do not. This is class-level confirmation on a fixed route, not instance verification or autonomous movement.

Table~\ref{tab:s-flight-all} includes all 15 configurations on all four routes. Matched all-frame/two-positive stopping processes 976 frames; the contextual allocator processes 222, 77.25\% fewer, with all four tasks confirmed. Both every-two phases also confirm all four, using 493 and 491 frames. Uniform sampling at 120 frames per route confirms one. Table~\ref{tab:s-flight-detail} reports route-level work and confirmation times: three routes confirm at the same time as the matched all-frame rule; QX-WL2 is six seconds later. These savings measure processed frames.
\begin{table*}[tp]
\centering\small
\caption{All 15 configurations under the four-route manual labels. Work is processed frames until the declared stop. Failed confirmation remains in the route denominator.}
\label{tab:s-flight-all}
\begin{tabular}{lrr}
\toprule
Configuration & Confirmed routes & Total processed frames \\
\midrule
All frames, no early stop & 4/4 & 1,818 \\
All frames, one positive & 4/4 & 813 \\
All frames, two positives & 4/4 & 976 \\
Uniform 120-frame cap & 1/4 & 458 \\
Every two, phase 0 / phase 1 & 4/4; 4/4 & 493; 491 \\
Every five, phases 0/1/2/3/4 & 0/4; 0/4; 1/4; 1/4; 0/4 & 366; 365; 262; 347; 362 \\
No context, no negative update & 2/4 & 377 \\
No context, negative diagnostic & 3/4 & 207 \\
Road/building context, no negative & 4/4 & 222 \\
Road/building context, negative diagnostic & 4/4 & 216 \\
\bottomrule
\end{tabular}
\end{table*}
\begin{table*}[tp]
\centering\small
\caption{Four-route manual-label replay under matched two-positive rules. Delay uses recorded-route time and does not measure flight time saved.}
\label{tab:s-flight-detail}
\begin{tabular}{lrrrrr}
\toprule
Route & Frames & All-frame processed & Context processed & All-frame confirm (s) & Context delay (s) \\
\midrule
D02-old & 356 & 248 & 76 & 1,112 & 0 \\
D02-new & 362 & 219 & 64 & 981 & 0 \\
QX-WL2 & 678 & 169 & 19 & 336 & 6 \\
e0928 & 422 & 340 & 63 & 677 & 0 \\
\midrule
Total & 1,818 & 976 & 222 & -- & -- \\
\bottomrule
\end{tabular}
\end{table*}

\begin{figure*}[tp]
\centering\includegraphics[width=.91\textwidth]{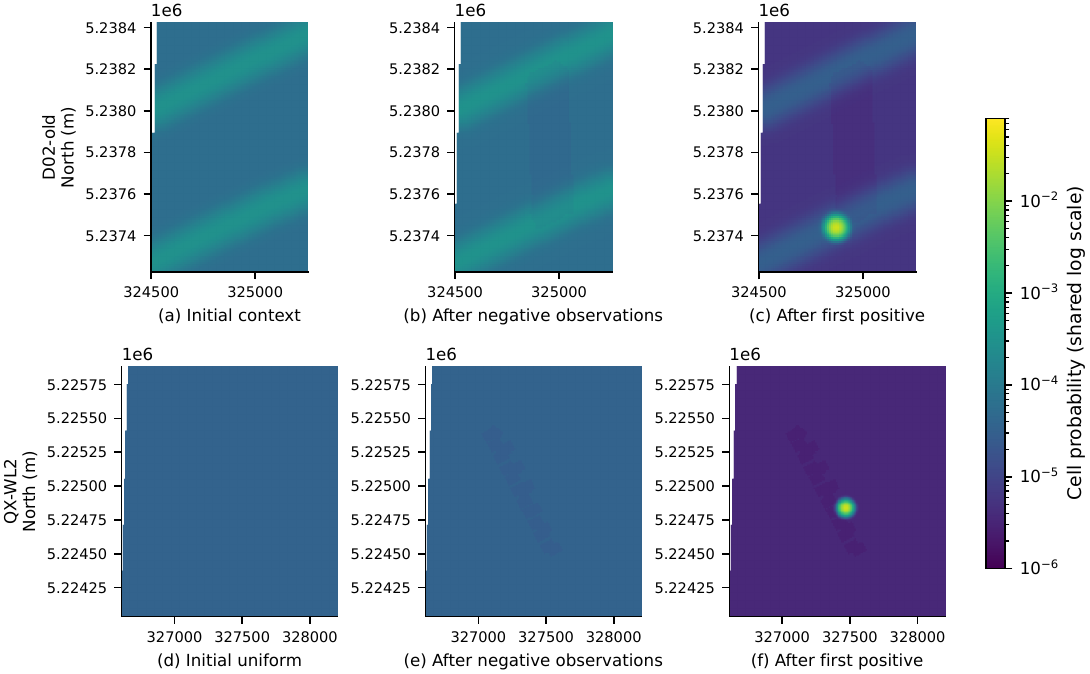}
\caption{Belief updates under the four-route manual-label protocol, separate from the final eight-task review in Section~\ref{sec:s-review-expanded}. Shared logarithmic scales show prescribed mass redistribution after processed observations; independent localization accuracy is not measured.}
\label{fig:s-belief}
\Description{Two four-route manual-label replay cases show prescribed belief changes on shared logarithmic scales. Their labels use the manual-observation protocol, separate from final adjudicated image-label results.}
\end{figure*}

Diagnostic negatives multiply each newly covered cell by 0.8 once and renormalize, alongside positive updates. Without context, this toggle raises confirmation from two routes to three but delays D02-new by 545\,s. With context, it saves six frames and delays one route by two seconds. QX-WL2 target projections have half the uniform prior density: context can alter processing allocation without correctly upweighting every target. Figure~\ref{fig:s-belief} shows the prescribed belief updates.

The four routes are selected after acquisition and the D02 areas partly overlap. The observed four-of-four result describes this selected case study. Forty-eight local timing rows repeat four configurations, four routes, and three rounds. They time RGB read/decode, 640$\times$480 preprocessing, manual observations and two-dimensional belief updates; learned detection, WM control, power and flight are excluded.

\clearpage
\subsection{Satellite metadata and regional context}
\label{sec:s-satellite-provenance}
Catalogue and XML metadata identify a Sentinel-2B L2A product for tile 53TLN, processed at 03:49:05 UTC. Sensing is September 20, 2025 at 02:06:49 UTC and catalogue publication is 04:50:35 UTC, both before the recorded camera times. All 2,685 positions and flat-ground footprints lie within the product extent.

The 20\,m scene classification layer (SCL) has 19.8841\% full-tile cloud/cirrus classification. Frozen crop and seven-route masks contain no cloud/cirrus/shadow or invalid class in 4,762 distinct route-union cells (1.9048\,km$^2$); the remaining classes are vegetation, nonvegetated ground and water. These are product classifications, without independent manual cloud validation.

The historical RGB exports cannot be uniquely bound to this product, and independent ground-control registration residuals are unavailable (Section~\ref{sec:s-current-availability}). Reprojection parity and extent containment do not measure registration accuracy. Manually digitized road/building regions therefore supply a broad prior mixed with uniform support; the prior arrays reproduce this distance-mixture construction. Evaluation uses fixed-route replay.

\FloatBarrier

\end{document}